\documentclass[twocolumn]{article}
\usepackage{arxiv}
\usepackage{placeins}
\usepackage{caption}

\usepackage[font={normal}]{caption} 
\usepackage{amssymb} 

\usepackage{algpseudocode} 
\usepackage{subfiles} 
\usepackage{mathrsfs} 
\usepackage{comment}  

\newenvironment{Itemize}%
{
\setlength{\leftmargini}{9pt}%
\begin{itemize}%
\setlength{\itemsep}{0pt}%
\setlength{\topsep}{0pt}%
\setlength{\partopsep}{0pt}%
\setlength{\parskip}{0pt}}%
{\end{itemize}}

\usepackage[utf8]{inputenc}
\usepackage[english]{babel}
\usepackage{float}
\usepackage{subfig}
\usepackage{bbm}
\usepackage{amsmath,amsfonts}
\usepackage{multirow}
\usepackage{array}
\usepackage{arydshln}
\usepackage{graphicx}
\usepackage{booktabs}
\usepackage[numbers,sort&compress,square]{natbib}
\usepackage{xcolor}
\definecolor{DarkGreen}{rgb}{0.43, 0.68, 0.28}
\usepackage{tabularx}
\graphicspath{{./images/}}

\usepackage{blindtext}

\usepackage{autobreak}
\usepackage{amsmath}
\usepackage{enumitem}
\setlist[itemize]{noitemsep,topsep=3pt,leftmargin=*}

\usepackage[original]{abstract} %

\usepackage{balance}

\usepackage[pagebackref,breaklinks,colorlinks]{hyperref}

\newcommand{\name} {TokenCut}
\usepackage[capitalize]{cleveref}
\crefname{section}{Sec.}{Secs.}
\Crefname{section}{Section}{Sections}
\Crefname{table}{Table}{Tables}
\crefname{table}{Tab.}{Tabs.}

\usepackage{colortbl}
\usepackage{pifont}%
\usepackage{makecell}

\definecolor{LightGray}{rgb}{0.9,0.9,0.9}

\definecolor{cssgreen}{rgb}{0.0, 0.5, 0.0}
\definecolor{cssred}{rgb}{1, 0, 0.0}
\newcommand{\feat}{v}
\newcommand{\indicator}{\mathbf{x}}
\newcommand{\indicatory}{\mathbf{y}}
\newcommand{\indicatorz}{\mathbf{z}}

\newcommand{\degree}{\mathbf{d}}

\newcommand{\nbnode}{N}

\newcommand{\graph}{\mathcal{G}}
\newcommand{\node}{\mathcal{V}}
\newcommand{\edge}{\mathcal{E}}
\newcommand{\Degree}{\mathcal{D}}

\newcommand{\connect}{C}

\newcommand{\simi}{S}
\newcommand{\minsimi}{\tau}

\newcommand{\cmark}{\ding{51}}%
\newcommand{\xmark}{\ding{55}}%
\newcommand\Tstrut{\rule{0pt}{2.6ex}}

\renewcommand{\authorversion}{Author version of CVPR22 paper}

\hypersetup{
  colorlinks=true,
  citecolor=cssgreen,
  urlcolor=red,
  pdftitle={Self-Supervised Transformers for Unsupervised Object Discovery using Normalized Cut},
  pdfauthor={Yangtao Wang, Xi Shen, Xu Hu, Yuan Yuan, James L. Crowley, Dominique Vaufreydaz},
  pdfkeywords={Object Discovery, Unsupervised Learning, Transformer},
}

\title{Self-Supervised Transformers for Unsupervised Object Discovery using Normalized Cut}

\author{Yangtao Wang$^{1}$, Xi Shen$^{2,3,}$\thanks{Corresponding Author}~~, Shell Xu Hu$^4$, Yuan Yuan$^5$, James L. Crowley$^1$, Dominique Vaufreydaz$^1$ \\
$^1$ Univ. Grenoble Alpes, CNRS, Grenoble INP, LIG, 38000 Grenoble, France\\
$^2$ Tencent AI Lab \hspace{5mm}
$^3$ LIGM (UMR 8049) - Ecole des Ponts, UPE \\
$^4$ Samsung AI Center, Cambridge \hspace{5mm}
$^5$ MIT CSAIL 
}

\begin{document}

\twocolumn[{%
  \begin{@twocolumnfalse}
    \maketitle
  \end{@twocolumnfalse}
}]

\setcounter{footnote}{0}
\renewcommand{\thefootnote}{\fnsymbol{footnote}}
\footnotetext[1]{Corresponding Author}

        \begin{abstract}
Transformers trained with self-supervision using self-distillation loss (DINO) have been shown to produce attention maps that highlight salient foreground objects. In this paper, we show a graph-based method that uses the self-supervised transformer features to discover an object from an image. Visual tokens are viewed as nodes in a weighted graph with edges representing a connectivity score based on the similarity of tokens. Foreground objects can then be segmented using a normalized graph-cut to group self-similar regions. We solve the graph-cut problem using spectral clustering with generalized eigen-decomposition and show that the second smallest eigenvector provides a cutting solution since its absolute value indicates the likelihood that a token belongs to a foreground object.

Despite its simplicity, this approach significantly boosts the performance of unsupervised object discovery: we improve over the recent state-of-the-art LOST by a margin of 6.9\%, 8.1\%, and 8.1\% respectively on the VOC07, VOC12, and COCO20K. The performance can be further improved by adding a second stage class-agnostic detector (CAD). Our proposed method can be easily extended to unsupervised saliency detection and weakly supervised object detection. For unsupervised saliency detection, we improve IoU for 4.9\%, 5.2\%, 12.9\% on ECSSD, DUTS, DUT-OMRON respectively compared to state-of-the-art. For weakly supervised object detection, we achieve competitive performance on CUB and ImageNet. Our code is available at:
\href{https://www.m-psi.fr/Papers/TokenCut2022/}{https://www.m-psi.fr/Papers/TokenCut2022/}
\end{abstract}

\section{Introduction}

\begin{figure}[!t]
  \centering
  \begin{minipage}[b]{.32\columnwidth}
    \subfloat[DINO~\cite{caron2021emerging}]{\includegraphics[width=\linewidth,
    height=2.5cm]{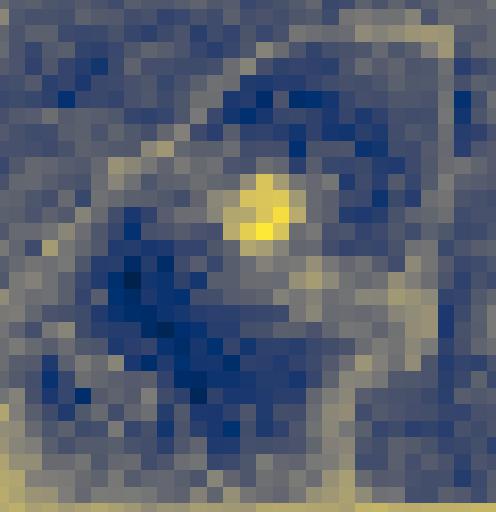}\label{fig:teasera}}
  \end{minipage}
  \hfill
  \begin{minipage}[b]{.32\columnwidth}
    \subfloat[LOST~\cite{simeoni2021localizing}.]{\includegraphics[width=\linewidth,
    height=2.5cm]{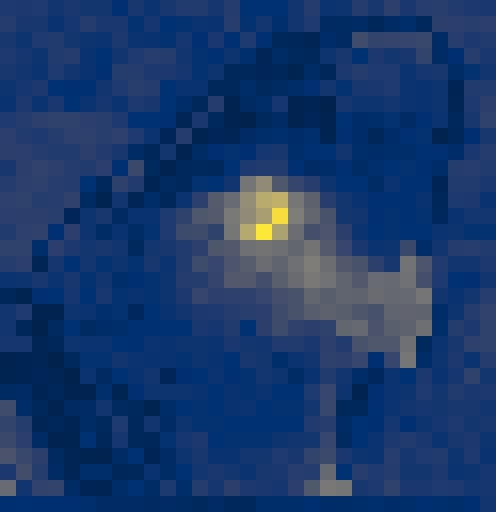}\label{fig:teaserb}}
  \end{minipage}
  \hfill
  \begin{minipage}[b]{.32\columnwidth}
    \subfloat[TokenCut (ours)]{\includegraphics[width=\linewidth,
    height=2.5cm]{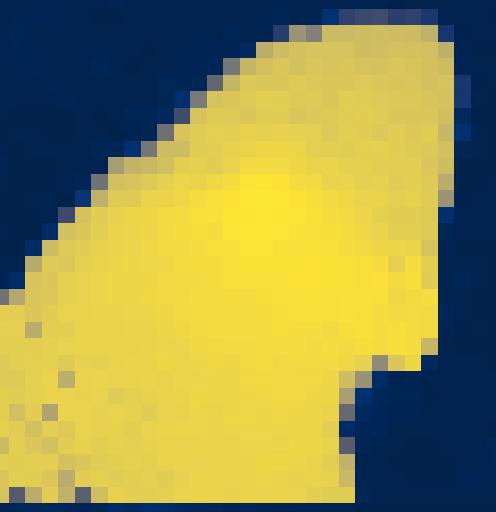}\label{fig:teaserc}}
  \end{minipage}
  \\
  \begin{minipage}[b]{\columnwidth}
    \subfloat[Attention maps associated to different patches ]{\includegraphics[width=\linewidth, height=4cm]{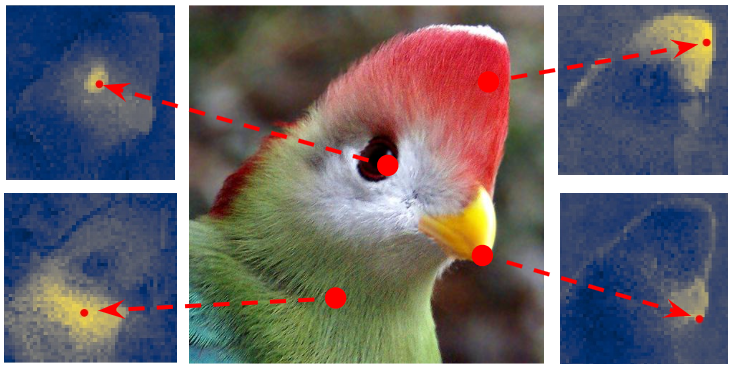}\label{fig:teaserd}}
  \end{minipage}
  \caption{The attention map of the class token used in DINO~\cite{caron2021emerging} (a) and the map of the inverse degrees used in LOST~\cite{simeoni2021localizing} (b) are noisy for foreground / background separation. Our proposed method provides a clean attention map that  can be used to segment salient objects (c). Considering the attention map associated to different patches highlight different regions of the object (d), it is reasonable to build a graph computing attention maps from multiple patches.}
  \label{fig:teaser}%
\end{figure}

Object detection is a key enabling technology for real-world vision systems for tasks such as robotics, autonomous driving, traffic monitoring, manufacturing, and embodied artificial intelligence ~\cite{wu2017squeezedet,geiger2013vision,xia2018dota}. However, the performance of current state-of-the-art object detectors is limited by %
the high cost of annotating sufficient training data~\cite{lin2014microsoft} for supervised learning. This limitation becomes even more apparent when using transfer learning to adapt a pre-trained object detector to a new application domain. Approaches such as active learning~\cite{aghdam2019active}, semi-supervised learning~\cite{liu2021unbiased}, and weakly-supervised learning~\cite{ren2020instance} have attempted to overcome this barrier by providing more efficient learning, but with only limited success. 

In this work, we focus on object discovery in natural images with no human annotations. This is an important problem and a critical step for many downstream applications~\cite{vo2020toward}. 
Poor object discovery can lead to poor overall system performance. Current approaches for this problem adopt some forms of bounding box proposal mechanism~\cite{cho2015unsupervised, wei2019unsupervised,vo2020toward,vo2019unsupervised} and formulate object discovery as an optimization problem. However, this approach can be computationally expensive~\cite{vo2019unsupervised} as each pair of bounding box proposals across different images needs to be compared, and the optimization may fail to scale to larger datasets due to the quadratic computation overhead~\cite{vo2021large}. 

Transformers have recently been shown to outperform convolutional neural networks for visual recognition. Vision Transformers, such as ViT~\cite{dosovitskiy2020image} accept image patches as input tokens and use stacked layers of encoders with self-attention to map tokens to image-level class labels. Recent results with DINO~\cite{caron2021emerging} have shown that when trained with self-distillation loss~\cite{hinton2015distilling}, the attention maps associated to the class token from the last layer indicate salient foreground regions. 
However, as illustrated in Fig.~\ref{fig:teasera}, such attention maps are noisy and it is not clear that they can be used for unsupervised object discovery. 

With LOST~\cite{simeoni2021localizing}, the authors propose construction of a graph and use the inverse degrees of nodes to segment objects.  A heuristic seed expansion strategy is used to overcome  noise (Fig.~\ref{fig:teaserb}) and detect a single bounding box for a foreground object. The attention maps associated with different nodes often contain meaningful information, as illustrated in Fig.~\ref{fig:teaserd}.
We have investigated whether it is possible to use the information in the entire graph by projecting the graph into a low dimensional subspace using eigendecomposition. We have discovered that such a projection can be used with Normalized Cut~\cite{shi2000normalized} (Ncut) to significantly improve  foreground / background segmentation (Fig.~\ref{fig:teaserc}). 

In this paper, we propose \name, a simple but effective graph-based approach for unsupervised object discovery. We build on the self-supervised vision transformer trained with DINO~\cite{caron2021emerging} as our backbone feature encoder and locate objects with the resulting features. Instead of using only the class token, we use all token features. We construct an undirected graph based on the token features in the last self-attention layer, where the visual tokens are viewed as graph nodes with edges representing a connectivity score based on  similarity of the features. We then use a normalized graph-cut to
group self-similar regions and delimit the foreground objects. We solve the graph-cut problem using spectral clustering with generalized eigen-decomposition and show that the second smallest eigenvector provides a cutting solution indicating the likelihood that a token belongs to a foreground object. Our approach can be considered as a run-time adaptation, which means that the model is able to adapt to each specific test image despite the shared training model.

Despite its simplicity, our approach significantly improves unsupervised object discovery. 
The method achieves 68.8\%, 72.1\% and 58.8\% on VOC07~\cite{pascal-voc-2007}, VOC12~\cite{pascal-voc-2012}, COCO20K~\cite{lin2014microsoft} respectively, thus outperforming LOST~\cite{simeoni2021localizing} by a margin of 6.9\%, 8.1\% and 8.1\% respectively.  TokenCut with second stage CAD further improves the performance to 71.4\%, 75.3\% and 62.6\% on VOC07, VOC12, COCO20k respectively, which outperforms LOST + CAD by 5.7\%, 4.9\% and 5.1\% respectively.

In addition, we show that \name~can be easily extended to weakly supervised object detection and unsupervised saliency detection. 
For weakly supervised object detection, the goal is to detect objects using only image-level annotations. 
We freeze the encoder and fine-tune a linear classifier with weakly-supervised image labels. 
We then  apply  \name~on the features extracted from the fine-tuned encoder. Our approach produces clearly improved results on the CUB dataset~\cite{WahCUB_200_2011} and competitive performance on ImageNet-1K~\cite{deng2009imagenet}. For unsupervised saliency detection, we use the foreground region discovered by the proposed approach and apply Bilateral Solver~\cite{barron2016fast} as a post-processing step to refine edges of the foreground region. In terms of results, our approach significantly improves previous state-of-the-art methods on ECSSD~\cite{shi2015hierarchical}, DUTS~\cite{wang2017learning} and DUT-OMRON~\cite{yang2013saliency}. %

In summary, our main contributions are as follows:
\begin{Itemize}

\item We propose a simple and effective method to discover objects in images without supervision based on the self-supervised vision transformers. This method significantly outperforms previous state-of-the-art methods for unsupervised object discovery when tested on multiple datasets;

\item We extend the proposed method to weakly-supervised object detection and show that the simple approach can achieve competitive performance;%

\item We also show that this method can be used for unsupervised saliency detection. The results demonstrate that \name~significantly improves the previous state-of-the-art performance on multiple datasets.

\end{Itemize}

\begin{figure*}[!thbp]
	\centering
	\includegraphics[width=0.8 \textwidth]{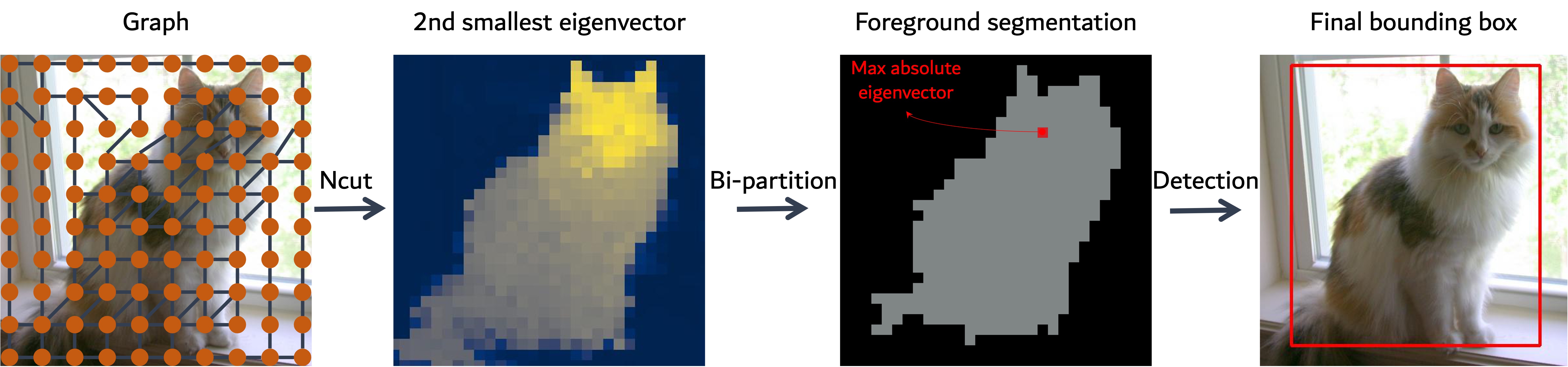}
	\caption{An overview of the TokenCut approach. We construct a graph where the nodes are tokens and the edges are similarities between the tokens using transformer features. The foreground and background segmentation can be solved by Ncut~\cite{shi2000normalized}. Performing bi-partition on the second smallest eigenvector allows to detect foreground object.}
	\label{fig:method}	
\end{figure*}
\section{Related Work}
\paragraph{Self-supervised vision transformers.} ViT~\cite{dosovitskiy2020image} has shown that a transformer architecture~\cite{vaswani2017attention} can be used as an effective encoder for images and provide useful features for supervised vision tasks.  
MoCo-v3~\cite{chen2021empirical} demonstrated that ViT can provide self-supervised representation learning and achieve strong results using contrastive learning. 
Recently, DINO~\cite{caron2021emerging} proposed to train  transformers  with self-distillation loss~\cite{hinton2015distilling}, showing that ViT contains explicit information that can be used for semantic segmentation of an image.
Inspired by BERT~\cite{devlin2018bert}, \cite{li2021mst} proposed MST, which dynamically masks some tokens and learns to recover missing tokens using a global image decoder. Also motivated by BERT~\cite{devlin2018bert}, BEIT~\cite{bao2021beit} first tokenizes the original image into visual tokens then randomly mask some tokens and learn to recover them using a transformer. Recently, MAE~\cite{he2021masked} masks a high proportion of images and reconstructs the missing pixels with an asymmetric encoder-decoder.

\paragraph{Unsupervised object discovery.} Given a group of images, unsupervised object discovery seeks to discover and delimit similar objects that appear in multiple images. Some methods~\cite{joulin2010discriminative,joulin2012multi,vicente2011object,hsu2018co,chen2020show} are designed to segment common repeated objects in an image collection, but rely on strong assumptions about the frequency of appearance of an object. Other approaches~\cite{tang2014co,cho2015unsupervised,vo2019unsupervised,vo2020toward} use bounding box proposals and formulate the object discovery as an optimization problem.
~\cite{vo2021large} proposed a novel formulation of unsupervised object discovery as a ranking problem and showed that discovery could be scaled to datasets with more than 10K images. 
Recently, LOST~\cite{simeoni2021localizing} significantly improved over state-of-the-art for unsupervised object discovery. LOST extracts features using a self-supervised transformer based on DINO~\cite{caron2021emerging} and designs a heuristic seed expansion strategy to obtain a single object region. Our work is closely related to LOST~\cite{simeoni2021localizing},
as we also use self-supervised transformer features. However, rather than  relying on the attention map of some specific nodes, we propose a graph-based method that employs the attention scores of all the nodes and can be used with  Ncut~\cite{shi2000normalized} to obtain a more precise segmentation of the image object.

\paragraph{Weakly supervised object detection.} Weakly supervised object detection~\cite{yuan2019marginalized,YuanyuanICCV2017,guo2021strengthen,zhang2021weakly} can be used to locate image objects using only  image-level annotation. Early approaches~\cite{zhou2016learning,selvaraju2017grad,chattopadhay2018grad} mainly relied on a Class Activation Map (CAM) which is introduced in ~\cite{zhou2016learning} to generate class-specific localization maps and find discriminant regions. Several  methods~\cite{choe2019attention,singh2017hide,mai2020erasing,zhang2018adversarial,yuan2019marginalized, choe2020evaluating} have been proposed to improve CAM by erasing the discriminant regions  and forcing the networks to capture additional object regions. Data augmentation techniques such as Cutout~\cite{devries2017cutout} and CutMix~\cite{yun2019cutmix} have been shown to provide improvement for both classification and localization performance. Some methods achieve both classification and localization using two separate networks~\cite{zhang2020rethinking,lu2020geometry,guo2021strengthen}.~\cite{zhang2020rethinking} trained the localization network using pseudo bounding boxes generated by~\cite{wei2019unsupervised}. ~\cite{zhang2020rethinking} first learns a classifier, then freeze its
weights and train another detector.~\cite{guo2021strengthen} learns a regressor and a classifier using the consistency of CAM between two transformations. Unlike these approaches, which are specifically designed for weakly supervised object detection, we propose an unified solution to both unsupervised object discovery and weakly supervised object detection based on transformer. 

\paragraph{Unsupervised saliency detection.}
Unsupervised saliency detection seeks to segment a salient objects within an image. Earlier work on this problem \cite{yan2013hierarchical, zhu2014saliency, jiang2013salient, li2015weighted} used classical techniques such as color contrast~\cite{cheng2014global},  certain background priors~\cite{wei2012geodesic}, or super-pixels~\cite{li2015weighted, yang2013saliency}. 
More recently,  unsupervised deep models~\cite{wang2017learning, zhang2018deep, nguyen2019deepusps} have  incorporated  heuristic saliency methods as pseudo ground truth to train deep CNN models. However, these methods rely on a CNN model pretrained with supervised training.  \cite{voynov2021object} has proposed an unsupervised Large-Scale GAN that does not make use of labels during training. In the following, we show that incorporating a simple post-processing step into our unsupervised object discovery can provide a strong baseline method for unsupervised saliency detection.

\section{Approach: TokenCut}
\label{sec:approach}

The TokenCut algorithm can be used to predict bounding boxes that locate a salient object in an image. Our approach, illustrated in Fig.~\ref{fig:method}, is based on a graph where the nodes are tokens and the edges are similarities between the tokens using features based on the latent variables of the transformer.
In the following, we first briefly present vision transformers in Section~\ref{sec:visiontransformers} and Normalized Cut in Section~\ref{sec:ncut}. We then introduce our solution and the implementation details in Section~\ref{sec:tokencut}.

\subsection{Background}
\subsubsection{Vision Transformers}
\label{sec:visiontransformers}
Given an image of size $H \times W$, vision transformers (ViT)~\cite{dosovitskiy2020image} take non-overlapping 2D image patches of resolutions $ K \times K $ as inputs, with the number of patches $N=HW/K^{2}$. 
Each patch is represented as a token, described by a vector of numerical features, referred to as an embedding. An extra learnable token, denoted as a class token $ CLS $, is used to represent the aggregated information of the entire set of patches. This $CLS$ token and the set of patch tokens are fed to a standard transformer network with a ``pre-norm" layer normalization~\cite{ba2016layer}.

The vision transformer is composed of a multiple layers of encoders, each with feed-forward networks and multiple attention heads for self-attention, paralleled with skip connections. %
For the unsupervised object discovery task, we use a vision transformer trained with self-supervised learning  using DINO~\cite{caron2021emerging} and extract latent variables from the final layer as the input features for our proposed method.

\begin{figure*}[!t]
\begin{tabular}{c@{\hskip 1.3pt}c@{\hskip 1.3pt}c@{\hskip 1.3pt}c@{\hskip 1.3pt}c@{\hskip 1.3pt}c}

        \includegraphics[width=0.16\textwidth, height=0.14\textwidth]{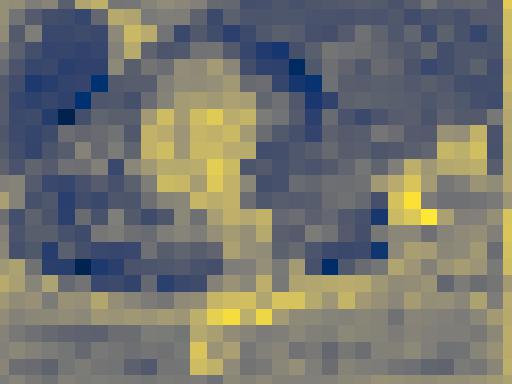} &
		\includegraphics[width=0.16\textwidth, height=0.14\textwidth]{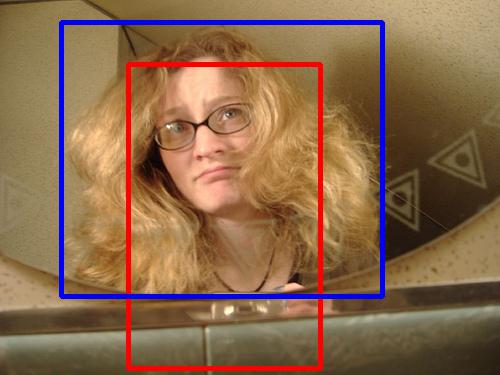} &
		\includegraphics[width=0.16\textwidth, height=0.14\textwidth]{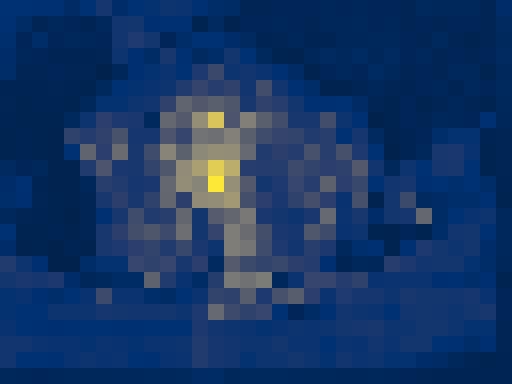} &
		\includegraphics[width=0.16\textwidth, height=0.14\textwidth]{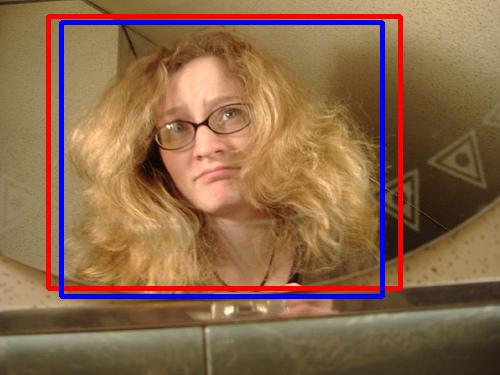} &
		\includegraphics[width=0.16\textwidth, height=0.14\textwidth]{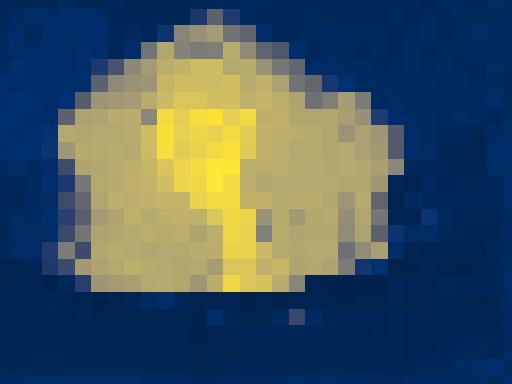} &
		\includegraphics[width=0.16\textwidth, height=0.14\textwidth]{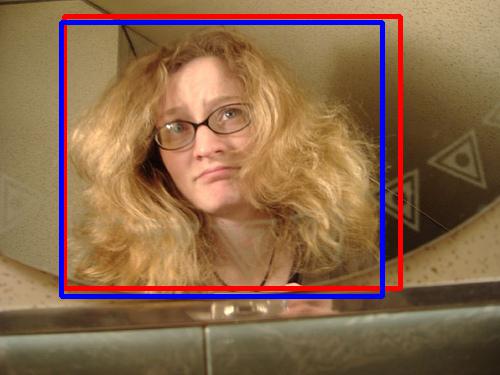} \\ 
		
		\includegraphics[width=0.16\textwidth, height=0.14\textwidth]{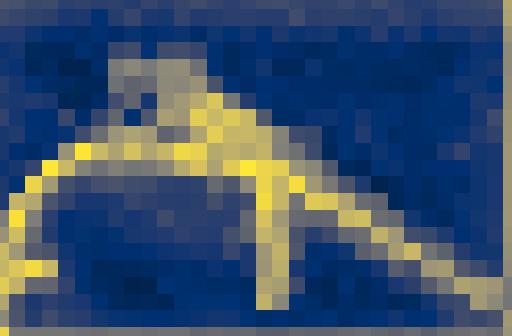} & 
		\includegraphics[width=0.16\textwidth, height=0.14\textwidth]{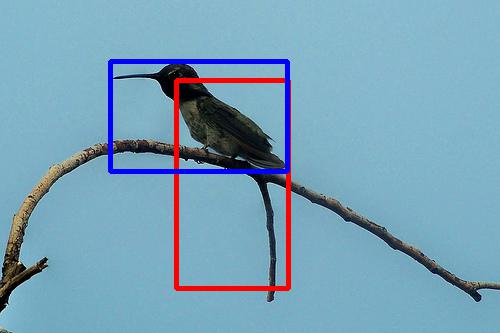} &
		\includegraphics[width=0.16\textwidth, height=0.14\textwidth]{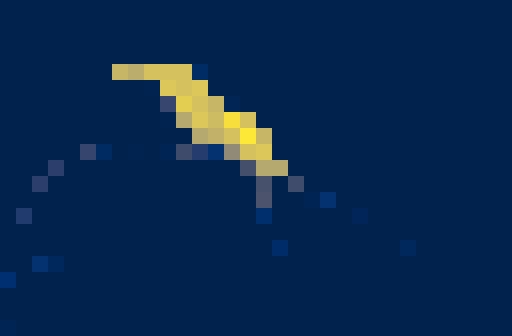} &
		\includegraphics[width=0.16\textwidth, height=0.14\textwidth]{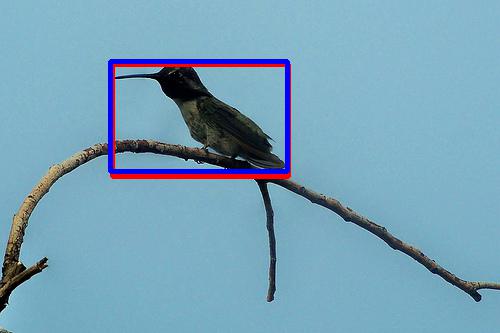} &
		\includegraphics[width=0.16\textwidth, height=0.14\textwidth]{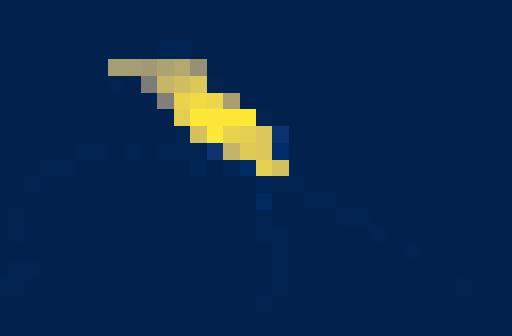} &
		\includegraphics[width=0.16\textwidth, height=0.14\textwidth]{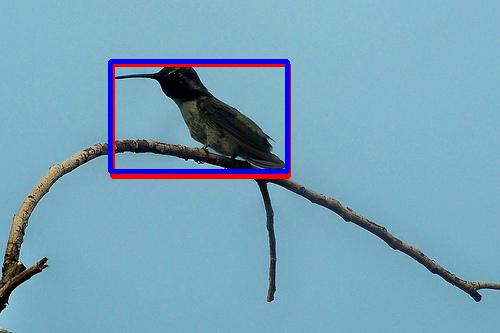} \\
		
		\includegraphics[width=0.16\textwidth, height=0.14\textwidth]{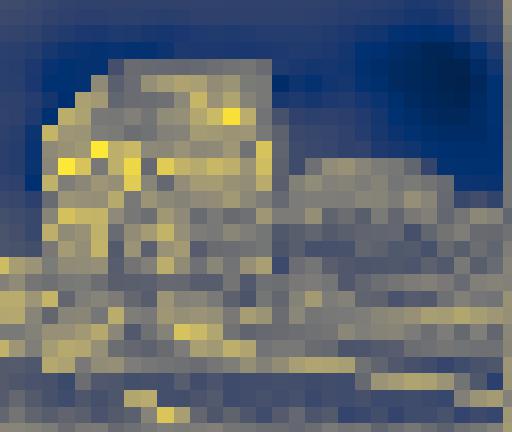} &
		\includegraphics[width=0.16\textwidth, height=0.14\textwidth]{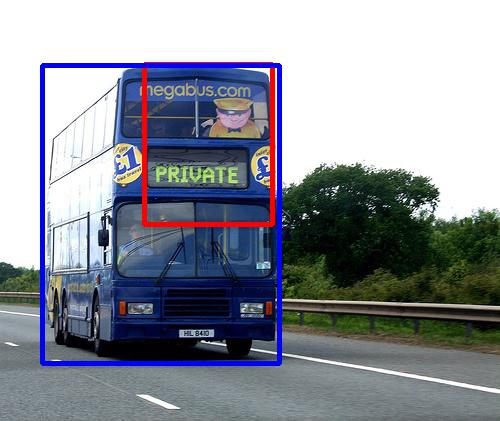} &
		\includegraphics[width=0.16\textwidth, height=0.14\textwidth]{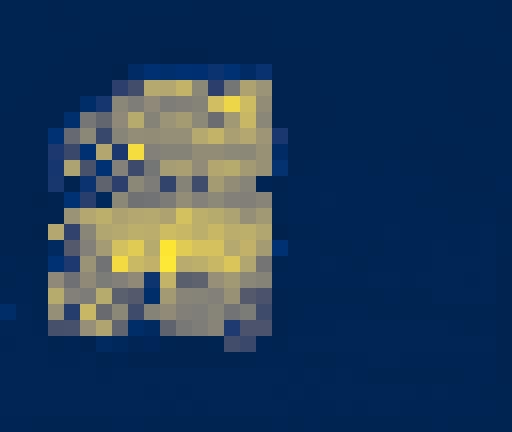} &
		\includegraphics[width=0.16\textwidth, height=0.14\textwidth]{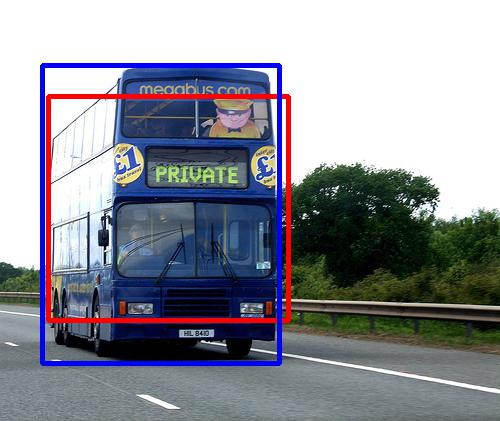} &
		\includegraphics[width=0.16\textwidth, height=0.14\textwidth]{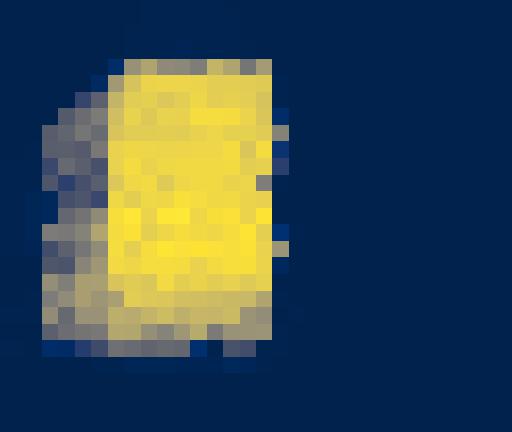} &
		\includegraphics[width=0.16\textwidth, height=0.14\textwidth]{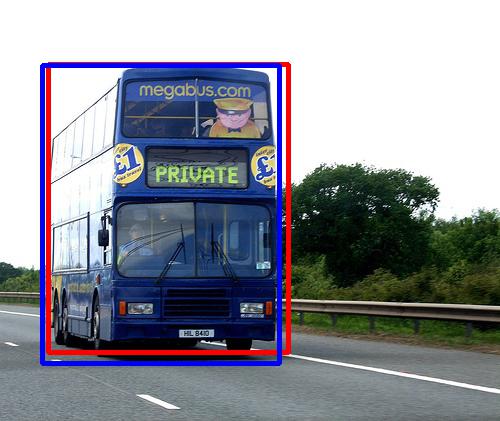} \\
		
		\includegraphics[width=0.16\textwidth, height=0.14\textwidth]{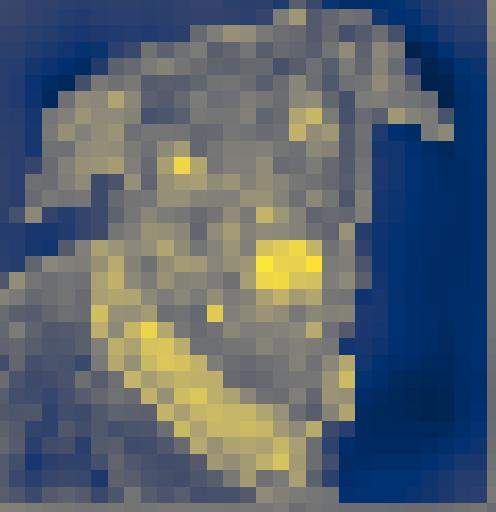} &
		\includegraphics[width=0.16\textwidth, height=0.14\textwidth]{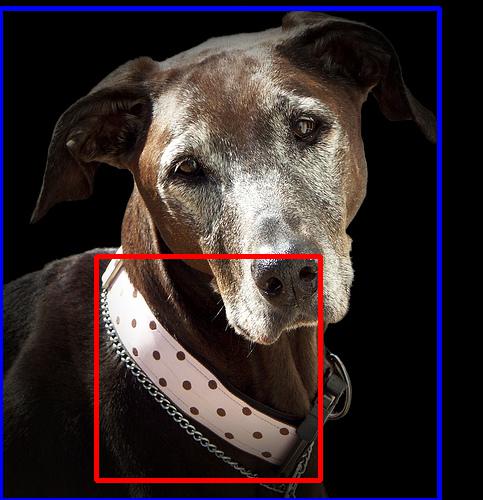} &
		\includegraphics[width=0.16\textwidth, height=0.14\textwidth]{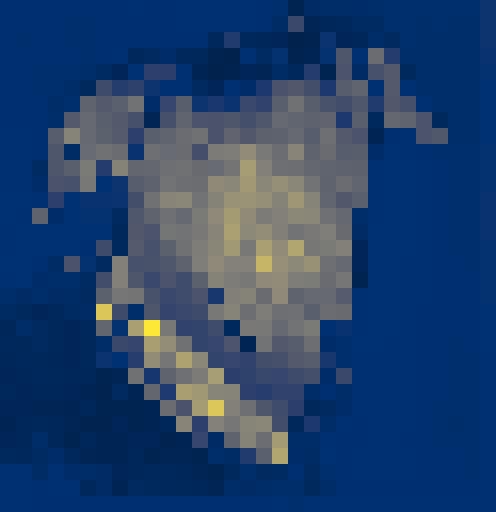} &
		\includegraphics[width=0.16\textwidth, height=0.14\textwidth]{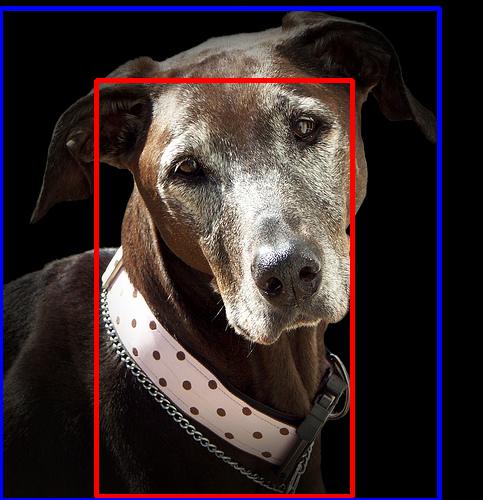} &
		\includegraphics[width=0.16\textwidth, height=0.14\textwidth]{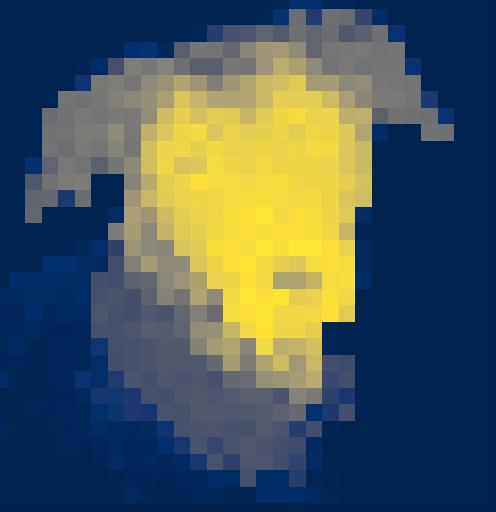} &
		\includegraphics[width=0.16\textwidth, height=0.14\textwidth]{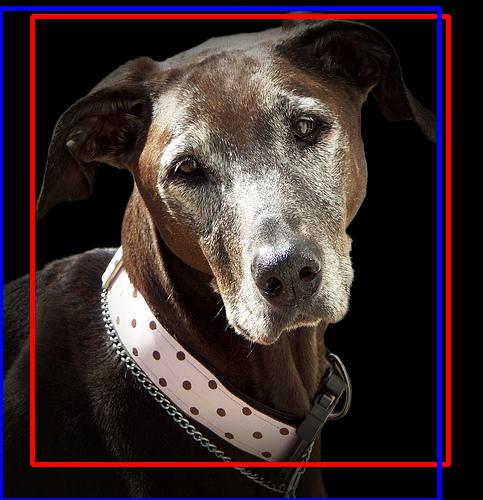} \\
		\makecell{(a) DINO CLS \\ Token Attention}  & \makecell{(b) DINO \\ Detection} & \makecell{(c) LOST Inverse \\Degree Attention} & \makecell{(d) LOST \\ Detection} & \makecell{(e) Our Eigen \\Attention} & \makecell{(f) Our \\Detection} \\
\end{tabular}
\caption{\textbf{Visual results of unsupervised single object discovery on VOC12.} In (a), we show the attention of the CLS token in DINO~\cite{caron2021emerging} which is used for detection (b). LOST~\cite{simeoni2021localizing} is mainly relied on the map of inverse degrees (c) to perform detection (d). For our approach, we illustrate the eigenvector in (e) and our detection in (f). \textcolor{blue}{Blue} and \textcolor{red}{Red} bounding boxes indicate the ground-truth and the predicted bounding boxes respectively. }
\label{fig:visual_res}
\end{figure*}

\subsubsection{Normalized Cut (Ncut)}
\label{sec:ncut}
\paragraph{Graph partitioning.} 
Ncut~\cite{shi2000normalized} can be used to partition a graph into two disjoint sets A and B. 
The method partitions the graph so as to minimizing the Ncut energy~\cite{shi2000normalized}:

\begin{align}
    \label{eqn:ncut_energy}
	Ncut(A, B) = \frac{ \connect(A, B)}{\connect(A, \node)} + \frac{\connect (A, B)}{\connect(B, \node)}
\end{align}

where $\connect$ measures the degree of similarity between two sets. $\connect(A, B) = \sum_{\feat_i \in A, \feat_j \in B} \edge_{i,j}$ and $\connect(A, \node)$ is the total connection from nodes in A to all nodes in the graph.

As shown by Shi and Malik~\cite{shi2000normalized}, the optimization problem in Eqn~\ref{eqn:ncut_energy} is equivalent to:

\begin{align}
	\label{eqn:ncut_final}
	min_{\indicator} Ncut(\indicator) = min_{\indicatory} \frac{\indicatory^T (\Degree - \edge) \indicatory}{\indicatory^T \Degree \indicatory}.
\end{align}
With the condition $\indicatory \in \{1, -b\}^N$,  b satisfies $\indicatory^T \Degree \mathbf{1} = 0$. $\Degree$ is a diagonal matrix with $\degree_i = \sum_{j} \edge_{i,j}$ on its diagonal. 

\paragraph{Ncut solution with the relaxed constraint.}
Taking $\indicatorz = \Degree ^{\frac{1}{2}}\indicatory$. Eqn~\ref{eqn:ncut_final} can be rewrite as:

\begin{align}
	\label{eqn:ncut_z}
	min_{\indicatorz} \frac{\indicatorz^T \Degree ^{-\frac{1}{2}} (\Degree - \edge) \Degree ^{-\frac{1}{2}} \indicatorz}{\indicatorz^T \indicatorz}.
\end{align}
Indicating in~\cite{shi2000normalized}, the formulation in Eqn~\ref{eqn:ncut_z} is equivalent to the Rayleigh quotient~\cite{van1996matrix}, which is equivalent to solve $\Degree ^{-\frac{1}{2}} (\Degree - \edge) \Degree ^{-\frac{1}{2}} \indicatorz = \lambda \indicatorz$, where $\Degree - \edge$ is the Laplacian matrix and known to be positive semidefinite~\cite{pothen1990partitioning}. Therefore $\indicatorz_0 =  \Degree ^{\frac{1}{2}} \mathbf{1}$ is an eigenvector associated to the smallest eigenvalue $ \lambda = 0$. According to Rayleigh quotient~\cite{van1996matrix}, the second smallest  eigenvector $\indicatorz_1$ is perpendicular to the smallest one ($\indicatorz_0$) and can be used to minimize the energy in Eqn~\ref{eqn:ncut_z},

\begin{align}
	\indicatorz_1 = argmin_{ \indicatorz^T \indicatorz_0} \frac{\indicatorz^T \Degree ^{-\frac{1}{2}} (\Degree - \edge) \Degree ^{-\frac{1}{2}} \indicatorz}{\indicatorz^T \indicatorz}. \nonumber 
\end{align}
Taking $\indicatorz = \Degree ^{\frac{1}{2}}\indicatory$,

\begin{align}
	\indicatory_1 = argmin_{ \indicatory^T \Degree \mathbf{1} = 0} \frac{\indicatory^T (\Degree - \edge)  \indicatory}{\indicatory^T \Degree \indicatory}. \nonumber 
\end{align}
Thus, the second smallest eigenvector of the generalized
eigensystem $(\Degree - \edge)  \indicatory = \lambda  \Degree \indicatory$ is the real valued solution to the Ncut~\cite{shi2000normalized} problem.

\subsection{TokenCut Algorithm}
\label{sec:tokencut}

\paragraph{Graph construction.}

Our method uses the vision transformers as described in Section~\ref{sec:visiontransformers} to produce a vector of features for each $ K \times K $ image patch. A fully connected undirected graph $\graph$~=~($\node$, $\edge$) of patches is constructed  where each $ \node $ represents a patch with a feature vector $ \{v_{i}\}_{i=1}^{N} $, and each patch is linked to adjacent patches by labeled edges,  $\edge$. Edge labels represent a similarity score $\simi$ based on the cosine similarity of the feature vectors of the two patches. 

\begin{align}
	\label{eqn:edge}
	\edge_{i,j} =
	\begin{cases} 
		1,  & \mbox{if }\simi(\feat_i, \feat_j) \ge \minsimi \\
		\epsilon, & \mbox{else}
	\end{cases},
\end{align}
where $\minsimi$ is a hyper-parameter and $\epsilon$ equals a small value $1e-5$ to assure that the graph is fully connected and  
$\simi$ is the cosine similarity between features. Note that the spatial information has been implicitly included in the features via positional encoding in transformer. 

\begin{align}
	\label{eqn:cos}
	\simi(\feat_i, \feat_j) = \frac{\feat_i \feat_j}{\lVert \feat_i \rVert_2 \lVert \feat_j \rVert_2}.
\end{align}

We apply Ncut algorithm, as described in Section~\ref{sec:ncut}, on constructed graph $\graph$ and obtain the second smallest eigenvector of the generalized eigensystem, which can be seen as an attention map of the potential objects. We provide visualization of this attention map in Section~\ref{sec:exp}.

\paragraph{Discovering Salient Object with TokenCut.} 
We assume that there is at least one object in the image and the object occupies the foreground region. To successfully segment the foreground objects from the image, we must solve three  problems: 
i) \emph{We must determine a means to partition the graph into two subgraphs} and ii) \emph{given a bi-partition of the graph, we must determine which partition represents the foreground}. iii) \emph{In case of detecting multiple connected components in the foreground, we  must identify the most salient object}.

For the first problem, in our initial experiments we have used a simple average value of the projection onto the second smallest eigenvector to determine the similarity value for cutting the graph $\overline{\indicatory_1} =  \frac{1}{\nbnode} \sum_{i} \indicatory^i_1$. 
Formally, $A = \{\feat_i | \indicatory^i_1 \le \overline{\indicatory_1} \}$ and $B = \{\feat_i | \indicatory^i_1 > \overline{\indicatory_1} \}$. We have compared this to using the classical clustering algorithms of K-means and EM to cluster the second smallest eigen vector into 2 partitions. The comparison is available in the supplementary material Table~\ref{tab:partition}, indicating that the mean generally provides better results. 

For the second problem, the foreground contains the salient object and is assumed to be less connected to the entire graph. Intuitively, $\degree_i < \degree_j$ if $\feat_i$ belongs to the foreground while $\feat_j$ is the background token. Therefore, the eigenvector of the foreground object should have a larger absolute value than the one of the background. We use the maximum absolute value $\feat_{max}$ to select the foreground partition and the most salient object. The partition that contains $\feat_{max}$ is taken as foreground. As our graph has no explicit spatial constraint, the foreground might contain more than one connected regions. We simply select the largest connected component existing in the foreground containing the maximum absolute value $\feat_{max}$ as our final object region.

\begin{table*}[!t]
\begin{center}
    
    \caption{\textbf{Comparisons for unsupervised single object discovery}. We compare TokenCut to state-of-the-art object discovery methods on VOC07~\cite{pascal-voc-2007}, VOC12 ~\cite{pascal-voc-2012} and COCO20K~\cite{lin2014microsoft,vo2020toward} datasets. Model performances are evaluated with CorLoc metric. ``Inter-image Simi.'' means the model leverages information from the entire dataset and explores inter-image similarities to localize objects.}
    
\resizebox{2\columnwidth}{!}{
  \begin{tabular}{lcllll}
    \toprule Method & Inter-image Simi.& DINO~\cite{caron2021emerging} Feat. & VOC07~\cite{pascal-voc-2007} & VOC12 ~\cite{pascal-voc-2012}& COCO20K~\cite{lin2014microsoft,vo2020toward}  \\
    \midrule
    Selective Search~\cite{uijlings2013selective, simeoni2021localizing}& \xmark &  - & 18.8 & 20.9 & 16.0 \\
    EdgeBoxes~\cite{zitnick2014edge, simeoni2021localizing} & \xmark & -& 31.1 & 31.6 & 28.8 \\
    Kim et al.~\cite{kim2009unsupervised, simeoni2021localizing}& \cmark  & -& 43.9 & 46.4& 35.1 \\
    Zhange et al.~\cite{zhang2020object, simeoni2021localizing}& \cmark & -& 46.2 & 50.5 & 34.8 \\
    DDT+~\cite{wei2019unsupervised, simeoni2021localizing}& \cmark & -& 50.2 & 53.1 & 38.2 \\
    rOSD~\cite{vo2020toward, simeoni2021localizing} &\cmark & -&  54.5 & 55.3 & 48.5 \\
    LOD~\cite{vo2021large, simeoni2021localizing}&\cmark &  -& 53.6 & 55.1 & 48.5 \\
    DINO-seg~\cite{caron2021emerging,simeoni2021localizing}& \xmark & ViT-S/16~\cite{dosovitskiy2020image} &  45.8 & 46.2 & 42.1 \\
    LOST~\cite{simeoni2021localizing}&  \xmark &  ViT-S/16~\cite{dosovitskiy2020image} & 61.9 & 64.0 & 50.7 \\
    \bf TokenCut & \xmark &  ViT-S/16~\cite{dosovitskiy2020image} & \bf 68.8 (\textcolor{cssgreen}{$\uparrow$ \bf 6.9}) &  \bf 72.1 (\textcolor{cssgreen}{$\uparrow$ \bf 8.1}) &  \bf 58.8 (\textcolor{cssgreen}{$\uparrow$ \bf 8.1})\\ 
    \midrule
    LOD + CAD$^{\star}$~\cite{simeoni2021localizing} & \cmark & -& 56.3 & 61.6 & 52.7 \\
    rOSD + CAD$^{\star}$~\cite{simeoni2021localizing} & \cmark & -& 58.3 & 62.3 & 53.0 \\
    LOST + CAD$^{\star}$~\cite{simeoni2021localizing} & \xmark & ViT-S/16~\cite{dosovitskiy2020image}& 65.7 & 70.4 & 57.5 \\
    \bf TokenCut + CAD$^{\star}$~\cite{simeoni2021localizing} & \xmark & ViT-S/16~\cite{dosovitskiy2020image}& \bf 71.4 (\textcolor{cssgreen}{$\uparrow$ \bf 5.7}) & \bf 75.3 (\textcolor{cssgreen}{$\uparrow$ \bf 4.9})&  \bf 62.6 (\textcolor{cssgreen}{$\uparrow$ \bf 5.1}) \\
\bottomrule

\end{tabular}}
  
  \footnotesize{$^{\star}$ +CAD indicates to train a second stage class-agnostic detector with ``pseudo-boxes'' labels.} 
  \label{tab:object_dis}
\end{center}

\end{table*}

In summary, the TokenCut algorithm consists of the following steps:

\begin{enumerate}
    
	\item Given an image, build a graph $\graph$ = ($\node$, $\edge$) according to Equation~\ref{eqn:edge} and~\ref{eqn:cos}.
	 
	\item Solve the generalized
	eigensystem $(\Degree - \edge)  \indicatory = \lambda  \Degree \indicatory$ for the eigenvector associated to the second smallest eigenvalue $\indicatory_1$.
	 
	\item Compute bi-partition using the average over $\indicatory_1$: $\overline{\indicatory_1} = \sum_{i} \frac{\indicatory^i_1}{\nbnode}$. $A = \{\feat_i | \indicatory^i_1 \le \overline{\indicatory_1} \}$ and $B = \{\feat_i | \indicatory^i_1 > \overline{\indicatory_1} \}$
	 
	\item Find the largest connected component associated to the maximum absolute value of  $\indicatory_1$.
	
\end{enumerate}

\paragraph{Implementation details.}
For our experiments, we use the ViT-S/16 model~\cite{dosovitskiy2020image} trained with self-distillation  loss (DINO)~\cite{caron2021emerging} to extract features of patches. 
We employ the keys features of the last layer as the input features $\feat$. Ablations on different features as well as  transformers trained with self-supervised learning are  provided in the supplementary material Table ~\ref{tab:backbone}. We set $\tau = 0.2$ for all datasets, the dependency on $\tau$ is provided in Section~\ref{sec:abl}. In terms of running time, our un-optimised implementation takes approximately 0.32 seconds to detect a bounding box of a single image with resolution 480 $\times$ 480 on a single GPU QUADRO RTX 8000.

\section{Experiments}
\label{sec:exp}
We evaluate our approach on three tasks: unsupervised single object discovery, weakly supervised object detection and unsupervised saliency detection. We present results of unsupervised single object discovery in Section~\ref{sec:unsupervised_object_discovery}. The results of weakly supervised object detection are in Section~\ref{sec:WSOL}. The results of unsupervised saliency detection in Section~\ref{sec:unsupervised_saliency_detection}. We provide analysis of $\tau$ in Section~\ref{sec:abl}, other ablation studies will be presented in supplementary material.

\subsection{Unsupervised Single Object Discovery}
\label{sec:unsupervised_object_discovery}

\paragraph{Evaluation metric.}

We report performance using the \emph{CorLoc} metric for precise localisation, as used by  ~\cite{deselaers2010localizing, vo2020toward,vo2021large,vo2019unsupervised, wei2019unsupervised,cho2015unsupervised,siva2013looking}. 
CorLoc counts a predicted bounding box as correct if the intersection over union (IoU) score between the predicted bounding box and one of the ground truth bounding boxes is superior to 0.5. 

\begin{table*}[!ht]
    
	\begin{center}
	\caption{\textbf{Comparisons for weakly supervised object localization.} We report Top-1 Cls, GT Loc and Top-1 Loc on CUB~\cite{WahCUB_200_2011} and ImageNet-1K~\cite{deng2009imagenet} datasets. Compared state-of-the-art methods are divided into two groups: with ImageNet-1K \textit{supervised} pretraining and with ImageNet-1K \textit{self-supervised} pretraining.}
	
	\resizebox{1\textwidth}{!}{
		\begin{tabular}{cll|lll|lll}
	    \toprule
			\multirow{2}{*}{Pretrained Dataset} &  \multirow{2}{*}{Method}  & \multirow{2}{*}{Backbone} &  \multicolumn{3}{c|}{CUB~\cite{WahCUB_200_2011}, Acc. (\%)} & \multicolumn{3}{c}{ImageNet-1K~\cite{deng2009imagenet}, Acc. (\%)}  \\
			& & & Top-1 Cls & GT Loc & Top-1 Loc & Top-1 Cls & GT Loc & Top-1 Loc   \\
		   \midrule
		   \multirow{7}{*}{\makecell{ImageNet-1K~\cite{deng2009imagenet} \\ \textcolor{red}{supervised} pretrain}} & CAM~\cite{zhou2016learning} &  GoogLeNet~\cite{szegedy2015going} & 73.8  & -  & 41.1 & 65  & -  & 43.6  \\
		   & HaS-32~\cite{singh2017hide} +~\cite{bae2020rethinkingCAM}&  GoogLeNet~\cite{szegedy2015going} & 75.4 & 61.1 & 47.4 &68.9 & 60.6 & 44.6\\
		   & ADL~\cite{choe2019attention} +~\cite{bae2020rethinkingCAM} &  ResNet50~\cite{he2016deep}& 75.0   & 77.6  & 59.5 & 75.8  & 62.2  & 49.4  \\
		   & ADL~\cite{choe2019attention}& InceptionV3~\cite{szegedy2016rethinking} & 74.6 & - & 53.0 & 72.8 & - & 48.7 \\
		   & I2C~\cite{zhang2020inter}  & InceptionV3~\cite{szegedy2016rethinking} & - & 72.6 & 56 & 73.3 & \bf 68.5 & 53.1\\
		   & PSOL~\cite{zhang2020rethinking}$^{\ddagger}$ & InceptionV3~\cite{szegedy2016rethinking} & - & - & 65.5 & - & 65.2 & 54.8 \\
		   & SLT-Net~\cite{guo2021strengthen}$^{\star}$ & InceptionV3~\cite{szegedy2016rethinking} & 76.4 & 86.5 & 66.1 & \bf 78.1 &  67.6 &  \bf 55.7 \\ 
		  \hline 
		  \Tstrut
		  \multirow{2}{*}{ \makecell{ImageNet-1K~\cite{deng2009imagenet} \\
		  \textcolor{red}{self-supervised} pretrain}} & LOST~\cite{simeoni2021localizing} & ViT-S/16~\cite{dosovitskiy2020image} & \bf 79.5 & 89.7    & 71.3 & 77.0 & 60.0 & 49 \\
		  & \textbf{TokenCut}  & ViT-S/16~\cite{dosovitskiy2020image} & \bf 79.5 & \bf 91.8 (\textcolor{cssgreen}{$\uparrow$ \bf 2.1}) & \bf 72.9 (\textcolor{cssgreen}{$\uparrow$ \bf 1.6}) & 77.0	&  65.4 (\textcolor{cssgreen}{$\uparrow$ \bf 5.4})&  52.3 (\textcolor{cssgreen}{$\uparrow$ \bf 3.3}) \\
		  \bottomrule
	\end{tabular}}
	
	\small{$\star$ uses ten crop augmentations to get final classification results. $\star$ and $\ddagger$ learn a classifer and a detector separately.}

	\label{tab:wsd}
	\end{center}

\end{table*}

\begin{table*}[!ht]

\centering
\caption{\textbf{Comparisons for unsupervised saliency detection} We compare \name~to state-of-the-art unsupervised saliency detection methods on ECSSD~\cite{shi2015hierarchical}, DUTS~\cite{wang2017learning} and DUT-OMRON~\cite{yang2013saliency}. TokenCut achieves better results comparing with other competitive approaches.}

\resizebox{1\textwidth}{!}{
\begin{tabular}{l|ccc|ccc|ccc}
\toprule

\multirow{2}{*}{Method} & \multicolumn{3}{c|}{ECSSD~\cite{shi2015hierarchical}} & \multicolumn{3}{c|}{DUTS~\cite{wang2017learning}}  & \multicolumn{3}{c}{DUT-OMRON~\cite{yang2013saliency}} \\
                        & \multicolumn{1}{c}{$maxF_{\beta}$(\%)} & \multicolumn{1}{c}{IoU(\%)} & \multicolumn{1}{c|}{Acc.(\%)} & \multicolumn{1}{c}{$maxF_{\beta}$(\%)} & \multicolumn{1}{c}{IoU(\%)} & \multicolumn{1}{c|}{Acc.(\%)} & \multicolumn{1}{c}{$maxF_{\beta}$(\%)} & \multicolumn{1}{c}{IoU(\%)} & \multicolumn{1}{c}{Acc.(\%)} \\
\midrule
HS~\cite{yan2013hierarchical}  & 67.3  & 50.8  & 84.7  & 50.4      & 36.9   & 82.6  & 56.1  & 43.3   & 84.3 \\
wCtr~\cite{zhu2014saliency}    & 68.4  & 51.7  & 86.2  & 52.2      & 39.2   & 83.5  & 54.1  & 41.6   & 83.8 \\
WSC~\cite{li2015weighted}      & 68.3  & 49.8  & 85.2  & 52.8      & 38.4   & 86.2  & 52.3  & 38.7   & 86.5 \\
DeepUSPS~\cite{nguyen2019deepusps} & 58.4  & 44.0  & 79.5  & 42.5      & 30.5   & 77.3  & 41.4  & 30.5   & 77.9 \\
BigBiGAN~\cite{voynov2021object}  & 78.2  & 67.2  & 89.9  & 60.8      & 49.8   & 87.8  & 54.9  & 45.3   & 85.6 \\
E-BigBiGAN~\cite{voynov2021object}  & 79.7  & 68.4  & 90.6  & 62.4      & 51.1   & 88.2  & 56.3  & 46.4   & 86.0 \\
LOST~\cite{simeoni2021localizing,shen2021learning}   & 75.8  & 65.4  & 89.5  & 61.1      & 51.8   & 87.1  & 47.3  & 41.0   & 79.7 \\
LOST~\cite{simeoni2021localizing,shen2021learning}+Bilateral Solver~\cite{barron2016fast}   & 83.7  & 72.3  & 91.6  & 69.7      & 57.2   & 88.7  & 57.8  & 48.9   & 81.8 \\                   
\midrule
\bf TokenCut & 80.3	& 71.2	& 91.8	& 67.2 &	57.6 & 	90.3 &	60.0 &	53.3 &	88.0 \\
\bf TokenCut + Bilateral Solver~\cite{barron2016fast} & \bf 87.4 (\textcolor{cssgreen}{$\uparrow$ \bf 3.7})& \bf 77.2(\textcolor{cssgreen}{$\uparrow$ \bf 4.9}) & \bf 93.4 (\textcolor{cssgreen}{$\uparrow$ \bf 1.8})& \bf 75.5(\textcolor{cssgreen}{$\uparrow$ \bf 5.8}) &	\bf 62.4 (\textcolor{cssgreen}{$\uparrow$ \bf 5.2})& \bf 91.4 (\textcolor{cssgreen}{$\uparrow$ \bf 2.7})& \bf 69.7 (\textcolor{cssgreen}{$\uparrow$ \bf 11.9}) & \bf 61.8 (\textcolor{cssgreen}{$\uparrow$ \bf 12.9})& \bf 89.7 (\textcolor{cssgreen}{$\uparrow$ \bf 7.9})\\

\bottomrule
\end{tabular}
}
\label{tab:salient_detection}

\end{table*}

\paragraph{Quantitative Results.}
We evaluate our approach on three commonly used benchmarks for unsupervised single object discovery: VOC07~\cite{pascal-voc-2007} , VOC12~\cite{pascal-voc-2012} and COCO20K ~\cite{lin2014microsoft,vo2020toward}. The qualitative results are provided in Tab.~\ref{tab:object_dis}. We evaluate the CorLoc scores in comparison with previous state-of-the-art single object discovery methods~\cite{uijlings2013selective, zitnick2014edge, simeoni2021localizing, kim2009unsupervised, zhang2020object, wei2019unsupervised, vo2020toward, vo2021large} on VOC07, VOC12, and COCO20K datasets. These methods can be roughly divided into two groups based on whether the model leverages information from the entire dataset and explores inter-image similarities or not. Because of quadratic complexity of region comparison among images, models with inter-image similarities are generally difficult to scale to larger datasets. The selective search~\cite{uijlings2013selective}, edge boxes~\cite{zitnick2014edge}, LOST~\cite{simeoni2021localizing} and  TokenCut do not require inter-image similarities and are thus much more efficient. As shown in the table,  TokenCut consistently outperforms all previous methods on all datasets by a large margin. Specifically, TokenCut improves the state-of-the-art by 6.9\%, 8.1\% and 8.1\% in VOC07, VOC12 and COCO20K respectively using the same ViT-S/16 features.

We also list a set of results that including a second stage unsupervised training strategy to boost the performance, This is referred to as class-agnostic detection (CAD). A CAD is trained by assigning the same ``foreground'' category to all the boxes produced by the first stage single object discovery model. As shown in Tab.~\ref{tab:object_dis}, TokenCut + CAD outperforms the state-of-the-art by 5.7\%, 4.9\% and 5.1\% on VOC07, VOC12 and COCO20k respectively. %

\paragraph{Qualitative Results.}
In Fig.~\ref{fig:visual_res}, we provide visualization for DINO-seg~\cite{caron2021emerging}, LOST~\cite{simeoni2021localizing} and Tokencut. For each method, we visualize the heatmap that is used to perform object detection. For DINO-seg, the heatmap is the attention map associated to the CLS token. For LOST, the detection is mainly based on the map of inverse degree ($\frac{1}{\degree_i}$). For TokenCut, we display the second smallest eigenvector. The visual result demonstrates that Tokencut can extract a high quality segmentation of the salient object. Comparing with DINO-seg and LOST, TokenCut is able to extract a more complete segmentation as can be seen in the first and the third samples in Fig.~\ref{fig:visual_res}. In some other cases, when all the methods have a high quality map, TokenCut has the strongest intensity on the object, this phenomenon can be viewed in the last sample in Fig.~\ref{fig:visual_res}. More visual results can be found in the supplementary material Fig.~\ref{fig:voc07} and Fig.~\ref{fig:coco}.

\subsection{Weakly Supervised Object Localization}
\label{sec:WSOL}

\paragraph{Evaluation metrics.} 
We report three standard metrics: \textit{Top-1 Cls}, \textit{GT Loc} and \textit{Top-1 Loc}. Top-1 Cls represents the top-1 accuracy of image classification. GT Loc is similar to CorLoc in which a predicted box is counted as correct if the IoU score is superior to 0.5 between the predicted bounding box and one of the ground-truth bounding boxes. Top-1 Loc is the most important metric as it considers measuring both the classification and the detection: a predicted bounding box is counted as a true positive if the class of the image is correctly predicted and the IoU is superior to 0.5 between the predicted bounding box and the ground-truth bounding box.

\paragraph{Results.}
\vspace{-10pt}
We use two datasets to evaluate model performances on weakly supervised object localization: CUB-200-2011~\cite{WahCUB_200_2011} (CUB) and ImangeNet-1k~\cite{russakovsky2015imagenet}. The fine-tuning details can be found in supplementary material. In Tab.~\ref{tab:wsd}, we compare TokenCut to the state-of-the-art weakly-supervised object localization approaches on CUB and ImageNet-1K datasets. The methods can be divided to two groups: models initialized with ImageNet-1K \textit{supervised} pre-training~\cite{zhou2016learning, singh2017hide, choe2019attention, zhang2020inter, zhang2020rethinking, guo2021strengthen, bae2020rethinkingCAM} and models inilialized with ImageNet-1K \textit{self-supervised} pre-training~\cite{simeoni2021localizing}. 

On the CUB dataset, TokenCut achieves the best performance over all methods, and outperforms the state-of-the-art LOST method by 2.1\% and 1.6\% on GT Loc and Top-1 Loc. Interestingly, all the ImageNet-1K self-supervised pretraining models are better than the supervised pretrained models. We believe that this is  because supervised pretraining learns a more discriminative representation of the pretrained dataset than  self-supervised pretraining, leading to a reduction in transferability to downstream datasets such as CUB. In comparison,  self-supervised pretraining can learn a more \textit{general} represenatation and thus provides better transferbility.

On the ImageNet-1K dataset,  TokenCut outperforms LOST by 5.4\% and 4.4\% on GT Loc and Top-1 Loc, and achieves a comparable performance with the ImageNet-1K supervised pretrain model. If the downstream task is ImageNet-1K, then the supervised pretraining with ImageNet-1K can provide discriminative features that improve the localization task as they are tuned to the dataset.

\begin{table}[!bh]
	\begin{center}
	\caption{\textbf{Analysis of $\tau$.} We report CorLoc for unsupervised single object discovery on VOC07, VOC12, COCO20K, and Top-1 Loc for weakly supervised object detection on CUB and ImageNet-1K.}  
 	
	\resizebox{0.8\columnwidth}{!}{
    \begin{tabular}{c | ccc | cc}
    \toprule \multirow{2}{*}{$\tau$} & \multicolumn{3}{c|}{CorLoc}& \multicolumn{2}{c}{Top-1 Loc} \\
    & VOC07 & VOC12& COCO20K & CUB & ImageNet-1K \\
    \midrule
     0 &   67.4 & 71.3  & 56.1 & 73.0 & \bf 53.8\\
     0.1 &   68.6  & 72.1  & 58.2 & \bf 73.2 & 53.4\\
     0.2 & \bf 68.8  & \bf 72.1  & \bf 58.8 & 72.9 & 52.3   \\
     0.3 & 67.7  & 72.1  &  58.2 & 70.8 & 50.4 \\
    \bottomrule
    \end{tabular}}
	\label{tab:tau}
	\end{center}

\end{table}

\subsection{Unsupervised Saliency detection}
\label{sec:unsupervised_saliency_detection}

\begin{figure}[!t]

\begin{tabular}{c@{\hskip 1.3pt}c@{\hskip 1.3pt}c@{\hskip 1.3pt}c}

		\includegraphics[width=0.23\columnwidth, height=0.2\columnwidth]{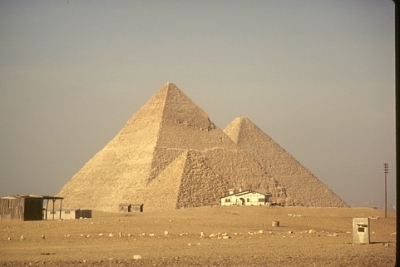} &
		\includegraphics[width=0.23\columnwidth, height=0.2\columnwidth]{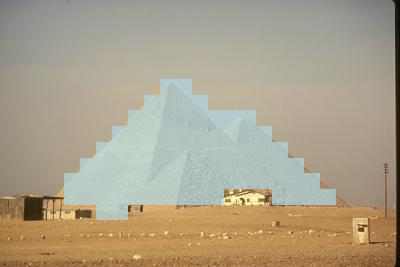} &
		\includegraphics[width=0.23\columnwidth, height=0.2\columnwidth]{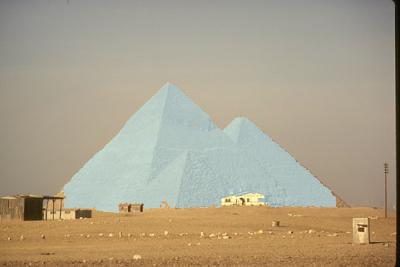} &
		\includegraphics[width=0.23\columnwidth, height=0.2\columnwidth]{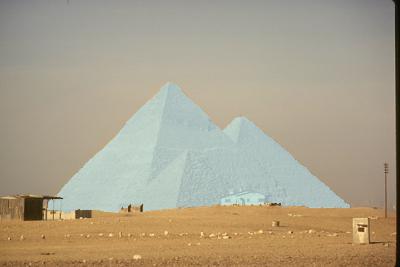} \\
		
		\includegraphics[width=0.23\columnwidth, height=0.2\columnwidth]{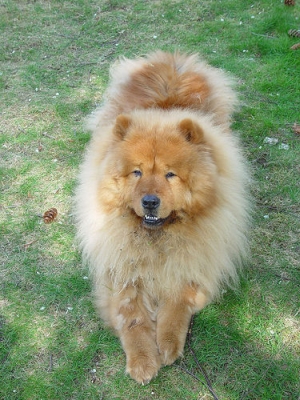} &
		\includegraphics[width=0.23\columnwidth, height=0.2\columnwidth]{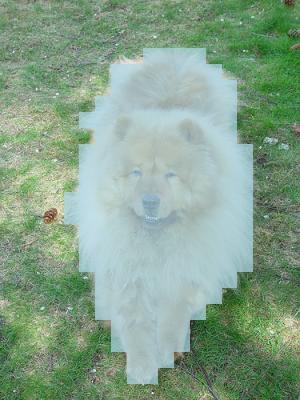} &
		\includegraphics[width=0.23\columnwidth, height=0.2\columnwidth]{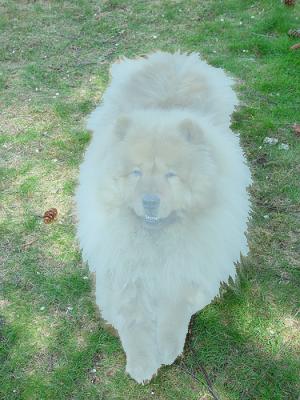} &
		\includegraphics[width=0.23\columnwidth, height=0.2\columnwidth]{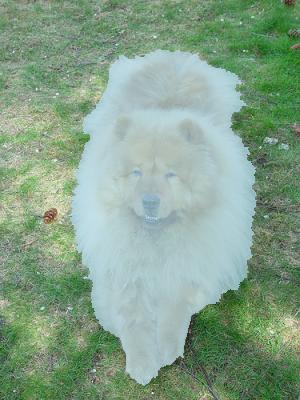} \\

		\makecell{(a) Input} & \makecell{(b) Ours} & \makecell{(c) Ours + BS} & \makecell{(d) GT} \\
\end{tabular}

\caption{\textbf{Visual results of unsupervised segments on ECSSD~\cite{shi2015hierarchical}}. In (a), we show the input image. TokenCut detection result is presented in (b). TokenCut + Bilateral Solver results is shown in (c). (d) is the ground truth.} 
\label{fig: saliency}

\end{figure}

\paragraph{Evaluation Metrics}
We report three standard metrics: F-measure, IoU and Accuracy. \textit{F-measure} is a standard measure in saliency detection. It is computed as $F_\beta = \frac{(1+\beta^2)Precision \times Recall}{\beta^2Precision + Recall}$, where the Precision and Recall are defined based on the binarized predicted mask and ground truth mask. The $maxF_{\beta}$ is the maximum value of 255 uniformly distributed binarization thresholds. Following previous works~\cite{shen2021learning,voynov2021object}, we set $\beta=0.3$ for consistency. \textit{IoU}(Intersection over Union) score is computed based on the binary predicted mask and the ground-truth, the threshold is set to 0.5. \textit{Accuracy} measures the proportion of pixels that have been correctly assigned to the object/background. The binarization threshold is set to 0.5 for masks.

\paragraph{Results}
We further evaluate \name~on three commonly datasets for unsupervised saliency detection: ECSSD~\cite{shi2015hierarchical}, DUTS~\cite{wang2017learning} and DUT-OMRON~\cite{yang2013saliency}. The qualitative results are in Tab.~\ref{tab:salient_detection}. TokenCut significantly outperforms previous state-of-the-art. Adding Bilateral Solver~\cite{barron2016fast} refines the boundary of an object and further boosts the performance over TokenCut, which can also be seen from the visual results presented in Fig.~\ref{fig: saliency}. 

\begin{figure}[!t]

\centering
\begin{tabular}{c@{\hskip 3pt}c@{\hskip 3pt}c}

        \includegraphics[width=0.15\textwidth, height=0.13\textwidth]{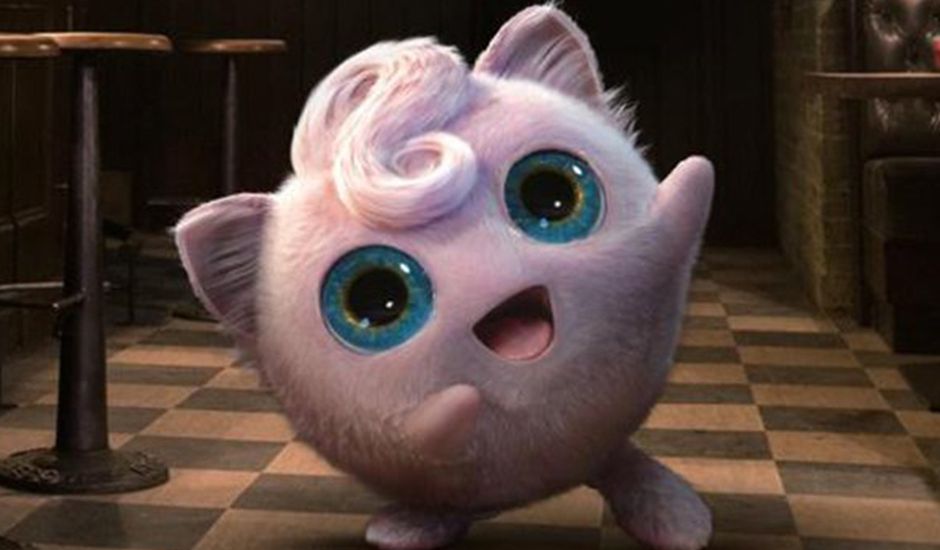} &
        \includegraphics[width=0.15\textwidth, height=0.13\textwidth]{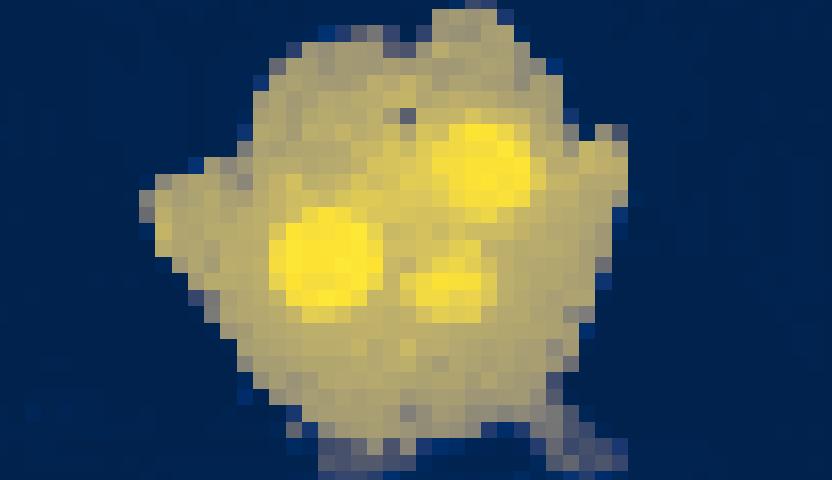}&
        \includegraphics[width=0.15\textwidth, height=0.13\textwidth]{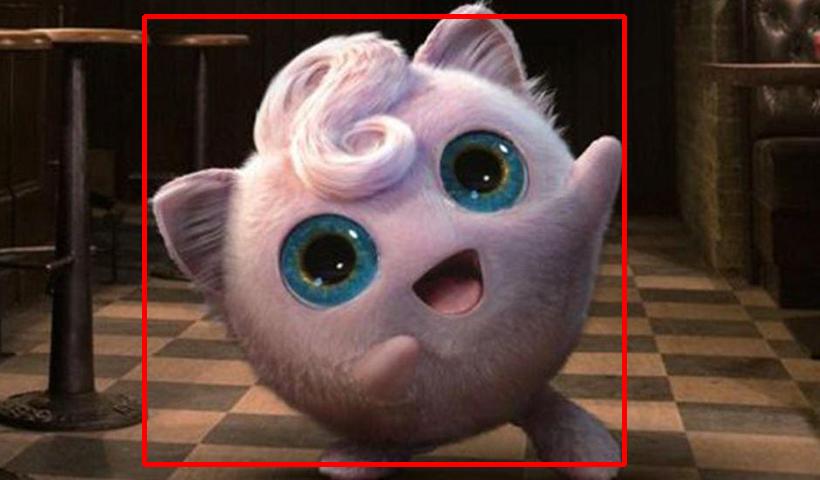} \\	
		
        \includegraphics[width=0.15\textwidth, height=0.13\textwidth]{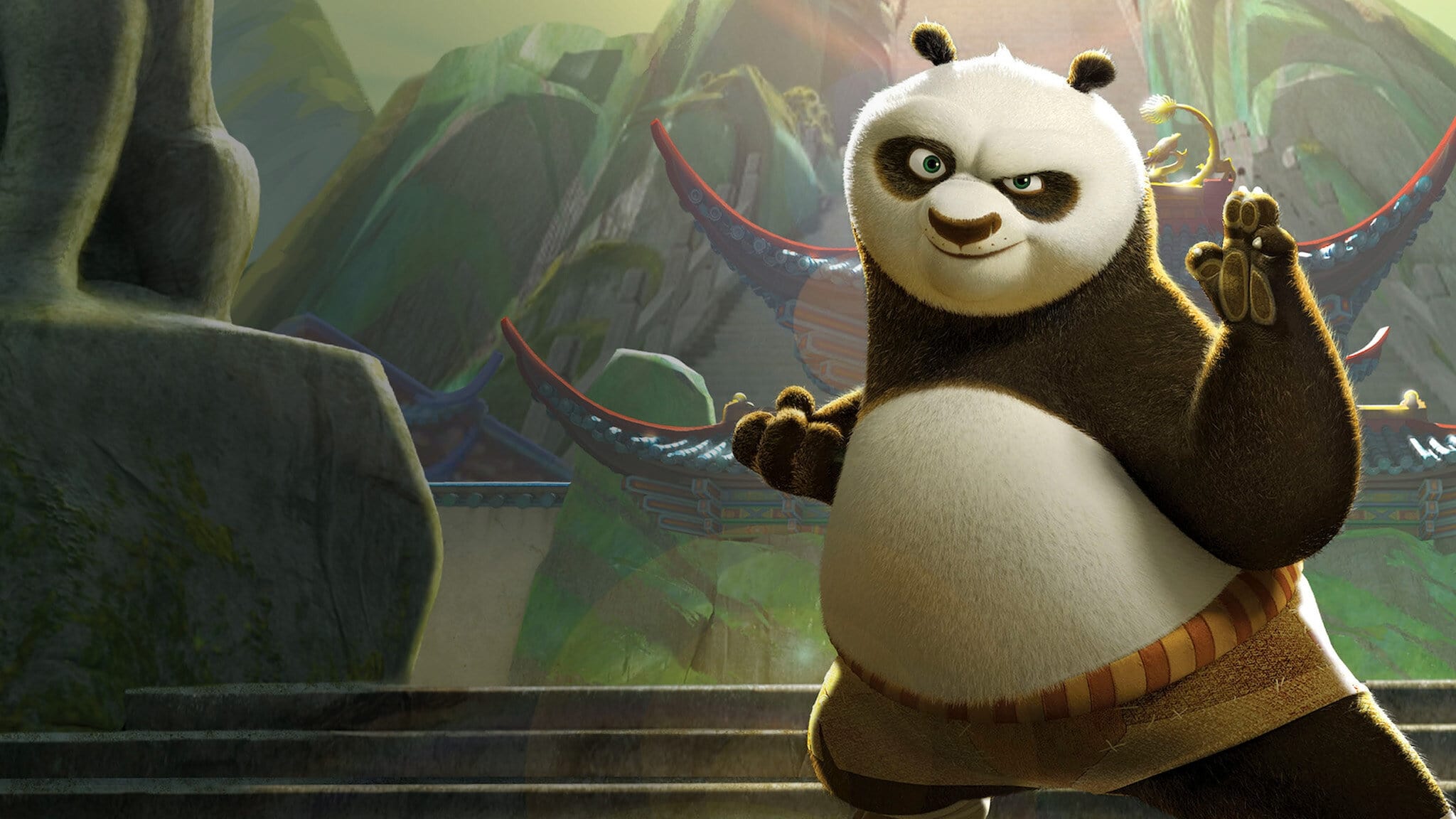} &
        \includegraphics[width=0.15\textwidth, height=0.13\textwidth]{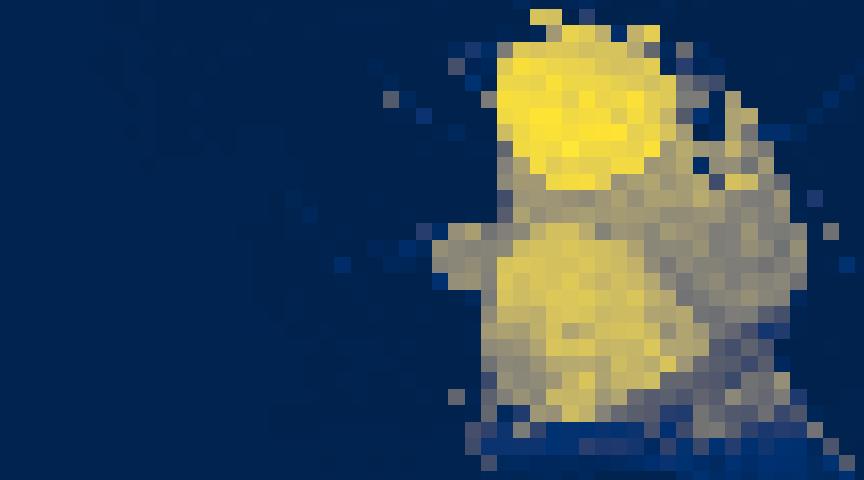}&
        \includegraphics[width=0.15\textwidth, height=0.13\textwidth]{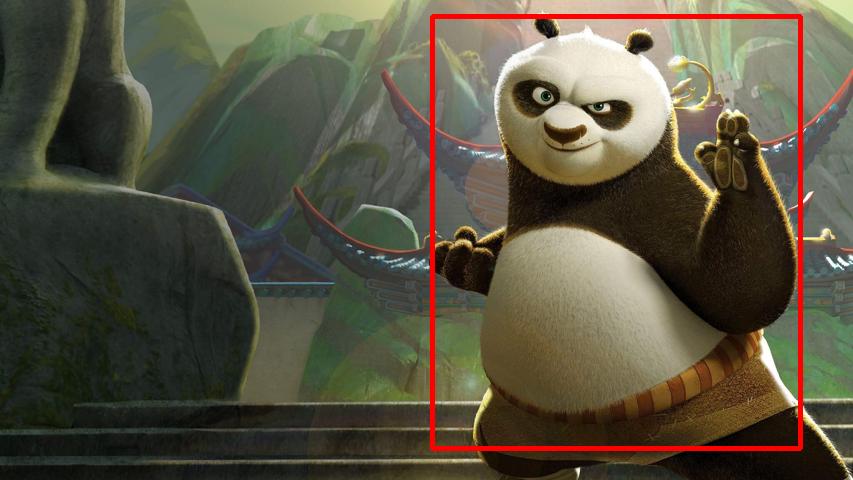} \\
        (a) Input & (b) Eigen Attention & (c) Detection \\ 
\end{tabular}

\caption{\textbf{Visual results of images taken from the Internet.} We show the input images, our eigen attention and final detection in (a) (b) (c) respectively.}
\label{fig:extra_samples}

\end{figure}

\subsection{Analysis and Discussion}
\label{sec:abl}

\paragraph{Analysis of $\tau$.} In Tab.~\ref{tab:tau}, we provide an analysis on $\tau$ defined in Equation~\ref{eqn:edge}. The results indicate that the effects of variations in $\tau$ value are not significant and that a suitable threshold is $\tau$ = 0.2. 

\vspace{-10pt}
\paragraph{Internet Images} We further test \name~ on Internet images. The results are in Figure~\ref{fig:extra_samples}. We can see even the inputs images are with noisy background, our algorithm can still provide precise attention map cover the object leading to accurate bounding box prediction, which demonstrates again the robustness of our approach.

\begin{figure}[!t]
\centering
\begin{tabular}{c@{\hskip 3pt}c@{\hskip 3pt}c@{\hskip 3pt}c}

        \includegraphics[width=0.11\textwidth, height=0.11\textwidth]{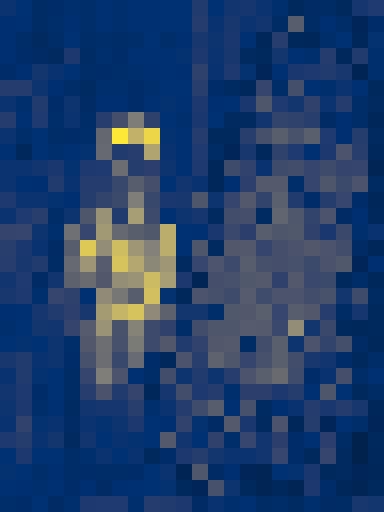} &
		\includegraphics[width=0.11\textwidth, height=0.11\textwidth]{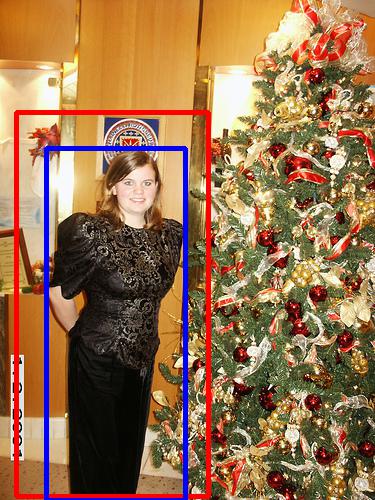} &
		\includegraphics[width=0.11\textwidth, height=0.11\textwidth]{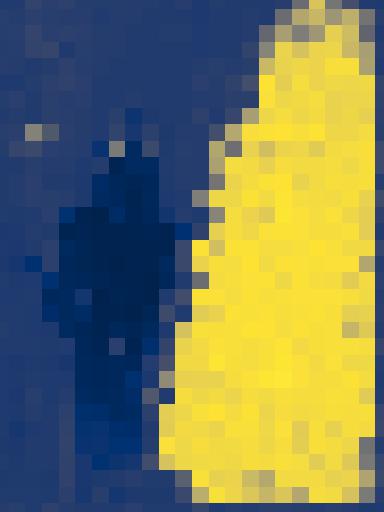} &
		\includegraphics[width=0.11\textwidth, height=0.11\textwidth]{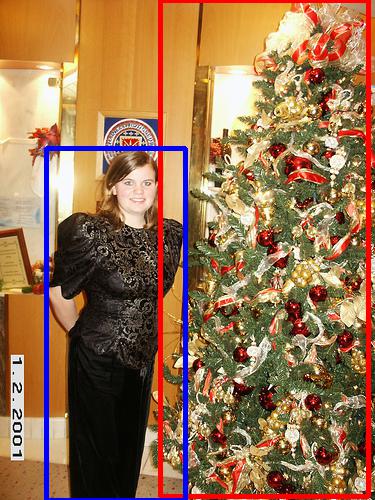}\\	
		
		\includegraphics[width=0.11\textwidth, height=0.11\textwidth]{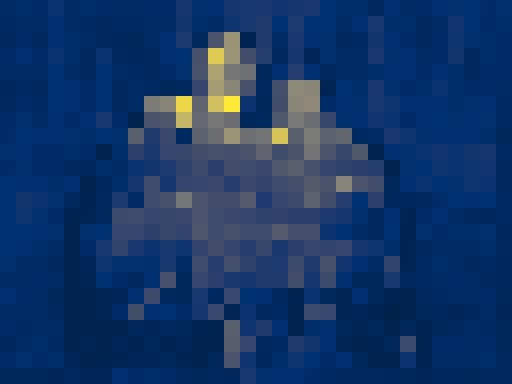} &
		\includegraphics[width=0.11\textwidth, height=0.11\textwidth]{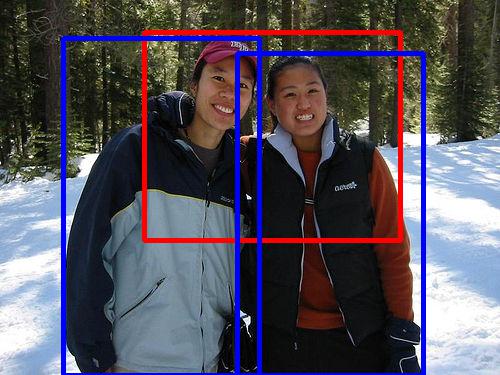} &
		\includegraphics[width=0.11\textwidth, height=0.11\textwidth]{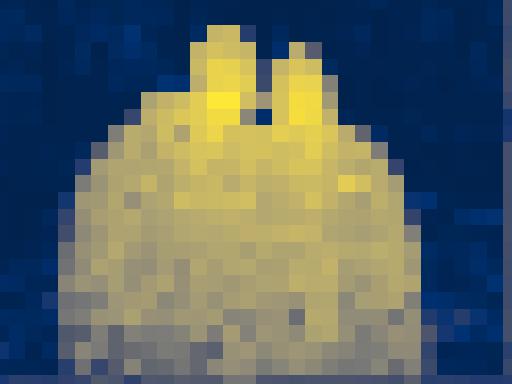} &
		\includegraphics[width=0.11\textwidth, height=0.11\textwidth]{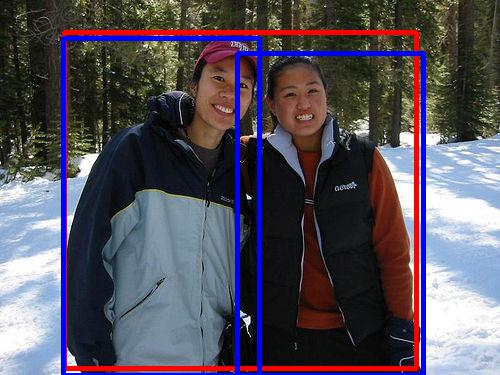}\\
		
		\includegraphics[width=0.11\textwidth, height=0.11\textwidth]{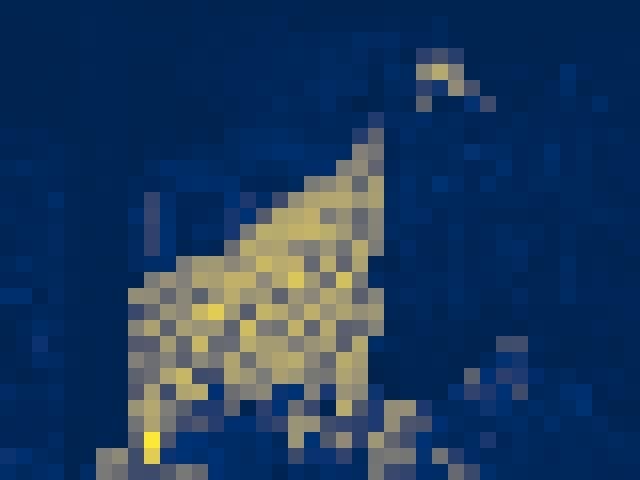} &
		\includegraphics[width=0.11\textwidth, height=0.11\textwidth]{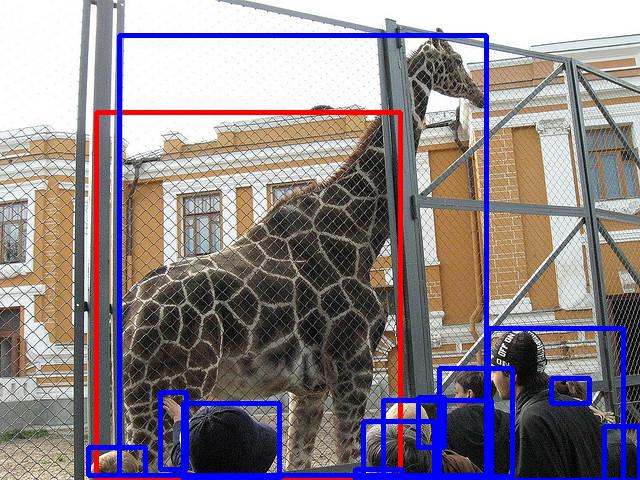} &
		\includegraphics[width=0.11\textwidth, height=0.11\textwidth]{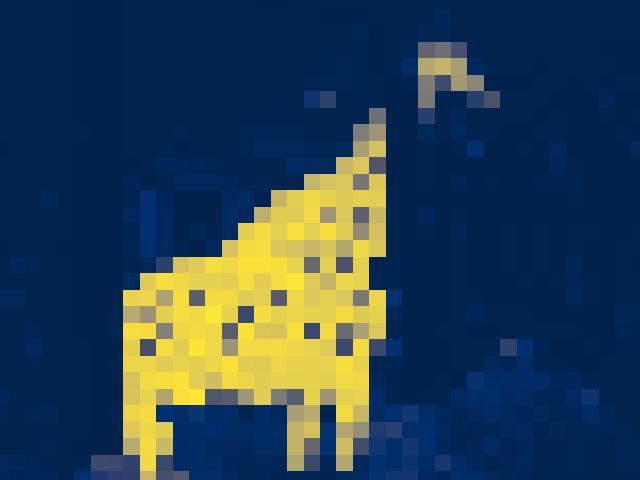} &
		\includegraphics[width=0.11\textwidth, height=0.11\textwidth]{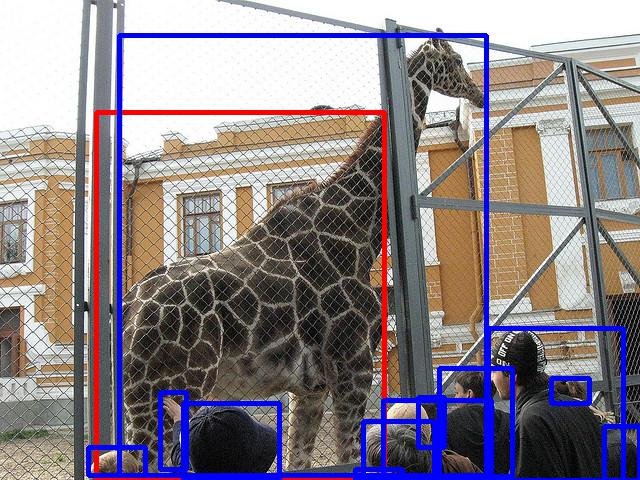}\\
        \makecell{(a) LOST \\ Inverse Attn.} & \makecell{(b) LOST \\Detection} & \makecell{(c) Our Eigen \\ Attention} & \makecell{(d) Our \\Detection} \\ 
\end{tabular}

\caption{\textbf{Failure cases on VOC12 (1st and 2nd row) and COCO (3rd row).} LOST~\cite{simeoni2021localizing}  mainly relies on the map of inverse degrees (a) to perform detection (b). For our approach, we illustrate the eigenvector in (c) and our detection in (d). \textcolor{blue}{Blue} and \textcolor{red}{Red} bounding boxes indicate the ground-truth and the predicted bounding boxes respectively.}
\label{fig:failure}

\end{figure}

\paragraph{Limitations} 
Despite the good performance of TokenCut, it has several limitations. Examples of failure cases are shown in Fig.~\ref{fig:failure}: i)  TokenCut focuses on the largest salient part in the image, which may not be the desired object (Fig.~\ref{fig:failure}, 1st row). ii) Similar to LOST~\cite{simeoni2021localizing}, TokenCut assumes that a single salient object occupies the foreground. If multiple overlapping objects are present in an image, both LOST and our approach will fail to detect one of the object (Fig.~\ref{fig:failure}, 2nd row). iii) Neither LOST nor our approach can handle occlusion (Fig.~\ref{fig:failure}, 3rd row).  

\section{Conclusion}

We have introduced TokenCut, a simple but effective approach for unsupervised object discovery. TokenCut uses self-supervised learning with transformers to constructs a graph where nodes are patches and edges represent similarities between patches. We showed that salient objects can be directly detected and delimited using Ncut. We evaluated this approach on unsupervised single object discovery, weakly supervised object detection and unsupervised saliency detection, and showed that it provides a significant improvement over previous approaches.
Our results indicate that self-supervised transformers can provide a rich and general set of features that may likely be used for a variety of computer vision problems.  

\footnotesize{\textbf{Acknowledgements} This work has been partially supported by the  MIAI  Multidisciplinary AI Institute at the Univ. Grenoble Alpes (MIAI@Grenoble Alpes - ANR-19-P3IA-0003),  and by the EU H2020 ICT48 project Humane AI Net under contract EU \#952026.}
{\small
\bibliographystyle{ieee_fullname}
\bibliography{egbib}
}

\clearpage
\onecolumn
\appendix

\section{Analysis of backbones.}

In Tab.~\ref{tab:backbone}, we provide an ablation study on different transformer backbones. The \textit{``-S''} and \textit{``-B''} are ViT small\cite{dosovitskiy2020image, caron2021emerging} and ViT base\cite{dosovitskiy2020image, caron2021emerging} architecture respectively. The \textit{``-16''} and \textit{``-8''} represents patch sizes 16 and 8 respectively. The \textit{``MocoV3''} is another pre-trained self-supervised transformer model~\cite{chen2021empirical}. The $\tau$ value is set to 0.3 for MoCov3, while for Dino the best tau value is 0.2. We observe that although the result of MoCov3 is slightly worse than the results of TokenCut with Dino, MoCov3 still outperforms previous state-of-the-art, indicating that TokenCut can provide similar results when used with other self-supervised Transformer architectures. Besides, the results demonstrates that a patch size of 16 provides  better results than a patch size of 8. Several insights can be found: i) TokenCut outperforms LOST for different backbones. ii) As LOST relies on a heuristic seeds expansion strategy, the performance varies significantly using different backbones. While our approach is more robust.

 \begin{table}[!ht]
	\begin{center}
	\caption{\textbf{Analysis of different backbones.} We report CorLoc for unsupervised single object discovery on VOC07, VOC12, COCO20K.}   
	
	\resizebox{0.8\columnwidth}{!}{
    \begin{tabular}{lllll}
    \toprule Method & Backbone &
    VOC07 & VOC12 & COCO20K\\
    \midrule
    LOST~\cite{simeoni2021localizing} &  ViT-S/16~\cite{dosovitskiy2020image,caron2021emerging} & 61.9 & 64.0 & 50.7 \\
    \bf TokenCut &   MoCoV3-ViT-S/16~\cite{dosovitskiy2020image,chen2021empirical} &  66.2  &  66.9 & 54.5\\ 
    \bf TokenCut &   ViT-S/16~\cite{dosovitskiy2020image,caron2021emerging} & \bf 68.8 (\textcolor{cssgreen}{$\uparrow$ \bf 6.9}) &  72.1 (\textcolor{cssgreen}{$\uparrow$ \bf 8.1}) &  58.8 (\textcolor{cssgreen}{$\uparrow$ \bf 8.1})\\ 
    \midrule 
    LOST~\cite{simeoni2021localizing}& ViT-S/8~\cite{dosovitskiy2020image,caron2021emerging} & 55.5 & 57.0 & 49.5 \\
    \bf TokenCut  & ViT-S/8~\cite{dosovitskiy2020image,caron2021emerging} &  67.3 (\textcolor{cssgreen}{$\uparrow$ \bf 11.8})    &  71.6 (\textcolor{cssgreen}{$\uparrow$ \bf 14.6}) &  \bf 60.7 (\textcolor{cssgreen}{$\uparrow$ \bf 11.2})\\ 
    \midrule
    LOST~\cite{simeoni2021localizing}& ViT-B/16~\cite{dosovitskiy2020image,caron2021emerging} & 60.1 & 63.3 & 50.0 \\
    \bf TokenCut  & ViT-B/16~\cite{dosovitskiy2020image,caron2021emerging} &  68.8 (\textcolor{cssgreen}{$\uparrow$ \bf 8.7})    &  \bf 72.4 (\textcolor{cssgreen}{$\uparrow$ \bf 9.1}) & 59.0 (\textcolor{cssgreen}{$\uparrow$ \bf 9.0})\\
    \bottomrule
    \end{tabular}}
   
	\label{tab:backbone}
		
	\end{center}

\end{table} 

We provide another an ablation study on different backbones for weakly supervised object localization. Results are shown in Tab.~\ref{tab:analysis}. The ``-S" and ``-B" designate ViT small~\cite{caron2021emerging, dosovitskiy2020image} and ViT base~\cite{caron2021emerging, dosovitskiy2020image} architecture respectively. The ``-16" and ``-8" indicate patch sizes 16 and 8 respectively. For our approach, we report results with $\tau = 0.2$, which is the same on all the datasets. Note that LOST with ViT-S/8 achieves much worse results, because the seed expansion strategy in LOST relies on the top-100 patches which are with lowest degrees. When the total number of patches is large, the proposed seed expansion strategy is not able to cover entire objects. While our approach provides more robust performance on different datasets across different backbones. 

\begin{minipage}{\linewidth}
	\begin{center}
	\captionsetup{type=table}+\caption{\textbf{Analysis of backbones for weakly supervised object localization.} We report Top-1 Cls, GT Loc and Top-1 Loc on CUB~\cite{WahCUB_200_2011} and Imagenet-1k~\cite{deng2009imagenet} datasets.}

	\resizebox{0.8\textwidth}{!}{
		\begin{tabular}{cll|lll|lll}
	    \toprule
			\multirow{2}{*}{Method}  & \multirow{2}{*}{Backbone} &  \multirow{2}{*}{$\tau$} & \multicolumn{3}{c|}{CUB [50], Acc. (\%)} & \multicolumn{3}{c}{ImageNet-1K [11], Acc. (\%)}  \\
			& & & Top-1 Cls & GT Loc & Top-1 Loc & Top-1 Cls & GT Loc & Top-1 Loc   \\
		   \midrule
		   
		LOST~\cite{simeoni2021localizing} & ViT-S/16~\cite{dosovitskiy2020image,caron2021emerging}& - & \bf 79.5 & 89.7 & 71.3 & \bf 77.0 & 60.0 & 49.0 \\
	     \textbf{TokenCut}  & ViT-S/16~\cite{dosovitskiy2020image,caron2021emerging} & 0.2 &  \bf 79.5 &  \bf 91.8 (\textcolor{cssgreen}{$\uparrow$ \bf 2.1}) & \bf 72.9 (\textcolor{cssgreen}{$\uparrow$ \bf 1.6}) & \bf 77.0	&  \bf 65.4 (\textcolor{cssgreen}{$\uparrow$ \bf 5.4}) &  \bf 53.4 (\textcolor{cssgreen}{$\uparrow$ \bf 4.4})  \\

	    \midrule
	    
	    LOST~\cite{simeoni2021localizing} & ViT-S/8~\cite{dosovitskiy2020image,caron2021emerging} & -  & \bf 82.3 & 78.0 & 64.4 & \bf 79.4 & 45.8 & 38.1 \\
	    \textbf{TokenCut}  & ViT-S/8~\cite{dosovitskiy2020image,caron2021emerging} & 0.2 & \bf 82.3 & \bf 89.9 (\textcolor{cssgreen}{$\uparrow$ \bf 11.9})  &  \bf 74.2 (\textcolor{cssgreen}{$\uparrow$ \bf 9.8})  & \bf 79.4 & \bf 66.0 (\textcolor{cssgreen}{$\uparrow$ \bf 20.2}) &  \bf 55.0 (\textcolor{cssgreen}{$\uparrow$ \bf 16.9}) \\

		\midrule
		LOST~\cite{simeoni2021localizing} & ViT-B/16~\cite{dosovitskiy2020image,caron2021emerging} & - & \bf 80.3 & \bf 90.7  & \bf 72.8 & \bf 78.3 & 58.6 & 48.3 \\
		\textbf{TokenCut}  & ViT-B/16~\cite{dosovitskiy2020image,caron2021emerging} & 0.2 & \bf 80.3 & 90.0 (\textcolor{cssred}{$\downarrow$ \bf 0.7})  & 72.5 (\textcolor{cssred}{$\downarrow$ \bf 0.3})	& \bf 78.3 &  \bf 63.2 (\textcolor{cssgreen}{$\uparrow$ \bf 4.8}) & \bf 52.3 (\textcolor{cssgreen}{$\uparrow$ \bf 4.0}) \\

		  \bottomrule
	\end{tabular}}
	\label{tab:analysis}
	\end{center}
\end{minipage}

\section{Analysis of bi-partition strategies.}

In Tab.~\ref{tab:partition}, we study different strategies to separate the nodes in our graph into two groups using the second smallest eigenvector. We consider three natural methods: mean value (Mean), Expectation-Maximisation (EM), K-means clustering (K-means). We use python sklearn library for EM and K-means algorithm implementation. For EM algorithm, we set number of iteration to 300 and each component has its own general covariance matrix. The convergence threshold is set to 1e-3. For K-means algorithm, we use ``k-means++'' for initialization. The maximum number of iterations is set to 300. The convergence threshold is set to 1e-4. The result suggests that the simple mean value as the splitting point performs well for most cases. We have also tried to search for the splitting point based on the best Ncut(A,B) value. Due to the quadratic complexity, this approach requires substantially more computations. Thus,  we finally obsolete it. 

\begin{table}[!ht]
	\begin{center}
	\caption{\textbf{Analysis of different bi-partition methods.} We report CorLoc for unsupervised single object discovery.}   
	\resizebox{0.4\columnwidth}{!}{
    \begin{tabular}{cccc}
    \toprule Bi-partition &
    VOC07 & VOC12 & COCO20K \\
    \midrule
    Mean        & \bf 68.8 & \bf 72.1 & 58.8 \\
    EM          & 63.0 & 65.7 & 59.3 \\
    K-means      & 67.5 &  69.2 &  \bf 61.6  \\
    
    \bottomrule
    \end{tabular}}
	\label{tab:partition}
	\end{center}

\end{table}

\section{Analysis of Graph edge weight}
\label{sec:backbone}

In this section, we provide an ablation study on graph edge weight defininig on equation~\ref{eqn:edge}. We have tested to directly use the similarity score as edge weights (i.e., $\edge_{ij}=S(x_i,x_j))$). However, it is not possible because there may exist negative edge values, which violates the Normalized Cut algorithm assumption. Thus, we also tried thresholding the similarity score (i.e., $\edge_{ij}=S(x_i,x_j)$ if $S(x_i,x_j) > \tau$, else $\epsilon$). We obtain 68.9\% on VOC07 dataset and 72\% on VOC12 dataset, which is similar to our reported results.

\section{Visual results for unsupervised single object discovery on VOC07 and COCO12}
\begin{figure*}[!ht]
\begin{tabular}{c@{\hskip 3pt}c@{\hskip 3pt}c@{\hskip 3pt}c@{\hskip 3pt}c@{\hskip 3pt}c}

\includegraphics[width=0.16\textwidth, height=0.14\textwidth]{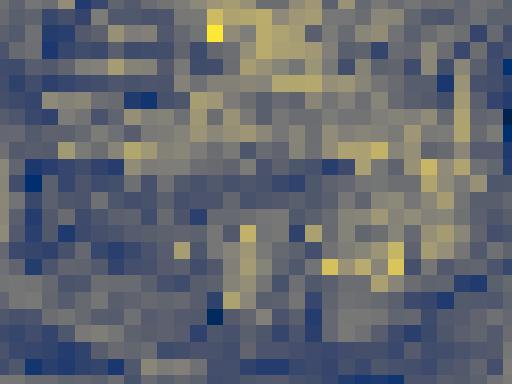} & \includegraphics[width=0.16\textwidth, height=0.14\textwidth]{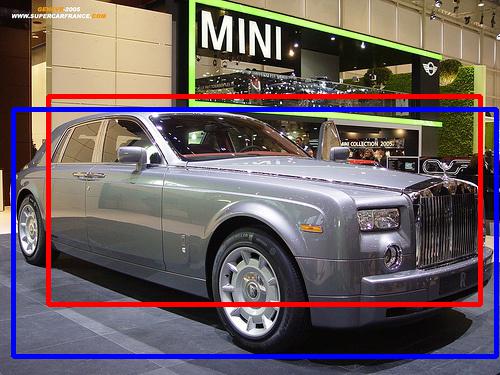} & \includegraphics[width=0.16\textwidth, height=0.14\textwidth]{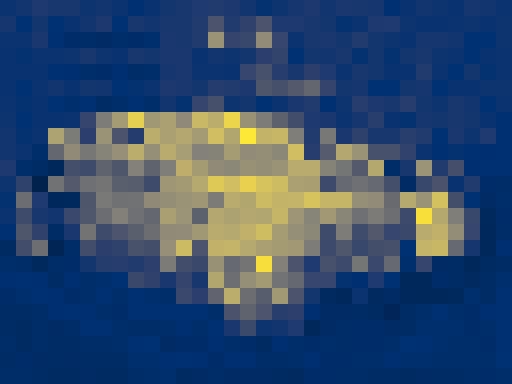} & \includegraphics[width=0.16\textwidth, height=0.14\textwidth]{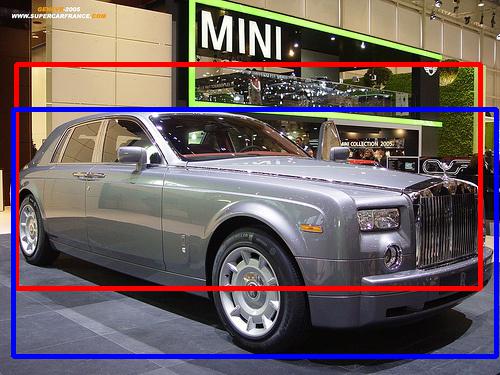} & \includegraphics[width=0.16\textwidth, height=0.14\textwidth]{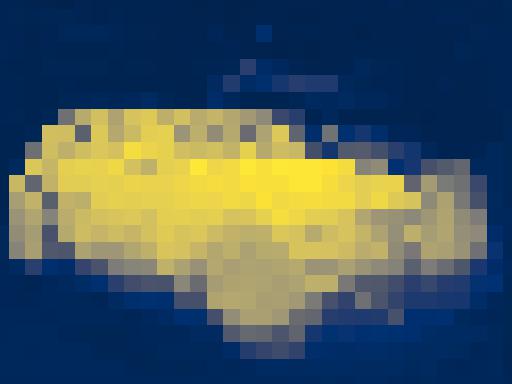} & \includegraphics[width=0.16\textwidth, height=0.14\textwidth]{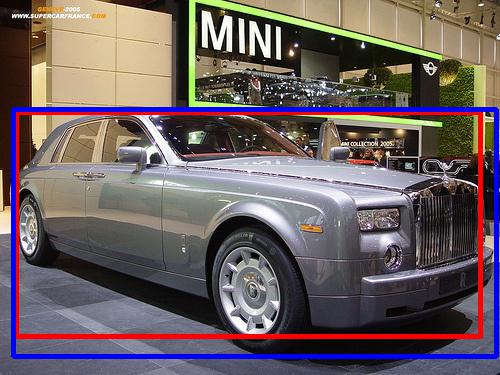}\\

\includegraphics[width=0.16\textwidth, height=0.14\textwidth]{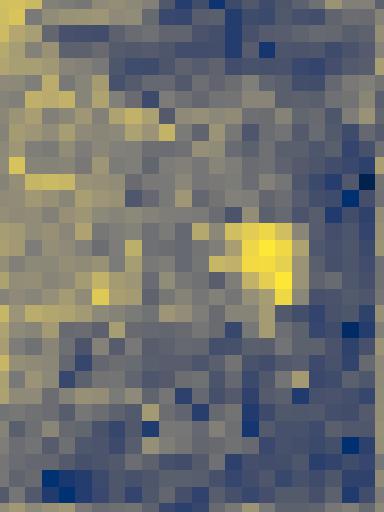} & \includegraphics[width=0.16\textwidth, height=0.14\textwidth]{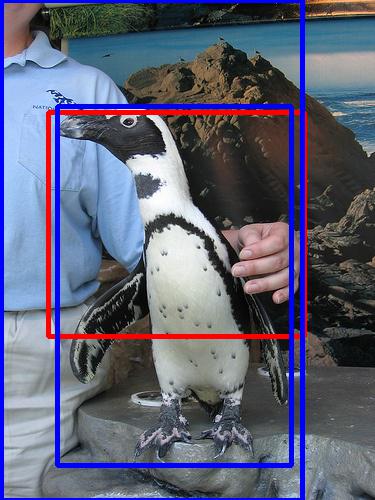} & \includegraphics[width=0.16\textwidth, height=0.14\textwidth]{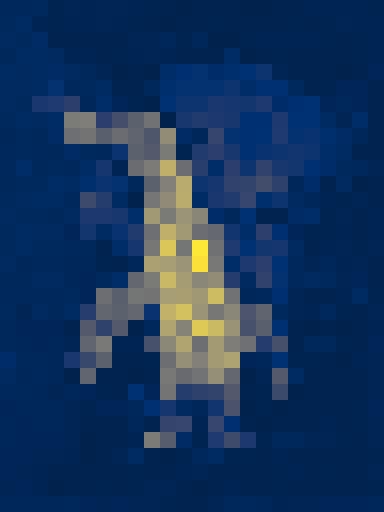} & \includegraphics[width=0.16\textwidth, height=0.14\textwidth]{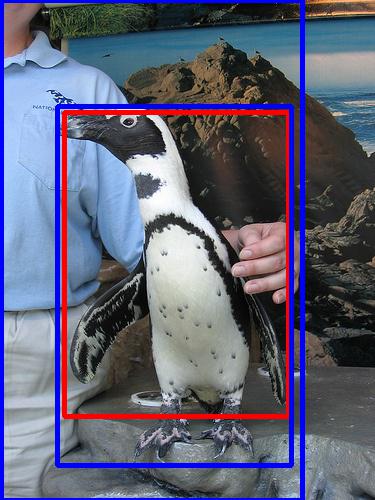} & \includegraphics[width=0.16\textwidth, height=0.14\textwidth]{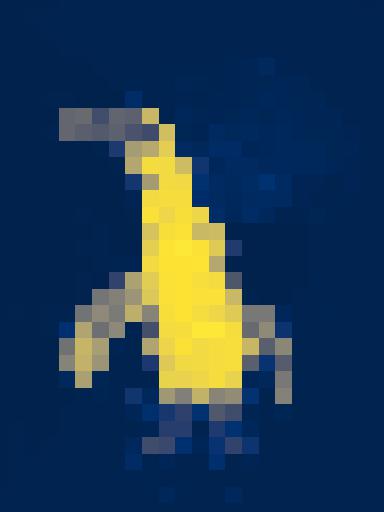} & \includegraphics[width=0.16\textwidth, height=0.14\textwidth]{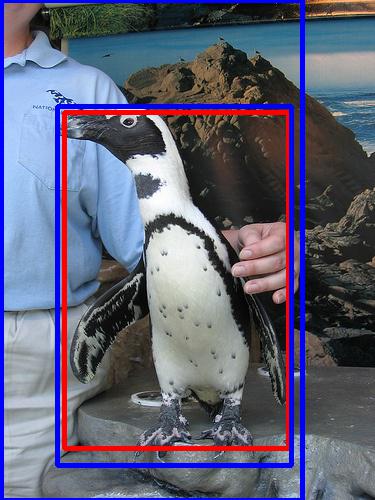}\\

\includegraphics[width=0.16\textwidth, height=0.14\textwidth]{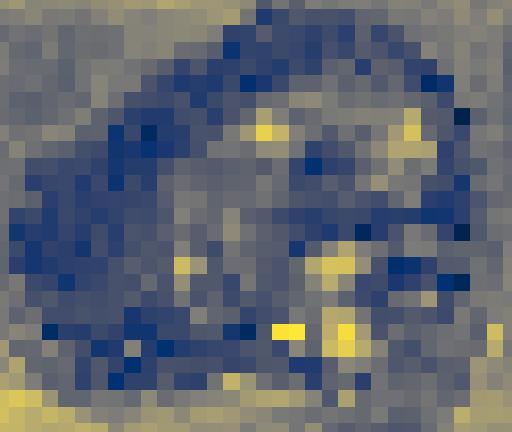} & \includegraphics[width=0.16\textwidth, height=0.14\textwidth]{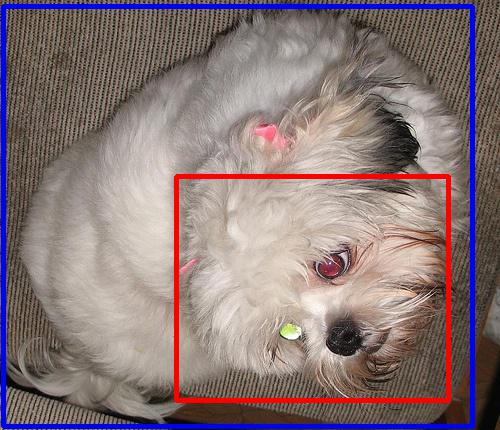} & \includegraphics[width=0.16\textwidth, height=0.14\textwidth]{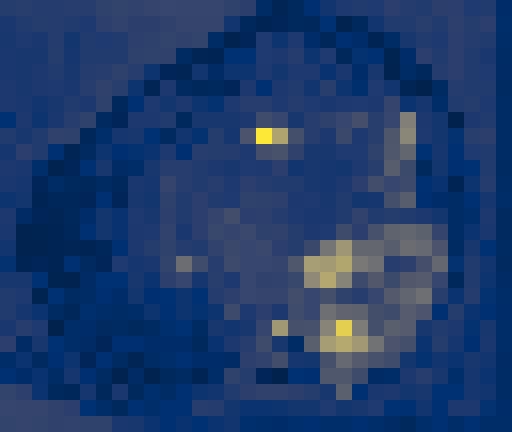} & \includegraphics[width=0.16\textwidth, height=0.14\textwidth]{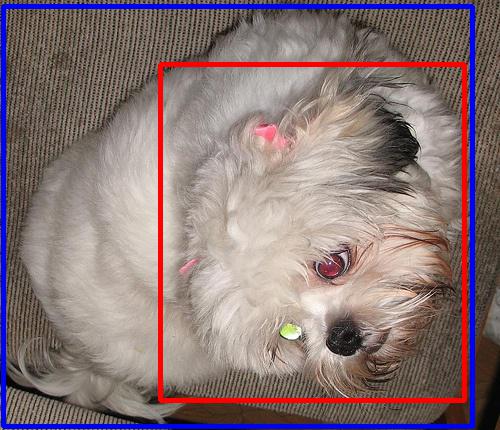} & \includegraphics[width=0.16\textwidth, height=0.14\textwidth]{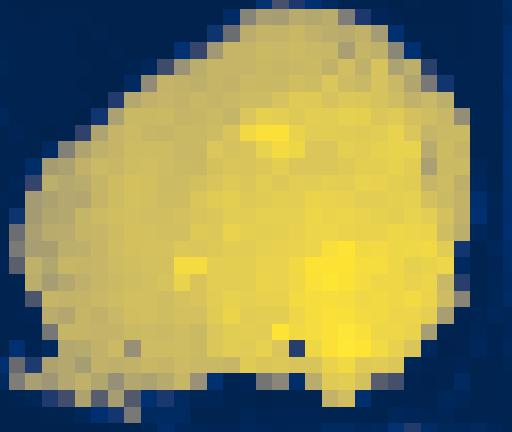} & \includegraphics[width=0.16\textwidth, height=0.14\textwidth]{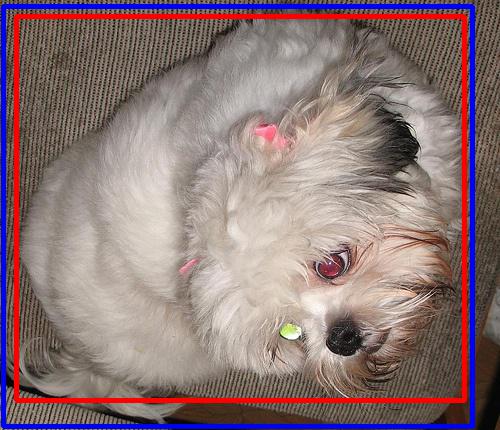}\\

\includegraphics[width=0.16\textwidth, height=0.14\textwidth]{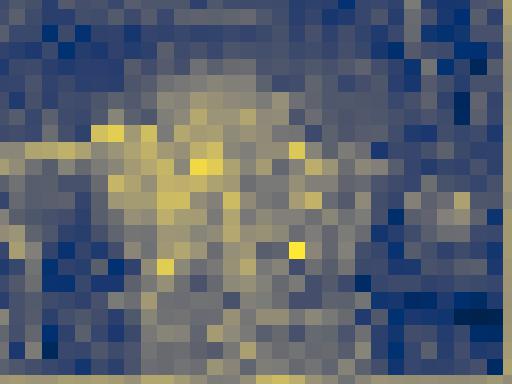} & \includegraphics[width=0.16\textwidth, height=0.14\textwidth]{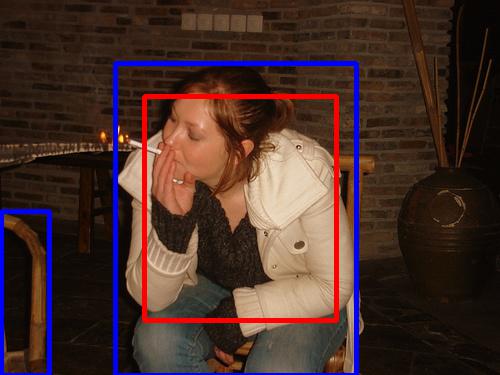} & \includegraphics[width=0.16\textwidth, height=0.14\textwidth]{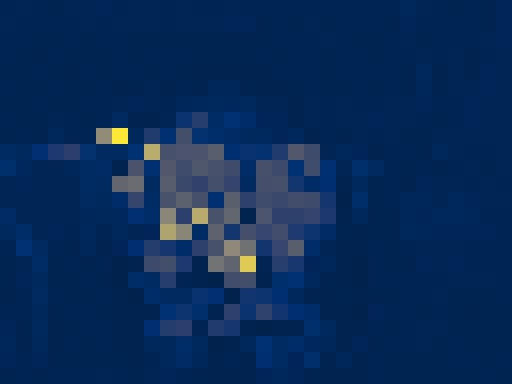} & \includegraphics[width=0.16\textwidth, height=0.14\textwidth]{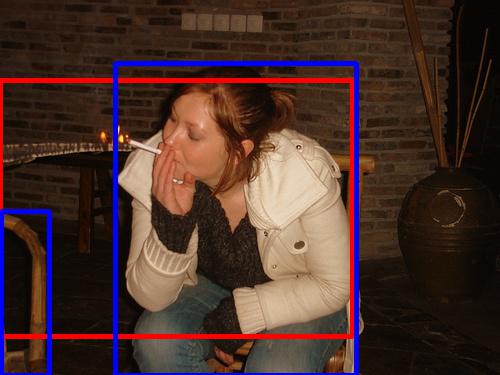} & \includegraphics[width=0.16\textwidth, height=0.14\textwidth]{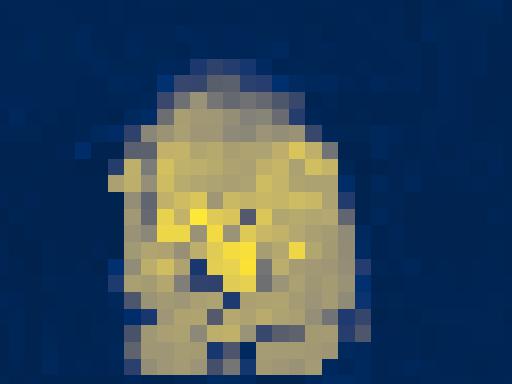} & \includegraphics[width=0.16\textwidth, height=0.14\textwidth]{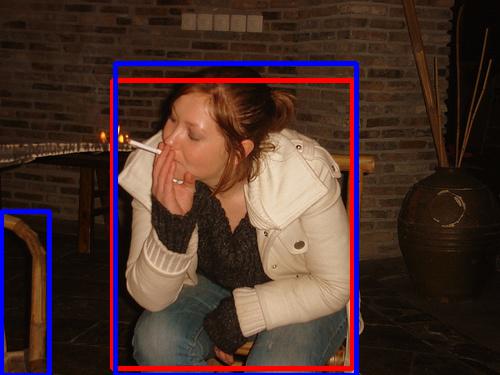}\\

\includegraphics[width=0.16\textwidth, height=0.14\textwidth]{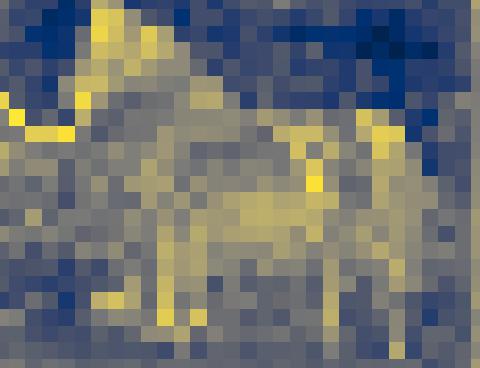} & \includegraphics[width=0.16\textwidth, height=0.14\textwidth]{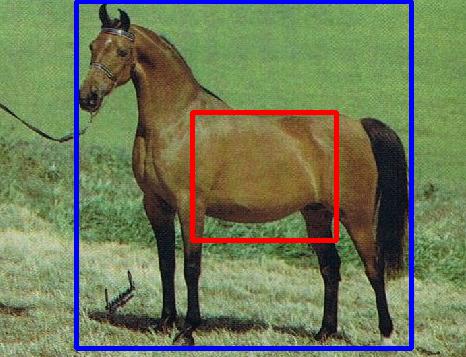} & \includegraphics[width=0.16\textwidth, height=0.14\textwidth]{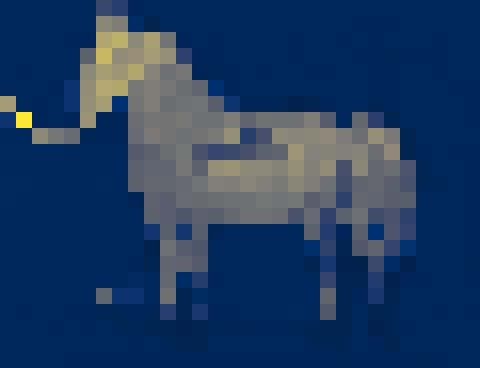} & \includegraphics[width=0.16\textwidth, height=0.14\textwidth]{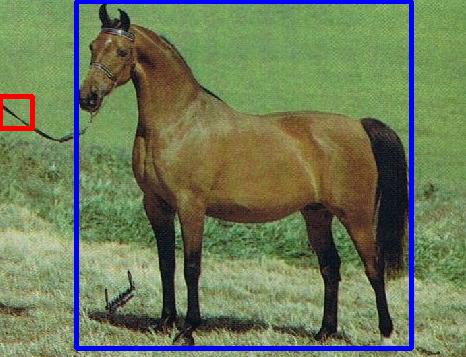} & \includegraphics[width=0.16\textwidth, height=0.14\textwidth]{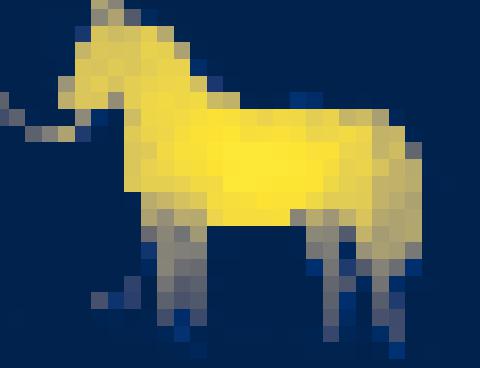} & \includegraphics[width=0.16\textwidth, height=0.14\textwidth]{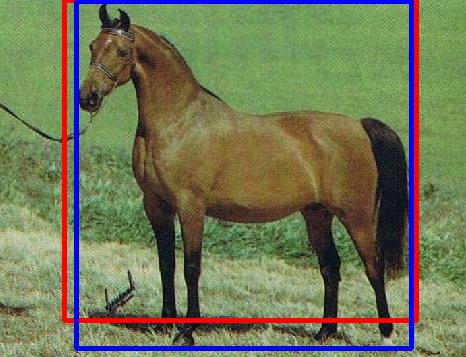}\\

\makecell{(a) DINO CLS \\ Token Attention}  & \makecell{(b) DINO \\ Detection} & \makecell{(c) LOST Inverse \\Degree Attention} & \makecell{(d) LOST \\ Detection} & \makecell{(e) Our Eigen \\Attention} & \makecell{(f) Our \\Detection} \\
\end{tabular}
\caption{\textbf{Visual results of unsupervised single object discovery on VOC07~\cite{pascal-voc-2007}} In (a), we show the attention of the CLS token in DINO~\cite{caron2021emerging} which is used for detection (b). LOST~\cite{simeoni2021localizing} is mainly relied on the map of inverse degrees (c) to perform detection (d). For our approach, we illustrate the eigenvector in (e) and our detection in (f). \textcolor{blue}{Blue} and \textcolor{red}{Red} bounding boxes indicate the ground-truth and the predicted bounding boxes respectively.}
\label{fig:voc07}
\end{figure*}

We show visual results for unsupervised single object discovery on VOC07~\cite{pascal-voc-2007} and COCO12~\cite{lin2014microsoft, vo2020toward}, which are illustrated in Fig.~\ref{fig:voc07} and Fig.~\ref{fig:coco} respectively.

For each dataset, we compare both attention maps and bounding box predictions among DINO~\cite{caron2021emerging}, LOST~\cite{simeoni2021localizing} and  TokenCut. The attention map for DINO is extracted from the CLS token attention map of the last layer of key features. The attention map for LOST is the inverse degree map used in LOST for detection. The TokenCut attention map is the second smallest eigenvector of Equation~\ref{eqn:ncut_final}. These results show that TokenCut provides clearly better segmentation of the object.

\begin{center}
\begin{tabular}{c@{\hskip 3pt}c@{\hskip 3pt}c@{\hskip 3pt}c@{\hskip 3pt}c@{\hskip 3pt}c}

\includegraphics[width=0.16\textwidth, height=0.14\textwidth]{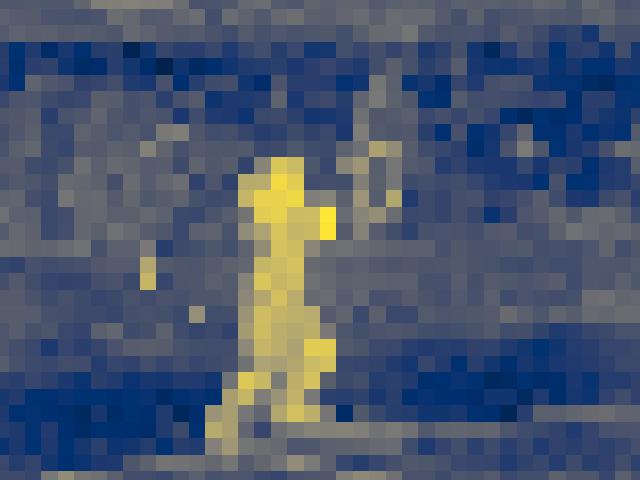} & \includegraphics[width=0.16\textwidth, height=0.14\textwidth]{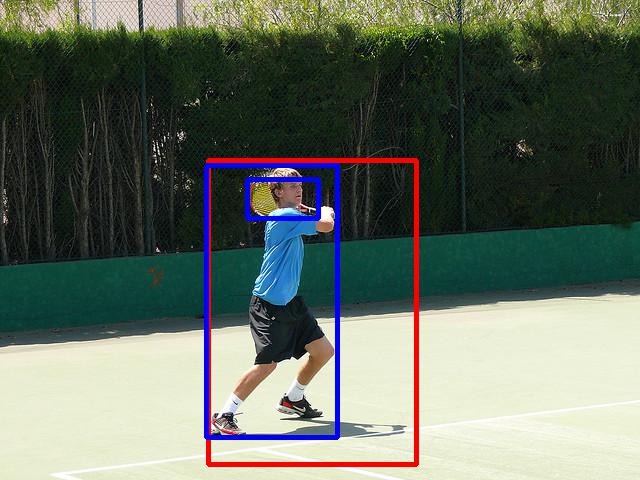} & \includegraphics[width=0.16\textwidth, height=0.14\textwidth]{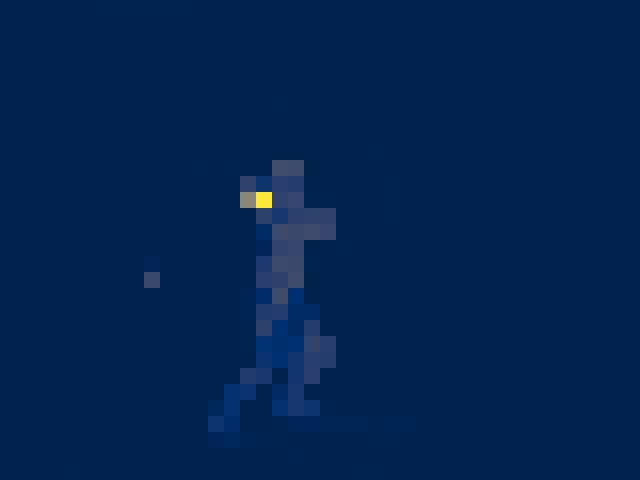} & \includegraphics[width=0.16\textwidth, height=0.14\textwidth]{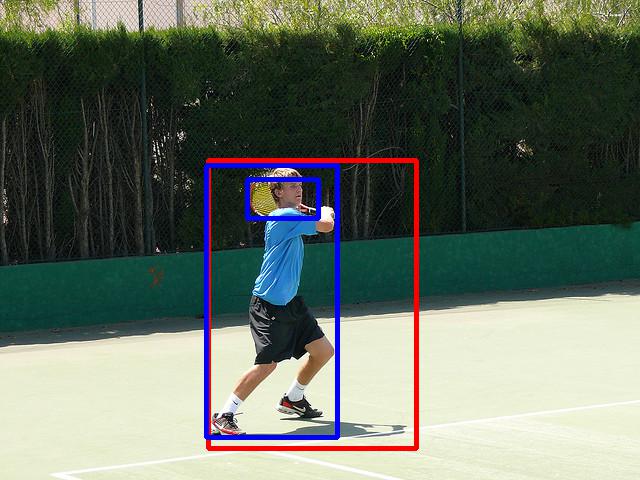} & \includegraphics[width=0.16\textwidth, height=0.14\textwidth]{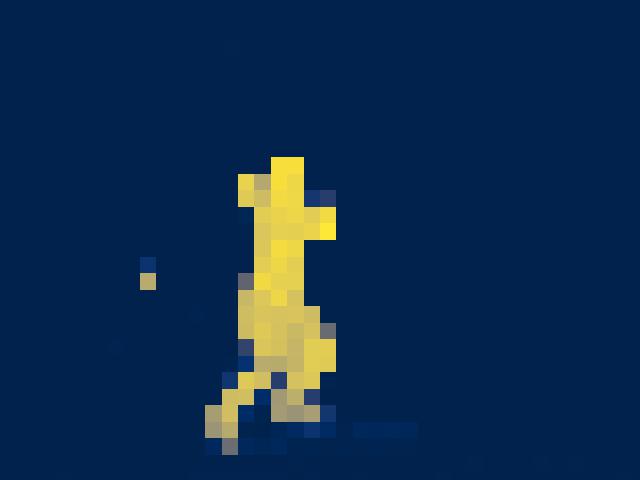} & \includegraphics[width=0.16\textwidth, height=0.14\textwidth]{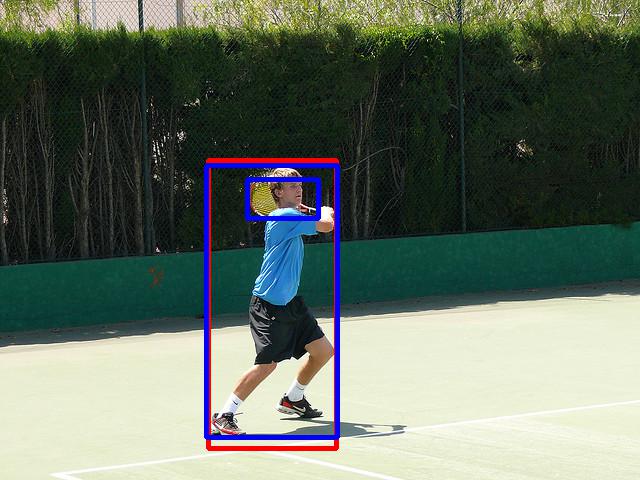}\\

\includegraphics[width=0.16\textwidth, height=0.14\textwidth]{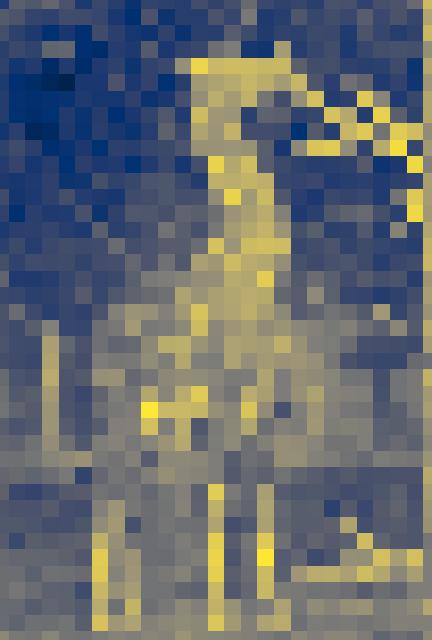} & \includegraphics[width=0.16\textwidth, height=0.14\textwidth]{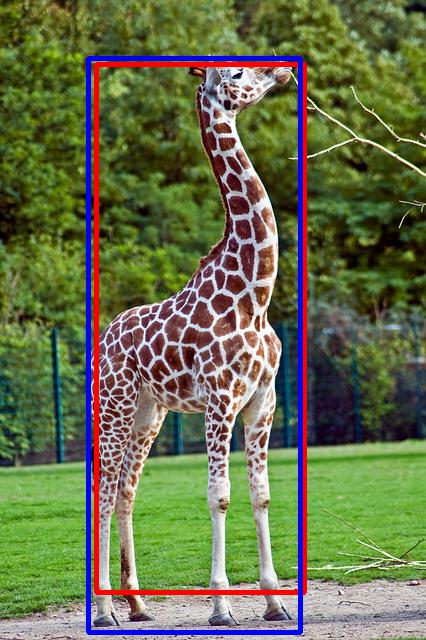} & \includegraphics[width=0.16\textwidth, height=0.14\textwidth]{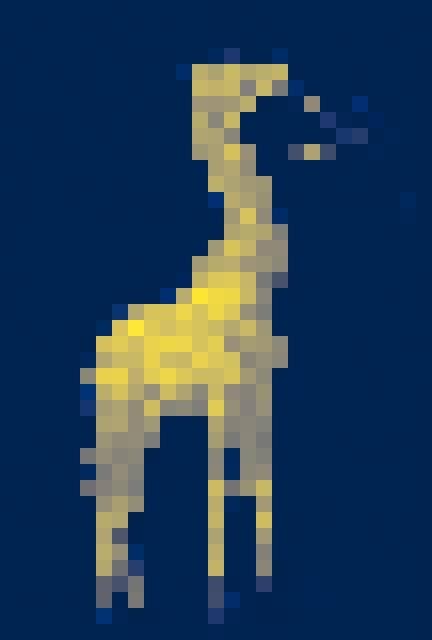} & \includegraphics[width=0.16\textwidth, height=0.14\textwidth]{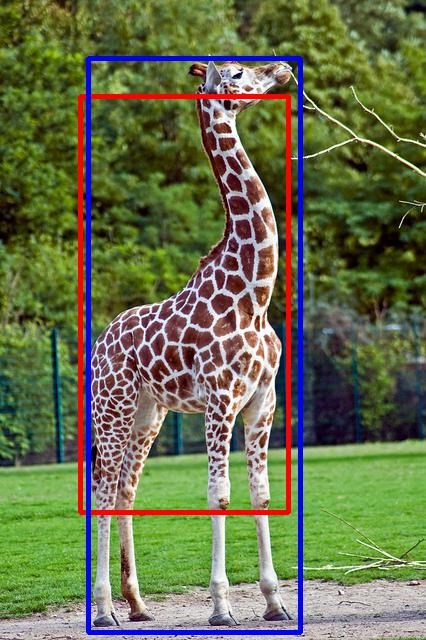} & \includegraphics[width=0.16\textwidth, height=0.14\textwidth]{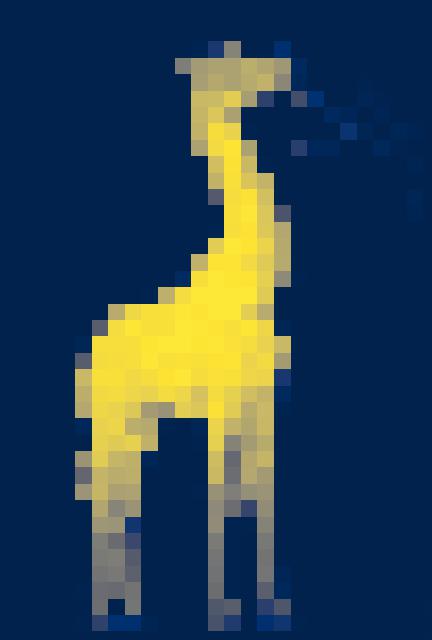} & \includegraphics[width=0.16\textwidth, height=0.14\textwidth]{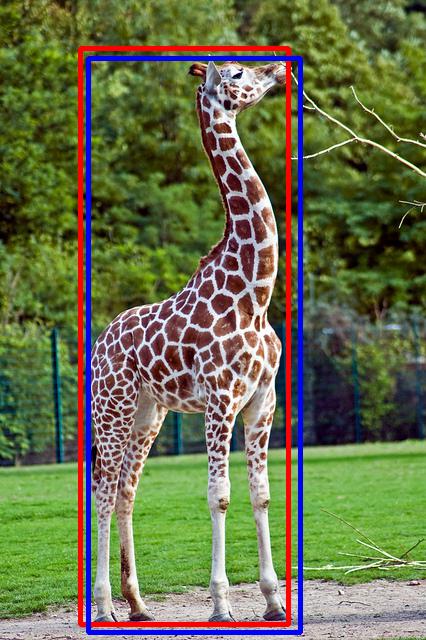}\\

\includegraphics[width=0.16\textwidth, height=0.14\textwidth]{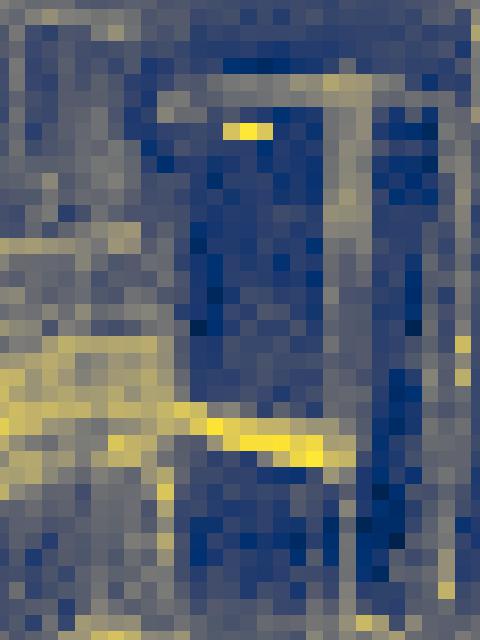} & \includegraphics[width=0.16\textwidth, height=0.14\textwidth]{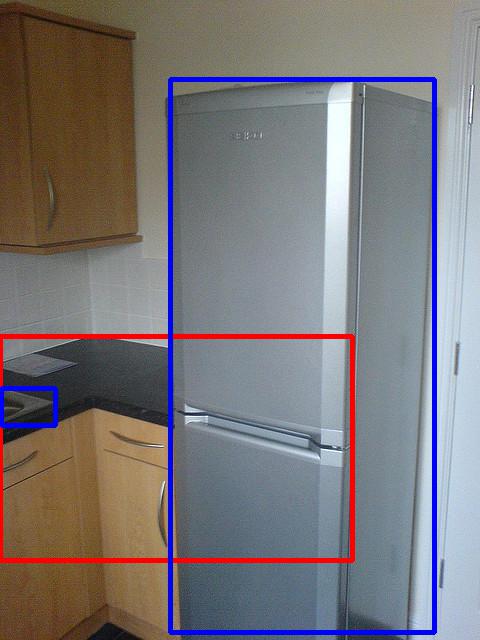} & \includegraphics[width=0.16\textwidth, height=0.14\textwidth]{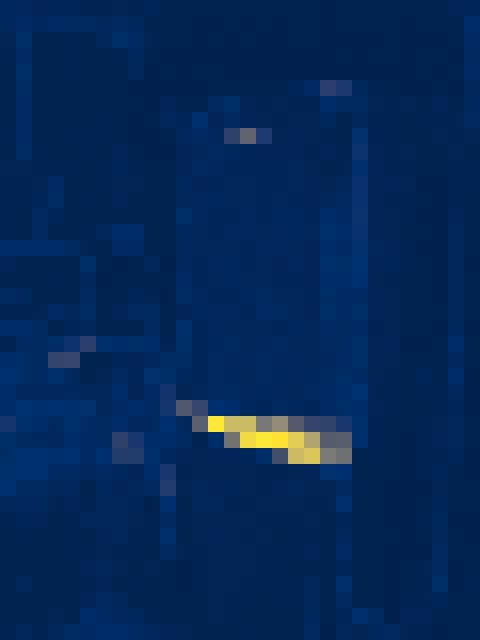} & \includegraphics[width=0.16\textwidth, height=0.14\textwidth]{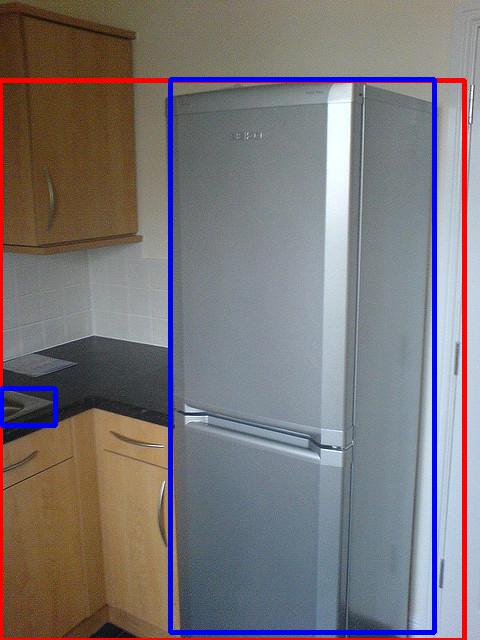} & \includegraphics[width=0.16\textwidth, height=0.14\textwidth]{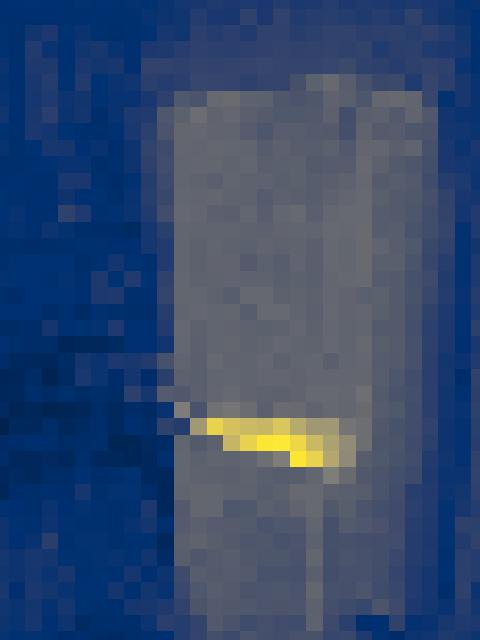} & \includegraphics[width=0.16\textwidth, height=0.14\textwidth]{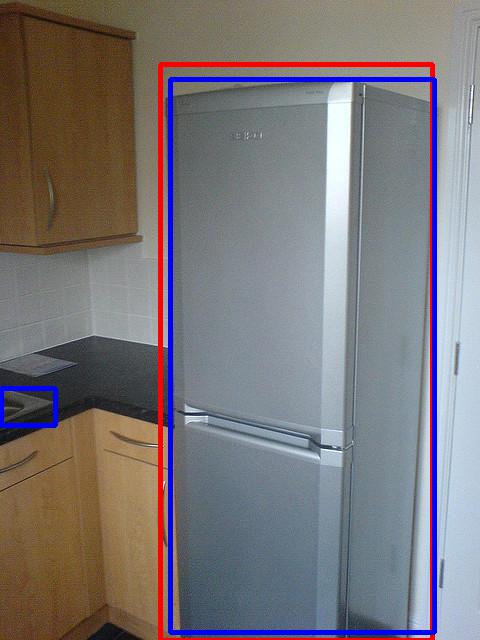}\\

\includegraphics[width=0.16\textwidth, height=0.14\textwidth]{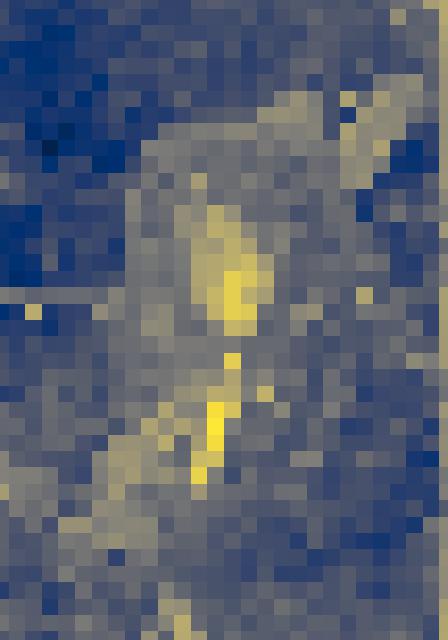} & \includegraphics[width=0.16\textwidth, height=0.14\textwidth]{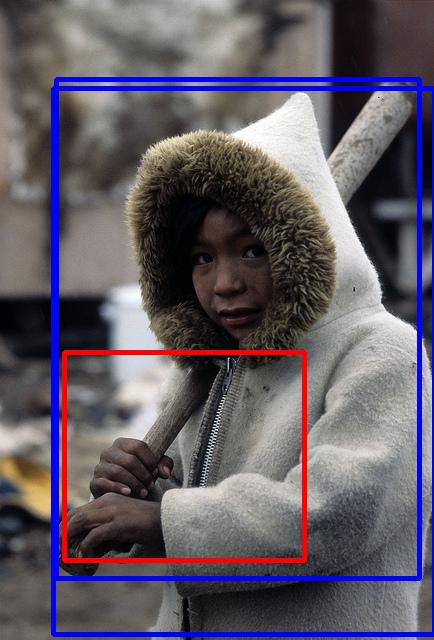} & \includegraphics[width=0.16\textwidth, height=0.14\textwidth]{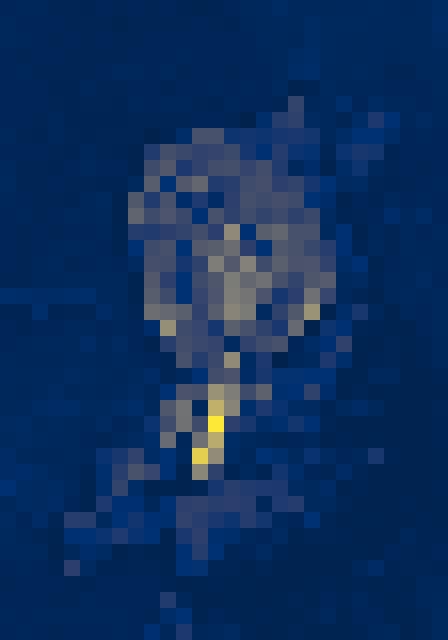} & \includegraphics[width=0.16\textwidth, height=0.14\textwidth]{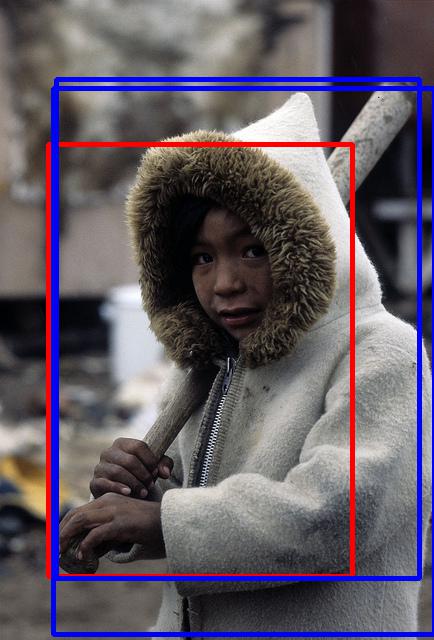} & \includegraphics[width=0.16\textwidth, height=0.14\textwidth]{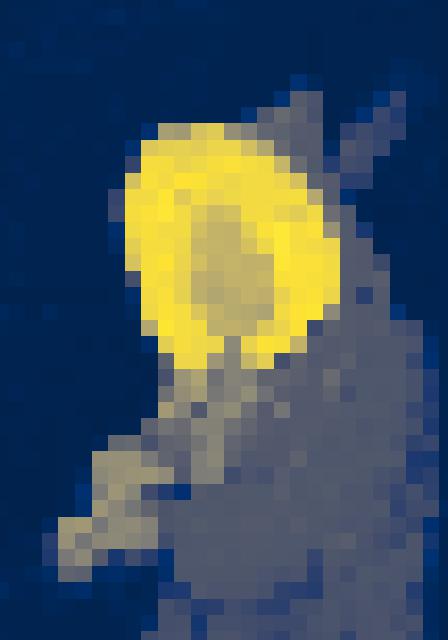} & \includegraphics[width=0.16\textwidth, height=0.14\textwidth]{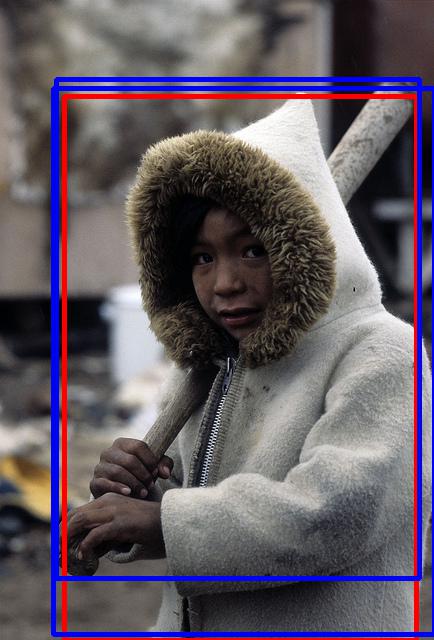}\\

\includegraphics[width=0.16\textwidth, height=0.14\textwidth]{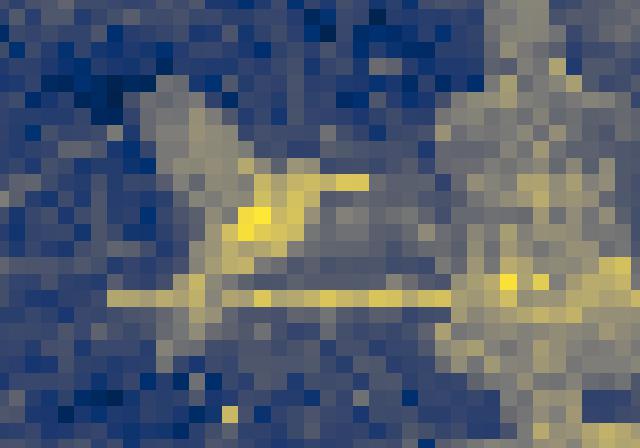} & \includegraphics[width=0.16\textwidth, height=0.14\textwidth]{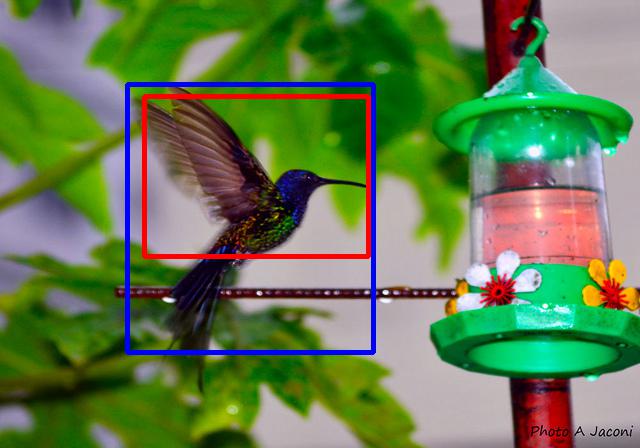} & \includegraphics[width=0.16\textwidth, height=0.14\textwidth]{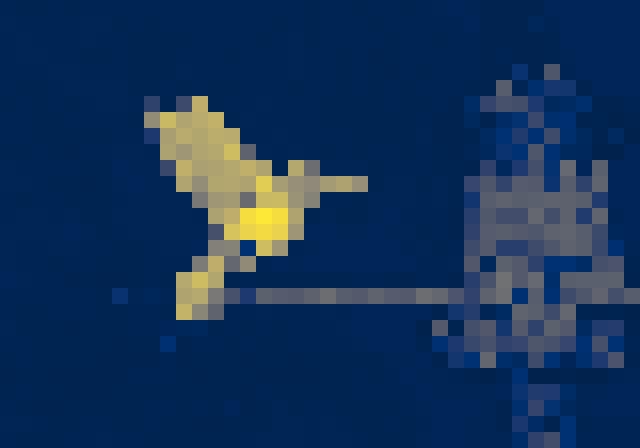} & \includegraphics[width=0.16\textwidth, height=0.14\textwidth]{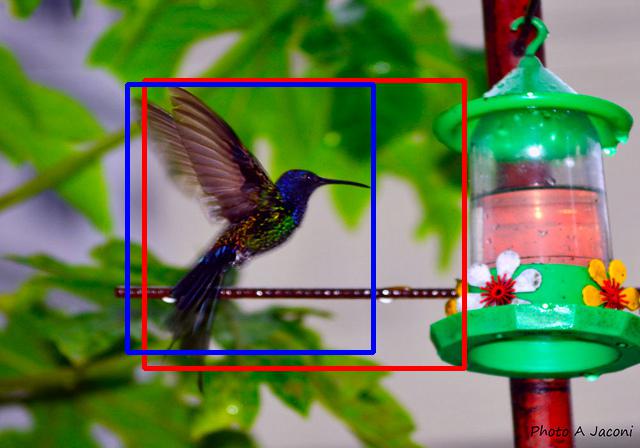} & \includegraphics[width=0.16\textwidth, height=0.14\textwidth]{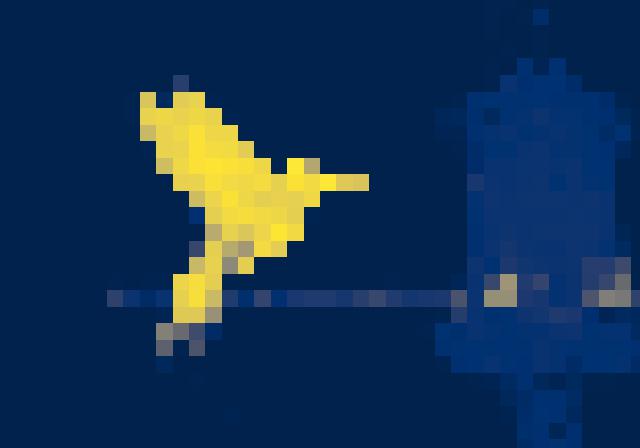} & \includegraphics[width=0.16\textwidth, height=0.14\textwidth]{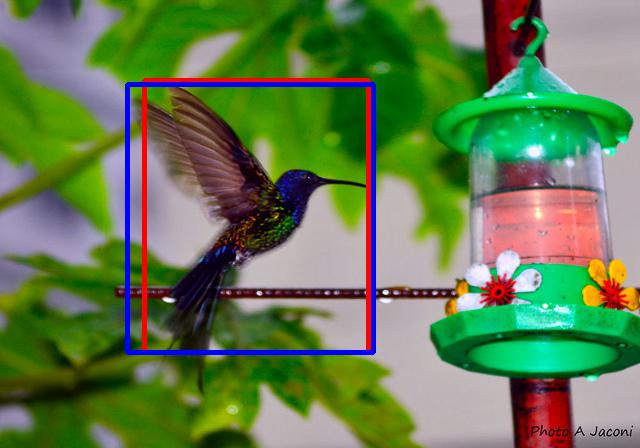}\\

\includegraphics[width=0.16\textwidth, height=0.14\textwidth]{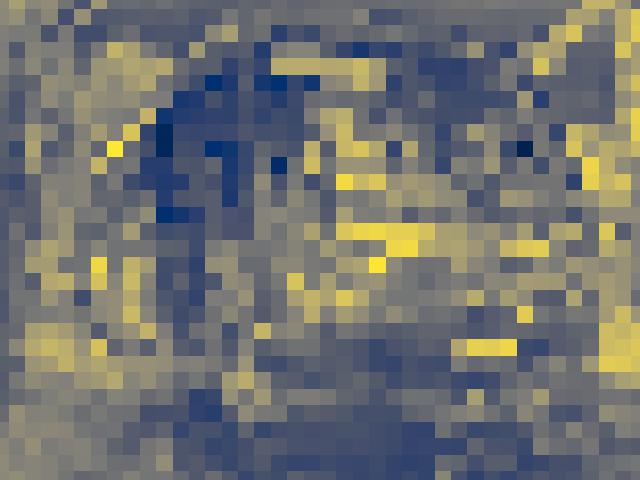} & \includegraphics[width=0.16\textwidth, height=0.14\textwidth]{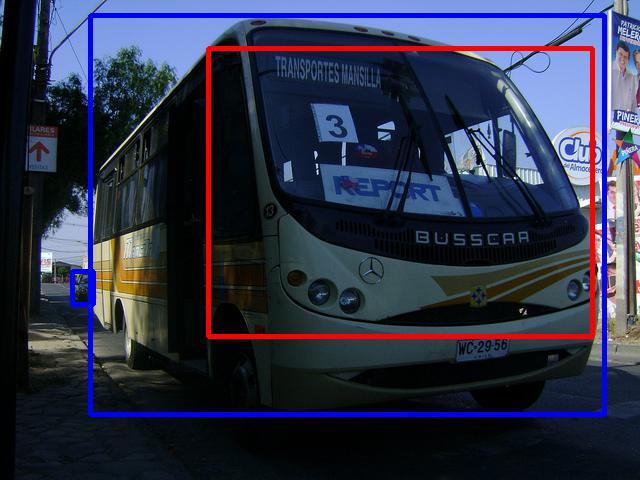} & \includegraphics[width=0.16\textwidth, height=0.14\textwidth]{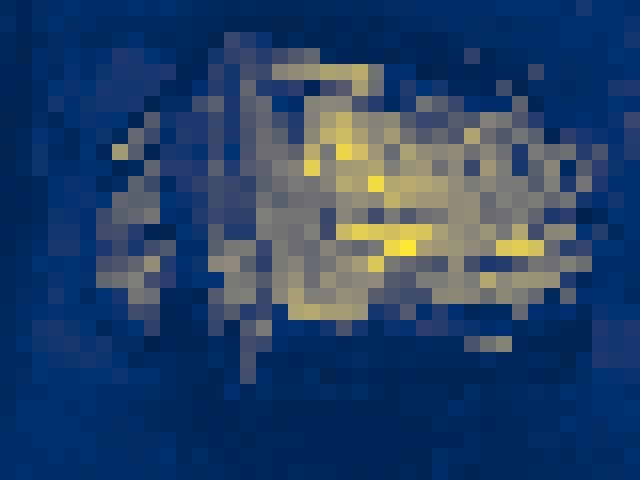} & \includegraphics[width=0.16\textwidth, height=0.14\textwidth]{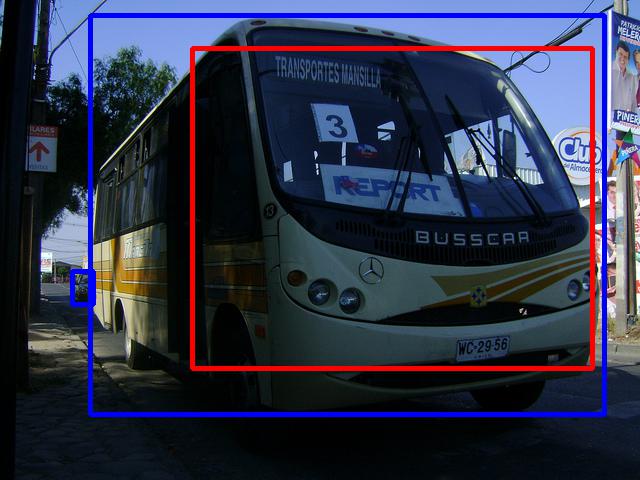} & \includegraphics[width=0.16\textwidth, height=0.14\textwidth]{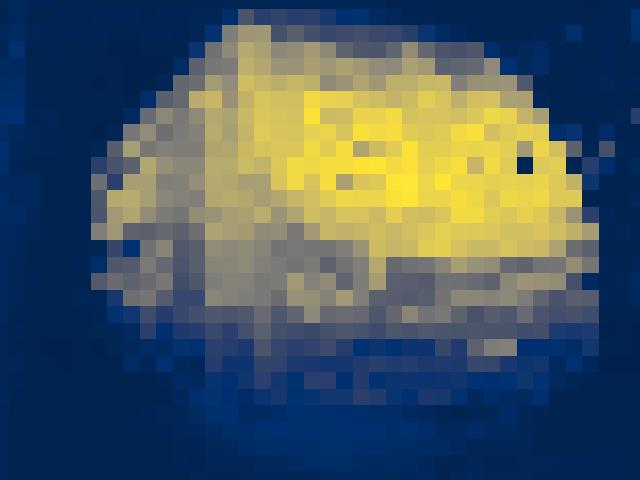} & \includegraphics[width=0.16\textwidth, height=0.14\textwidth]{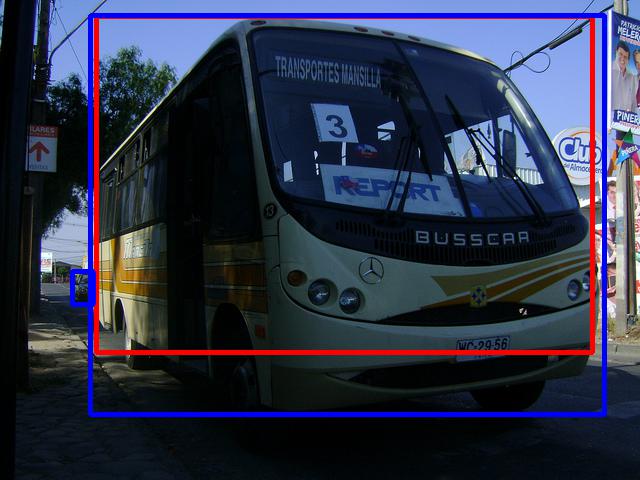}\\

\makecell{(a) DINO CLS \\ Token Attention}  & \makecell{(b) DINO \\ Detection} & \makecell{(c) LOST Inverse \\Degree Attention} & \makecell{(d) LOST \\ Detection} & \makecell{(e) Our Eigen \\Attention} & \makecell{(f) Our \\Detection} \\
\end{tabular}
\captionsetup{type=figure}+\caption{\textbf{Visual results of unsupervised single object discovery on COCO20K~\cite{lin2014microsoft, vo2020toward}.} In (a), we show  the attention of the CLS token in DINO~\cite{caron2021emerging} used for detection (b). LOST~\cite{simeoni2021localizing}  mainly relies on the map of inverse degrees (c) to perform detection (d). For TokenCut, we illustrate the eigenvector in (e) and the detection in (f). \textcolor{blue}{Blue} and \textcolor{red}{Red} bounding boxes indicate the ground-truth and the predicted bounding boxes respectively.}
\label{fig:coco}
\end{center}

\section{Fine-tuning self-supervised transformers}  
For weakly supervised object localization, we use a pre-trained DINO  model as our backbone and learn a linear classifier on the training set where we only have access to the class labels. We freeze the backbone weights and fine-tune a linear classifier, as shown in Tab.~\ref{tab:wsd}. For CUB, We train with a SGD optimizer for 1000 epochs and set the batch size to 256 per GPU, distributed over 4 GPUs. The learning rate is linearly warmed during the first 50 epochs, then follows a cosine learning rate scheduler. We decay the learning rate from $\frac{\text{batch~size}}{256} \times$5e-4 to 1e-6. The weight decay is set to 0.005. For ImageNet-1K, we use the models released by DINO. Other training setups and details can be found in the supplementary material.

\section{Visual results for weakly supervised object localizatio on CUB and Imagenet-1k}
\label{sec:wsod}

We present visual results for weakly supervised object localization on CUB~\cite{WahCUB_200_2011} and Imagenet-1k~\cite{deng2009imagenet} in Fig.~\ref{fig:cub} and Fig.~\ref{fig:imagenet} respectively.

For each dataset, we compare the attention map and bounding box prediction with LOST~\cite{simeoni2021localizing} and our approach. The eigenvector of  TokenCut provides better segmentation on objects and leads to better detection results.

\begin{minipage}{\linewidth}
\begin{center}
\resizebox{\linewidth}{!}{ 
\begin{tabular}{c@{\hskip 3pt}c@{\hskip 3pt}c@{\hskip 3pt}c@{\hskip 3pt}c@{\hskip 3pt}c@{\hskip 3pt}c}
\rotatebox{90}{\makecell{(a)~LOST Inverse \\ Degree Attention}}& 
\includegraphics[width=0.16\textwidth, height=0.14\textwidth]{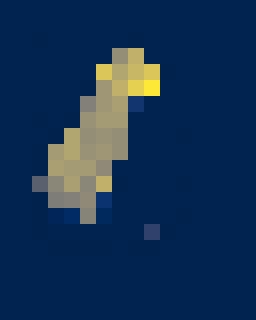} &
\includegraphics[width=0.16\textwidth, height=0.14\textwidth]{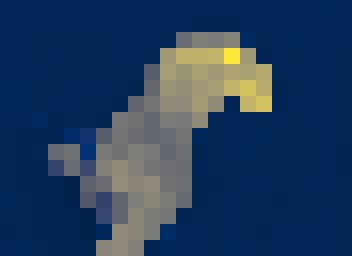} &
\includegraphics[width=0.16\textwidth, height=0.14\textwidth]{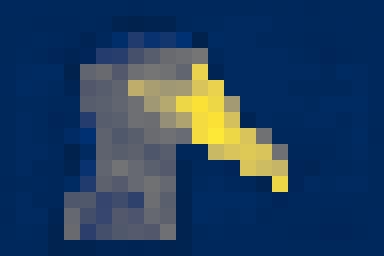} & \includegraphics[width=0.16\textwidth, height=0.14\textwidth]{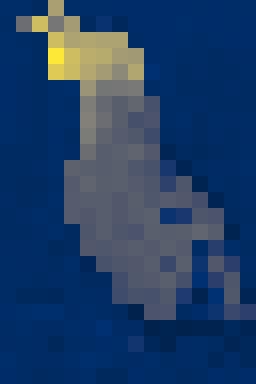}  & \includegraphics[width=0.16\textwidth, height=0.14\textwidth]{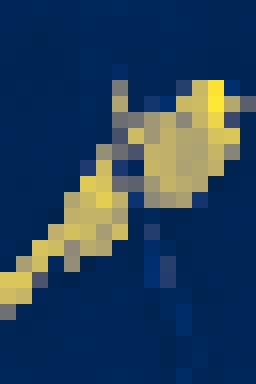}  & 
\includegraphics[width=0.16\textwidth, height=0.14\textwidth]{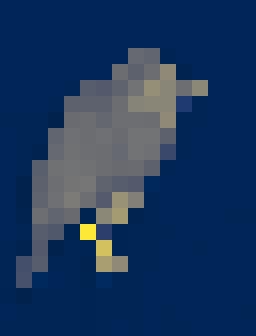} \\

\rotatebox{90}{\makecell{(b)~LOST \\Detection}} &  
\includegraphics[width=0.16\textwidth, height=0.14\textwidth]{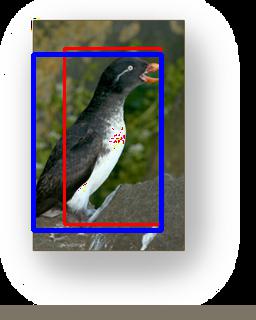} &
\includegraphics[width=0.16\textwidth, height=0.14\textwidth]{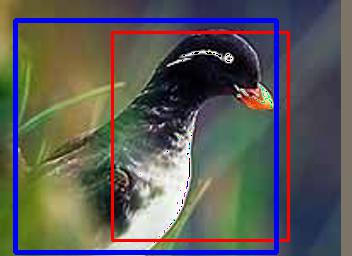} &
\includegraphics[width=0.16\textwidth, height=0.14\textwidth]{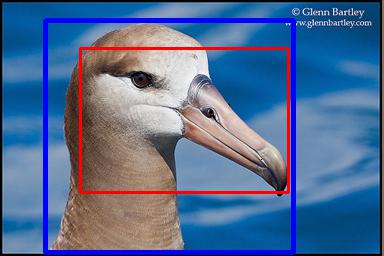} & 
\includegraphics[width=0.16\textwidth, height=0.14\textwidth]{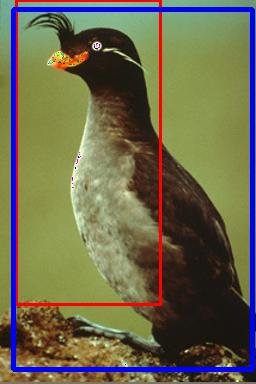} & 
\includegraphics[width=0.16\textwidth, height=0.14\textwidth]{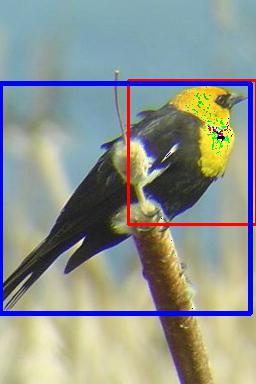} & 
\includegraphics[width=0.16\textwidth, height=0.14\textwidth]{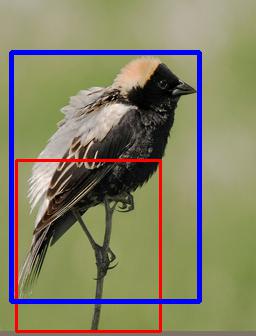}  \\

\rotatebox{90}{\makecell{(c)~Ours Eigen \\Attention}} & 
\includegraphics[width=0.16\textwidth, height=0.14\textwidth]{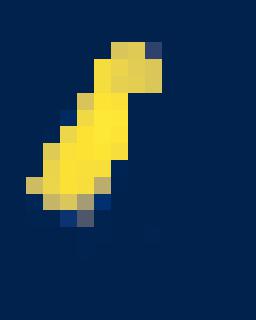} & 
\includegraphics[width=0.16\textwidth, height=0.14\textwidth]{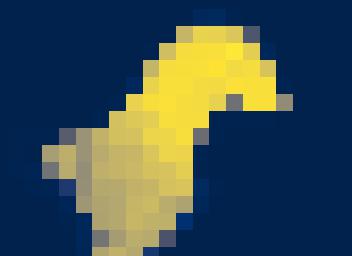} & 
\includegraphics[width=0.16\textwidth, height=0.14\textwidth]{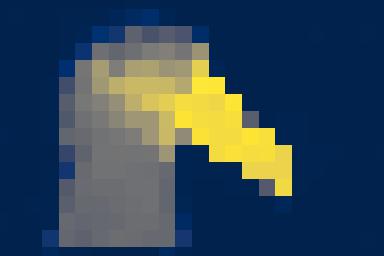} & 
\includegraphics[width=0.16\textwidth, height=0.14\textwidth]{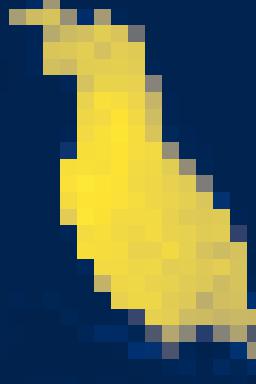} & 
\includegraphics[width=0.16\textwidth, height=0.14\textwidth]{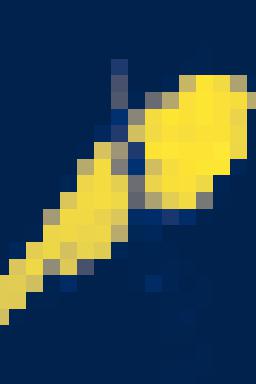} & 
\includegraphics[width=0.16\textwidth, height=0.14\textwidth]{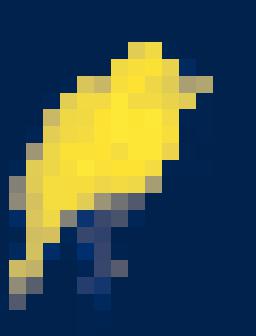} \\

\rotatebox{90}{\makecell{(d)~Ours \\Detection}} & 
\includegraphics[width=0.16\textwidth, height=0.14\textwidth]{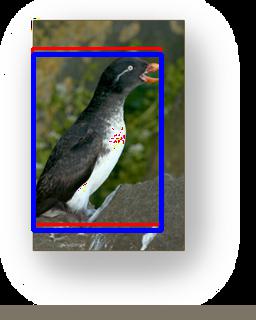} &
\includegraphics[width=0.16\textwidth, height=0.14\textwidth]{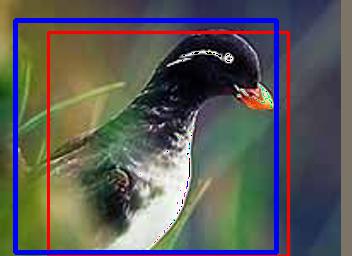} &
\includegraphics[width=0.16\textwidth, height=0.14\textwidth]{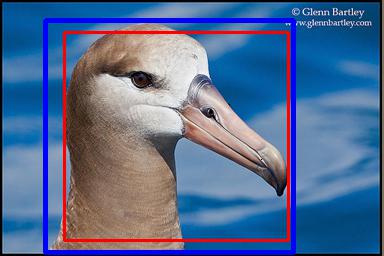} &
\includegraphics[width=0.16\textwidth, height=0.14\textwidth]{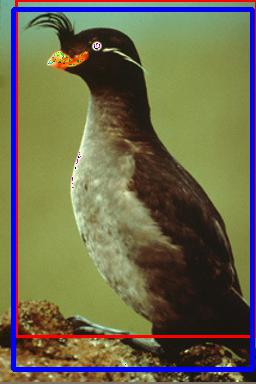} & 
\includegraphics[width=0.16\textwidth, height=0.14\textwidth]{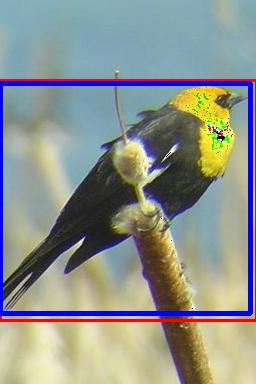} & 
\includegraphics[width=0.16\textwidth, height=0.14\textwidth]{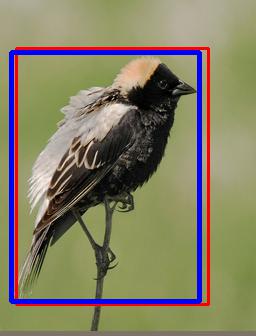} \\

\end{tabular}}

\captionsetup{type=figure} \caption{\textbf{Visual results for weakly supervised object localization on CUB~\cite{WahCUB_200_2011}.} In (a), we show the map of inverse degrees used to perform detection with LOST (b)~\cite{simeoni2021localizing}. For TokenCut, we illustrate the eigenvector in (c) used for detection in (d). \textcolor{blue}{Blue} and \textcolor{red}{Red} bounding boxes indicate the ground-truth and the predicted bounding boxes respectively.}
\label{fig:cub}
\end{center}
\end{minipage}

\begin{minipage}{\linewidth}
\begin{center}
\resizebox{\linewidth}{!}{ 
\begin{tabular}{c@{\hskip 3pt}c@{\hskip 3pt}c@{\hskip 3pt}c@{\hskip 3pt}c@{\hskip 3pt}c@{\hskip 3pt}c}
\rotatebox{90}{\makecell{(a)~LOST Inverse \\ Degree Attention}}& 
\includegraphics[width=0.16\textwidth, height=0.14\textwidth]{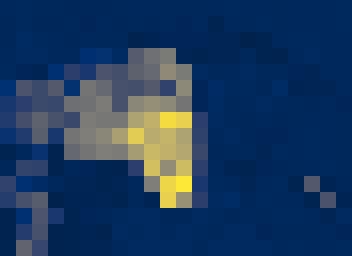} &
\includegraphics[width=0.16\textwidth, height=0.14\textwidth]{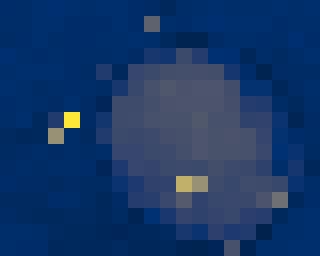} &
\includegraphics[width=0.16\textwidth, height=0.14\textwidth]{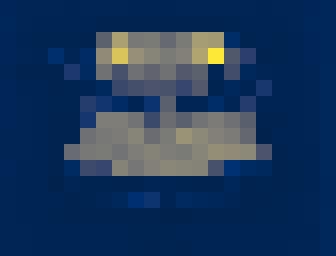} &
\includegraphics[width=0.16\textwidth, height=0.14\textwidth]{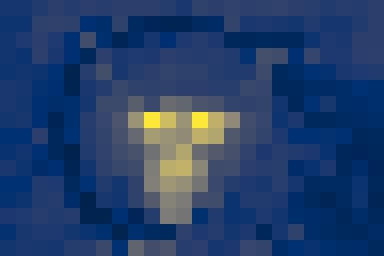} & \includegraphics[width=0.16\textwidth, height=0.14\textwidth]{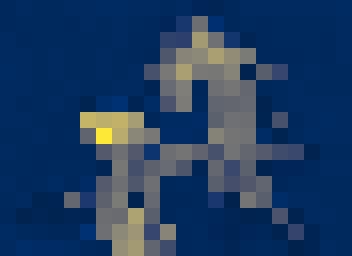} & 
\includegraphics[width=0.16\textwidth, height=0.14\textwidth]{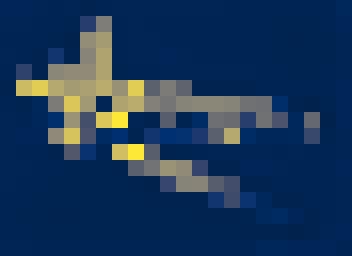} \\
\rotatebox{90}{\makecell{(b)~LOST \\Detection}} &  \includegraphics[width=0.16\textwidth, height=0.14\textwidth]{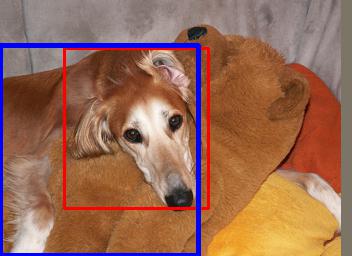} &
\includegraphics[width=0.16\textwidth, height=0.14\textwidth]{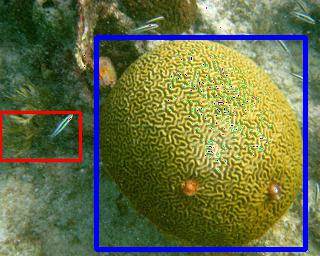} & 
\includegraphics[width=0.16\textwidth, height=0.14\textwidth]{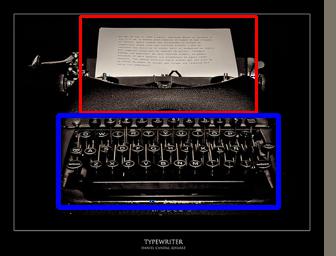} & 
\includegraphics[width=0.16\textwidth, height=0.14\textwidth]{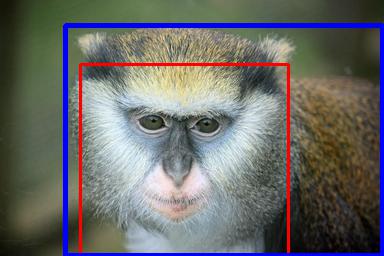} & 
\includegraphics[width=0.16\textwidth, height=0.14\textwidth]{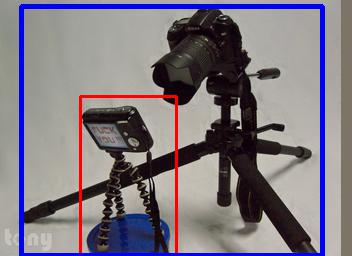} & 
\includegraphics[width=0.16\textwidth, height=0.14\textwidth]{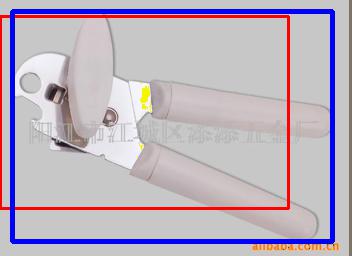} \\

\rotatebox{90}{\makecell{(c)~Ours Eigen \\Attention}} & \includegraphics[width=0.16\textwidth, height=0.14\textwidth]{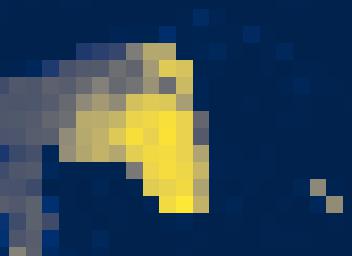} & 
\includegraphics[width=0.16\textwidth, height=0.14\textwidth]{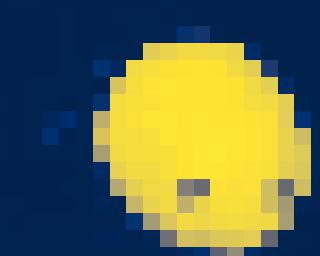} & 
\includegraphics[width=0.16\textwidth, height=0.14\textwidth]{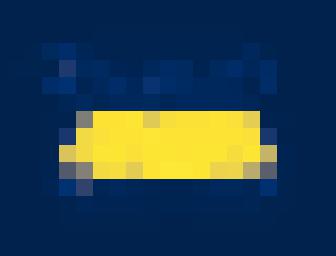} & 
\includegraphics[width=0.16\textwidth, height=0.14\textwidth]{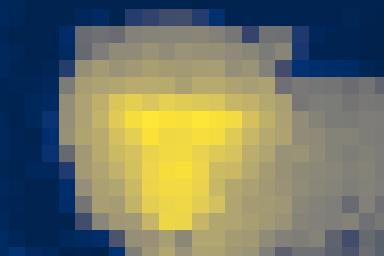} & 
\includegraphics[width=0.16\textwidth, height=0.14\textwidth]{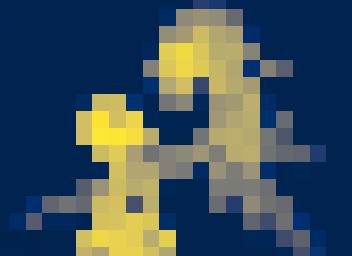} & 
\includegraphics[width=0.16\textwidth, height=0.14\textwidth]{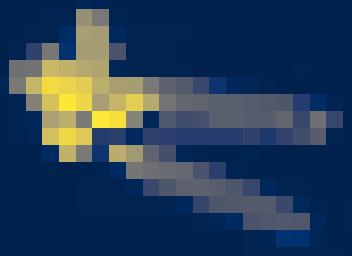} \\

\rotatebox{90}{\makecell{(d)~Ours \\Detection}} & \includegraphics[width=0.16\textwidth, height=0.14\textwidth]{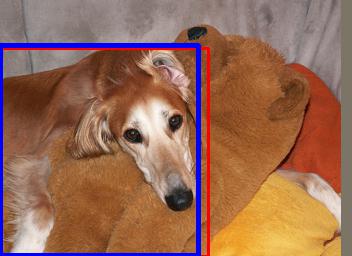} &
\includegraphics[width=0.16\textwidth, height=0.14\textwidth]{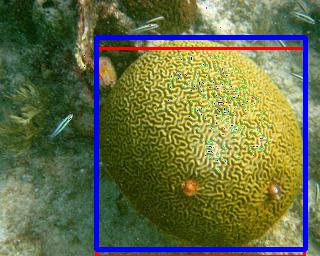} &
\includegraphics[width=0.16\textwidth, height=0.14\textwidth]{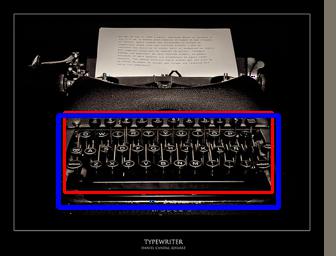} &
\includegraphics[width=0.16\textwidth, height=0.14\textwidth]{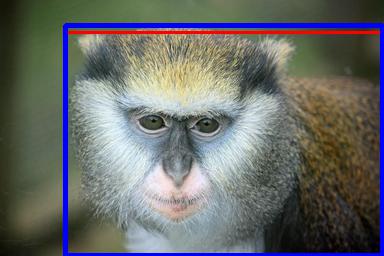} & 
\includegraphics[width=0.16\textwidth, height=0.14\textwidth]{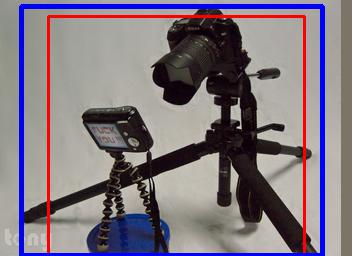} & 
\includegraphics[width=0.16\textwidth, height=0.14\textwidth]{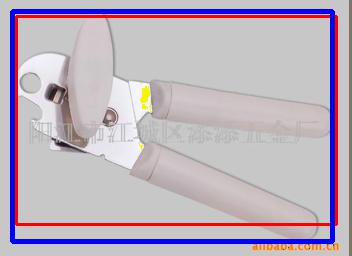} \\

\end{tabular}}
\captionsetup{type=figure}+\caption{\textbf{Visual results of unsupervised single object discovery on Imagenet-1k~\cite{deng2009imagenet}.} In (a), we show LOST ~\cite{simeoni2021localizing} the map of inverse degrees, which is used to perform detection (b). For TokenCut, we illustrate the eigenvector in (c) and the detection in (d). \textcolor{blue}{Blue} and \textcolor{red}{Red} bounding boxes indicate the ground-truth and the predicted bounding boxes respectively.}
\label{fig:imagenet}
\vspace{-5pt}
\end{center}
\end{minipage}

\section{Failure cases on CUB and Imagenet-1k}
\label{sec:failure}

We illustrate additional failure cases in Fig.~\ref{fig:failure2}. Those failure cases can be organised into three categories: 1) Where TokenCut focus on the largest salient object, whereas the annotation is highlights a different object, shown in the first and the second column in Fig.~\ref{fig:failure2}. 2) Similar to LOST, Tokencut is not able to differentiate the connected objects, such as the third and the fourth column in Fig.~\ref{fig:failure2}. 3) In case of occlusion, neither LOST nor our approach can't detect the entire object, such as the last two columns in Fig.~\ref{fig:failure2}. 

\begin{minipage}{\linewidth}
\begin{center}
\vspace{1em}
\resizebox{\linewidth}{!}{ 
\begin{tabular}{c@{\hskip 3pt}c@{\hskip 3pt}c@{\hskip 3pt}c@{\hskip 3pt}c@{\hskip 3pt}c@{\hskip 3pt}c}
\rotatebox{90}{\makecell{(a)~LOST Inverse \\ Degree Attention}} & 
\includegraphics[width=0.155\textwidth, height=0.14\textwidth]{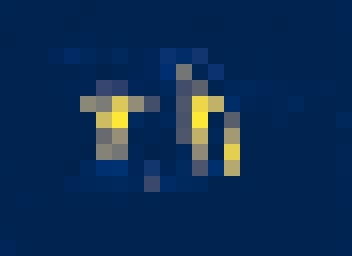} &
\includegraphics[width=0.16\textwidth, height=0.14\textwidth]{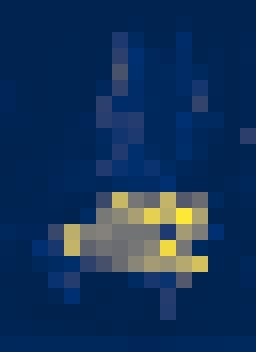} &
\includegraphics[width=0.16\textwidth, height=0.14\textwidth]{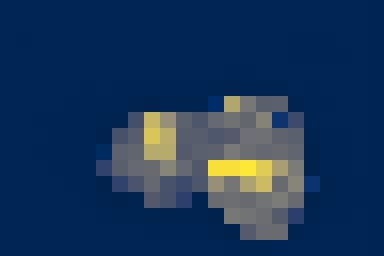} &
\includegraphics[width=0.16\textwidth, height=0.14\textwidth]{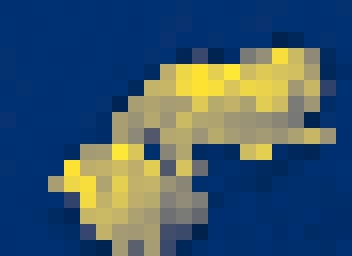} &
\includegraphics[width=0.16\textwidth, height=0.14\textwidth]{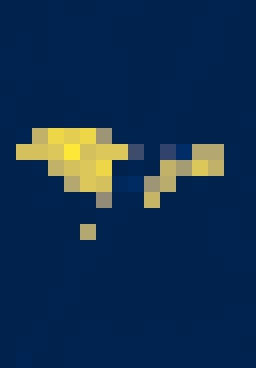} &
\includegraphics[width=0.16\textwidth, height=0.14\textwidth]{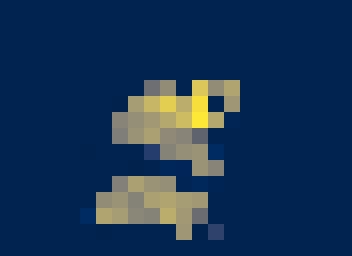} \\

\rotatebox{90}{\makecell{(b)~LOST \\Detection}} & 
\includegraphics[width=0.16\textwidth, height=0.14\textwidth]{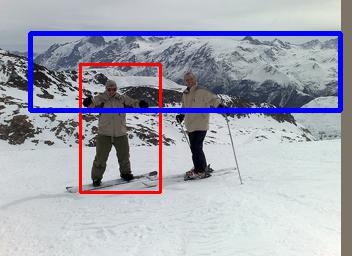} &
\includegraphics[width=0.16\textwidth, height=0.14\textwidth]{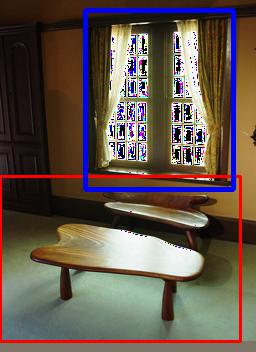} &
\includegraphics[width=0.16\textwidth, height=0.14\textwidth]{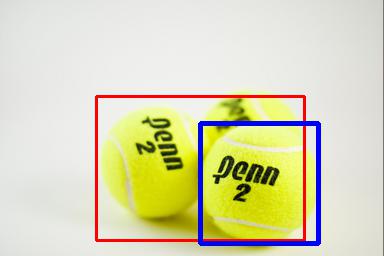} &
\includegraphics[width=0.16\textwidth, height=0.14\textwidth]{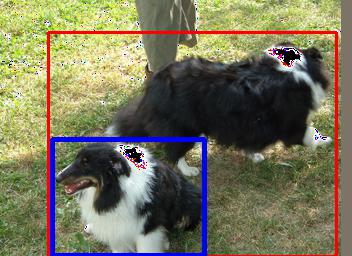} &
\includegraphics[width=0.16\textwidth, height=0.14\textwidth]{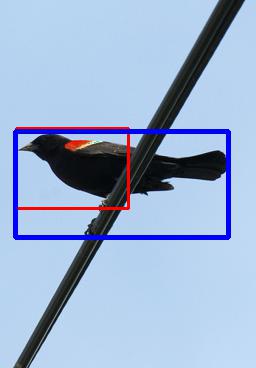} &
\includegraphics[width=0.16\textwidth, height=0.14\textwidth]{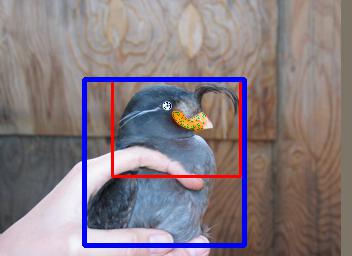}\\

\rotatebox{90}{\makecell{(c)~Ours Eigen \\Attention}} & 
\includegraphics[width=0.16\textwidth, height=0.14\textwidth]{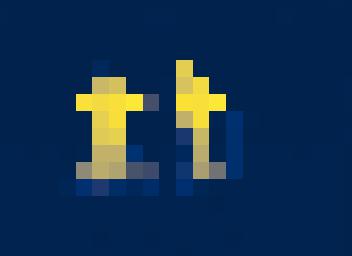} & 
\includegraphics[width=0.16\textwidth, height=0.14\textwidth]{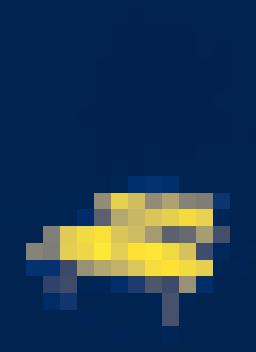} & 
\includegraphics[width=0.16\textwidth, height=0.14\textwidth]{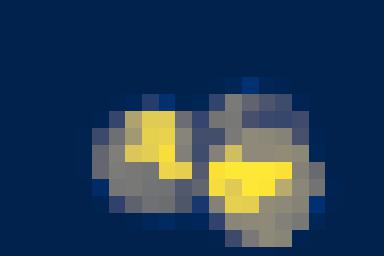} & 
\includegraphics[width=0.16\textwidth, height=0.14\textwidth]{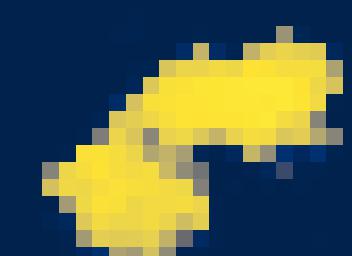} &
\includegraphics[width=0.16\textwidth, height=0.14\textwidth]{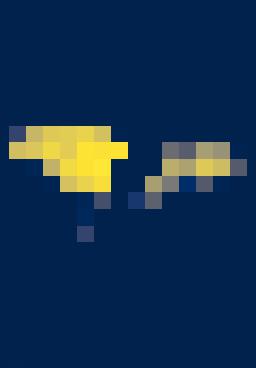} & 
\includegraphics[width=0.16\textwidth, height=0.14\textwidth]{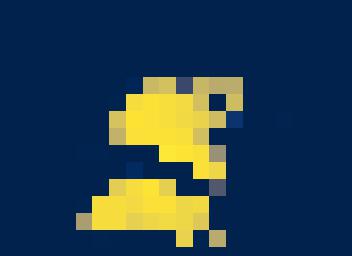}\\

\rotatebox{90}{\makecell{(d)~Ours \\Detection}} & 
\includegraphics[width=0.16\textwidth, height=0.14\textwidth]{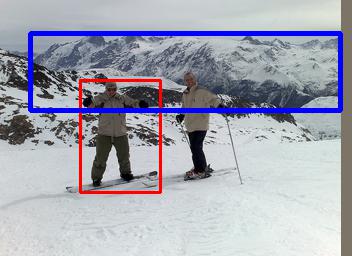} &
\includegraphics[width=0.16\textwidth, height=0.14\textwidth]{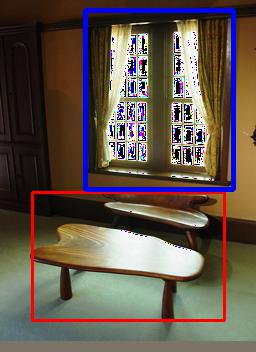} &
\includegraphics[width=0.16\textwidth, height=0.14\textwidth]{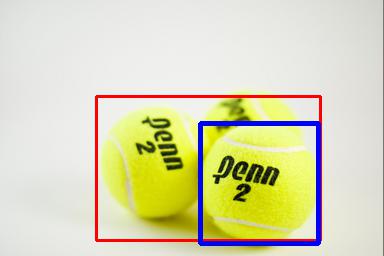} &
\includegraphics[width=0.16\textwidth, height=0.14\textwidth]{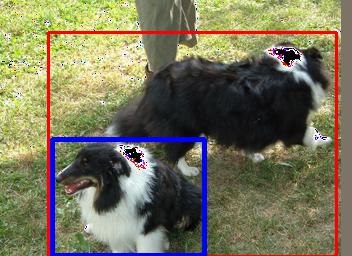} & 
\includegraphics[width=0.16\textwidth, height=0.14\textwidth]{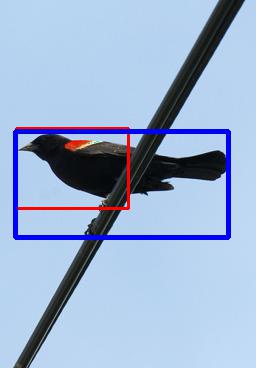} &
\includegraphics[width=0.16\textwidth, height=0.14\textwidth]{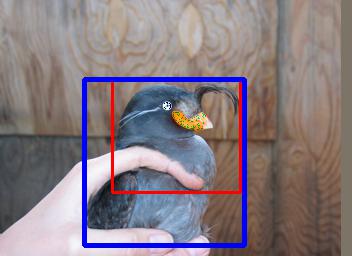}\\

\end{tabular}}
\captionsetup{type=figure}+\caption{\textbf{Failure cases on Imagenet-1k~\cite{deng2009imagenet} and CUB~\cite{WahCUB_200_2011}.} In (a), we show LOST~\cite{simeoni2021localizing} the map of inverse degrees, which is used to perform detection (b). For TokenCut, we illustrate the eigenvector in (c) and the detection in (d). \textcolor{blue}{Blue} and \textcolor{red}{Red} bounding boxes indicate the ground-truth and the predicted bounding boxes respectively.}
\label{fig:failure2}
\vspace{-5pt}
\end{center}
\end{minipage}

\section{Datasets}

We present in this section the details of the datasets used in our experiments: 
\begin{Itemize}
\item \textbf{VOC07 and VOC12}  correspond to the training and validation set of PASCAL-VOC07 and PASCAL-VOC12. VOC07 and VOC12 contain 5 011 and 11 540 images respectively which belong to 20 categories. They are commonly evaluated for unsupervised object discovery~\cite{vo2020toward,vo2021large,vo2019unsupervised, wei2019unsupervised,cho2015unsupervised}. 

\item \textbf{COCO20K} consists of 19 817 randomly chosen images from the COCO2014 dataset~\cite{lin2014microsoft}. It is used as a benchmark in~\cite{vo2020toward} for a large scale evaluation. 

\item \textbf{CUB} consists of 200 bird species, including 6 033 and 5 755 images in training and test sets respectively, which is commonly used to evaluate weakly supervised object localization~\cite{zhou2016learning,singh2017hide,bae2020rethinkingCAM,szegedy2015going,choe2019attention}.

\item \textbf{ImageNet}~\cite{deng2009imagenet} is a widely used benchmark for image classification and object detection, which consists of 1 000 different categories. The number of images in training and validation sets are 1.3 million and 50,000 respectively. Each image contains a single object supposed to be detected. During the training, only class labels are available. 

\item \textbf{ECSSD} contains 1 000 real-world images of complex scenes for testing. 

\item \textbf{DUTS} contains 10 553 train and 5 019 test images. The training set is collected from the ImageNet detection train/val set. The test set is collected from ImageNet test, and the SUN dataset~\cite{Xiao2010Sun}. Following the previous works~\cite{shen2021learning}, we report the performance on the DUTS-test subset. 

\item \textbf{DUT-OMRON}~\cite{yang2013saliency} contains 5 168 images of high quality natural images for testing.
\end{Itemize}

\section{Visual results for unsupervised saliency detecion on ECSSD, DUTS and DUT-OMRON}
\label{sec:usd}

\begin{figure*}[!ht]
\begin{tabular}{c@{\hskip 1.3pt}c@{\hskip 1.3pt}c@{\hskip 1.3pt}c@{\hskip 1.3pt}c@{\hskip 1.3pt}c}
		
		\includegraphics[width=0.16\textwidth, height=0.14\textwidth]{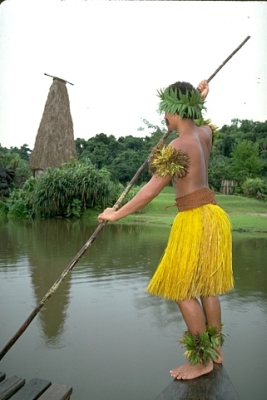} &
		\includegraphics[width=0.16\textwidth, height=0.14\textwidth]{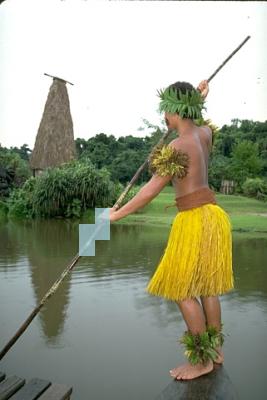} &
		\includegraphics[width=0.16\textwidth, height=0.14\textwidth]{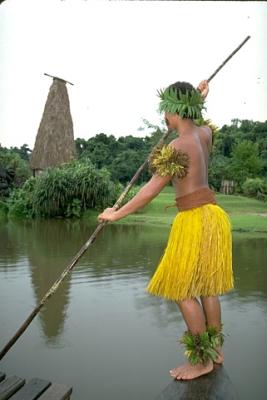} &
		\includegraphics[width=0.16\textwidth, height=0.14\textwidth]{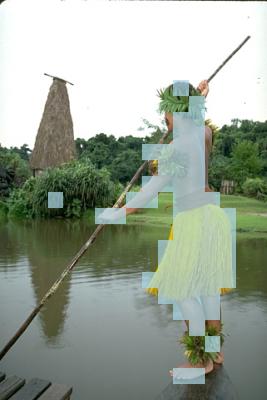} &
		\includegraphics[width=0.16\textwidth, height=0.14\textwidth]{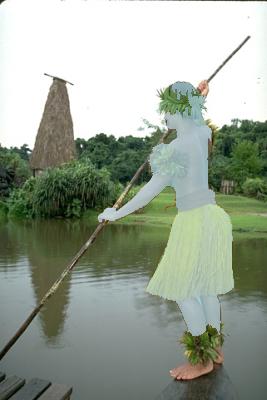} &
		\includegraphics[width=0.16\textwidth, height=0.14\textwidth]{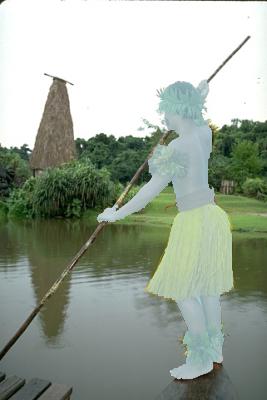} \\
		
		\includegraphics[width=0.16\textwidth, height=0.14\textwidth]{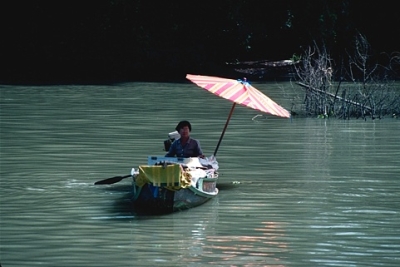} &
		\includegraphics[width=0.16\textwidth, height=0.14\textwidth]{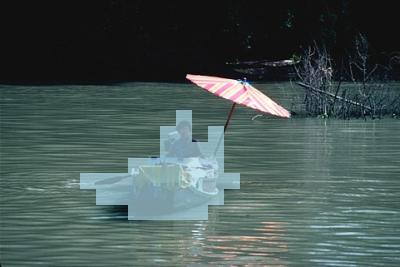} &
		\includegraphics[width=0.16\textwidth, height=0.14\textwidth]{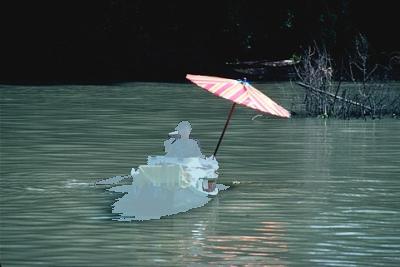} &
		\includegraphics[width=0.16\textwidth, height=0.14\textwidth]{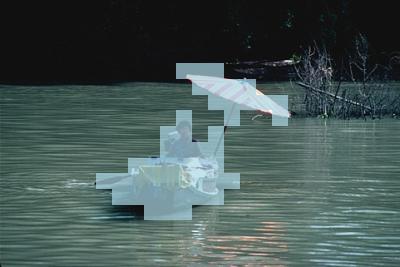} &
		\includegraphics[width=0.16\textwidth, height=0.14\textwidth]{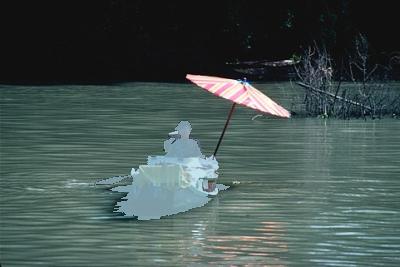} &
		\includegraphics[width=0.16\textwidth, height=0.14\textwidth]{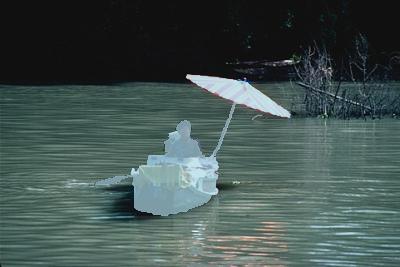} \\
		
		\includegraphics[width=0.16\textwidth, height=0.14\textwidth]{./supplementary/ecssd/0021.jpg} &
		\includegraphics[width=0.16\textwidth, height=0.14\textwidth]{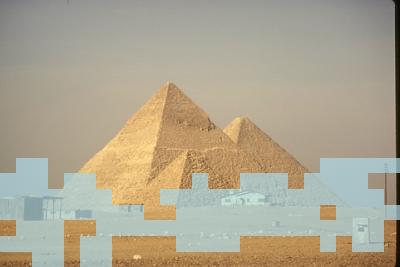} &
		\includegraphics[width=0.16\textwidth, height=0.14\textwidth]{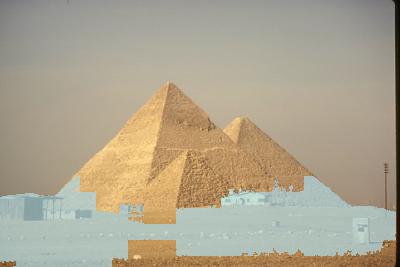} &
		\includegraphics[width=0.16\textwidth, height=0.14\textwidth]{./supplementary/ecssd/0021_tokencut.jpg} &
		\includegraphics[width=0.16\textwidth, height=0.14\textwidth]{./supplementary/ecssd/0021_tokencut_bfs.jpg} &
		\includegraphics[width=0.16\textwidth, height=0.14\textwidth]{./supplementary/ecssd/0021_gt.jpg} \\
		
		\includegraphics[width=0.16\textwidth, height=0.14\textwidth]{./supplementary/ecssd/0049.jpg} &
		\includegraphics[width=0.16\textwidth, height=0.14\textwidth]{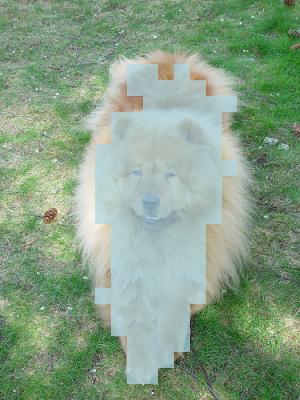} &
		\includegraphics[width=0.16\textwidth, height=0.14\textwidth]{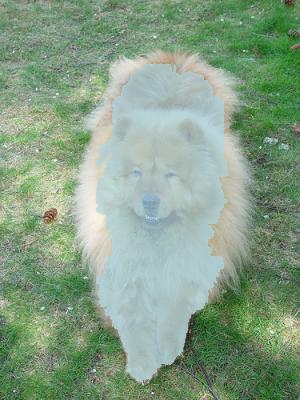} &
		\includegraphics[width=0.16\textwidth, height=0.14\textwidth]{./supplementary/ecssd/0049_tokencut.jpg} &
		\includegraphics[width=0.16\textwidth, height=0.14\textwidth]{./supplementary/ecssd/0049_tokencut_bfs.jpg} &
		\includegraphics[width=0.16\textwidth, height=0.14\textwidth]{./supplementary/ecssd/0049_gt.jpg} \\

		\makecell{(a) Input}  & \makecell{(b) LOST} & \makecell{(c) LOST + BS} & \makecell{(d) Ours} & \makecell{(e) Ours + BS} & \makecell{(f) GT} \\
\end{tabular}
\vspace{-0.2cm}
\caption{\textbf{Visual results of unsupervised segments on ECSSD~\cite{shi2015hierarchical}}}
\label{fig: ecssd}
\vspace{-0.2cm}
\end{figure*}

\begin{figure*}[!t]
\vspace{-10pt}
\begin{tabular}{c@{\hskip 1.3pt}c@{\hskip 1.3pt}c@{\hskip 1.3pt}c@{\hskip 1.3pt}c@{\hskip 1.3pt}c}
		
		\includegraphics[width=0.16\textwidth, height=0.14\textwidth]{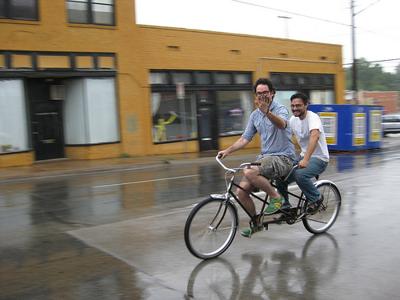} &
		\includegraphics[width=0.16\textwidth, height=0.14\textwidth]{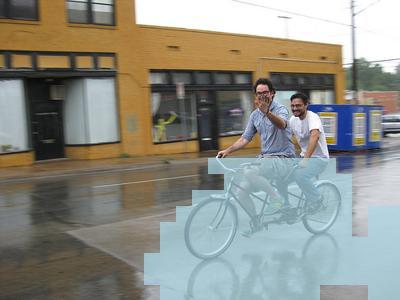} &
		\includegraphics[width=0.16\textwidth, height=0.14\textwidth]{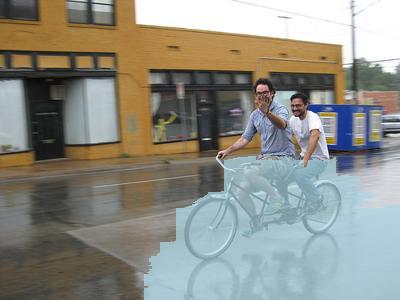} &
		\includegraphics[width=0.16\textwidth, height=0.14\textwidth]{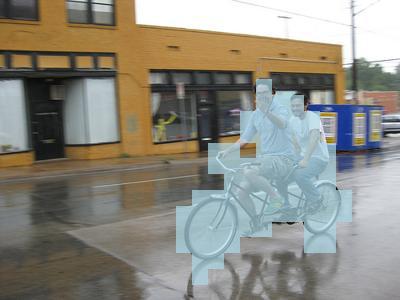} &
		\includegraphics[width=0.16\textwidth, height=0.14\textwidth]{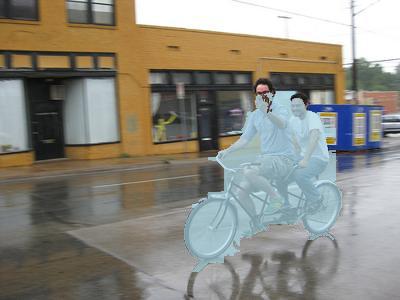} &
		\includegraphics[width=0.16\textwidth, height=0.14\textwidth]{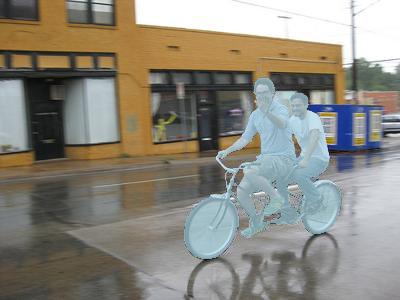} \\

		\includegraphics[width=0.16\textwidth, height=0.14\textwidth]{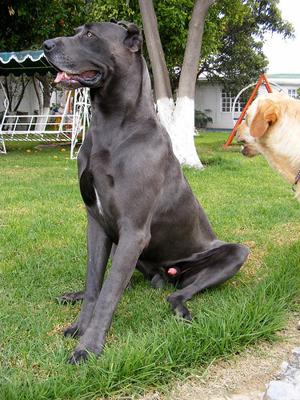} &
		\includegraphics[width=0.16\textwidth, height=0.14\textwidth]{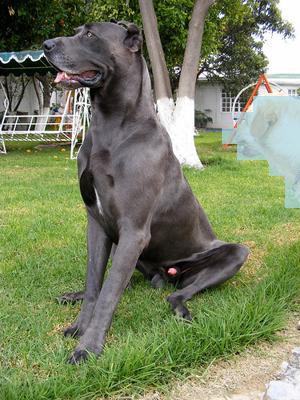} &
		\includegraphics[width=0.16\textwidth, height=0.14\textwidth]{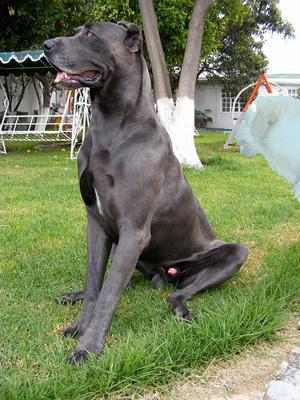} &
		\includegraphics[width=0.16\textwidth, height=0.14\textwidth]{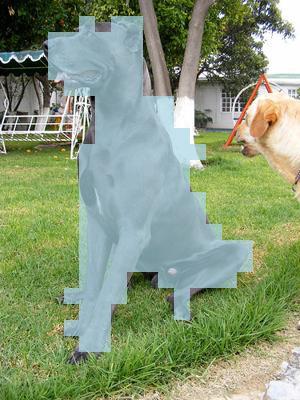} &
		\includegraphics[width=0.16\textwidth, height=0.14\textwidth]{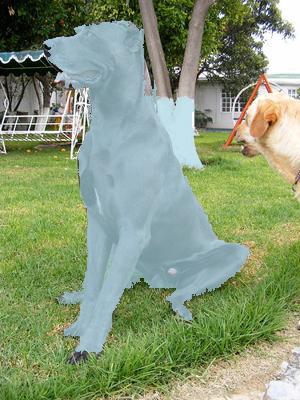} &
		\includegraphics[width=0.16\textwidth, height=0.14\textwidth]{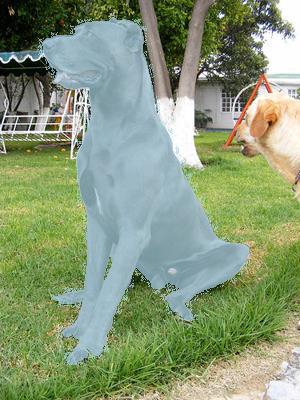} \\
		
		\includegraphics[width=0.16\textwidth, height=0.14\textwidth]{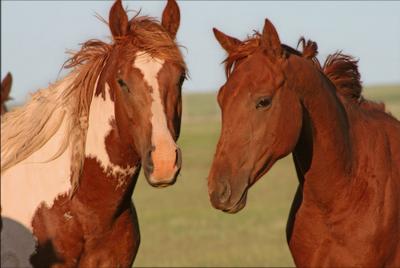} &
		\includegraphics[width=0.16\textwidth, height=0.14\textwidth]{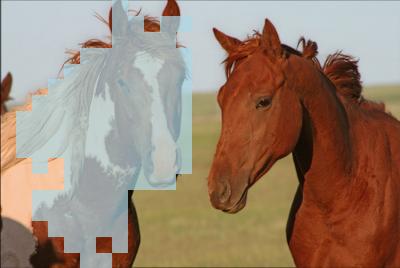} &
		\includegraphics[width=0.16\textwidth, height=0.14\textwidth]{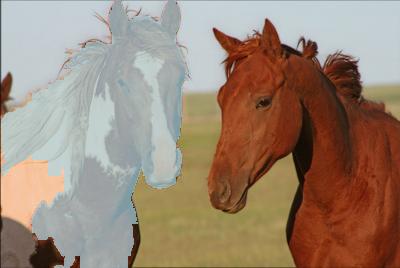} &
		\includegraphics[width=0.16\textwidth, height=0.14\textwidth]{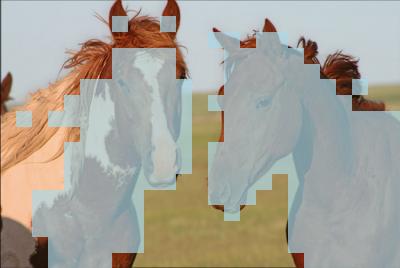} &
		\includegraphics[width=0.16\textwidth, height=0.14\textwidth]{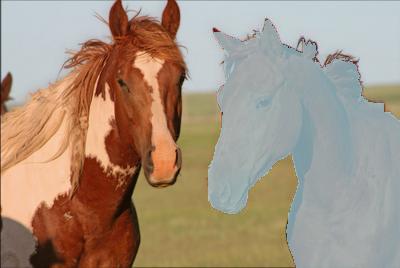} &
		\includegraphics[width=0.16\textwidth, height=0.14\textwidth]{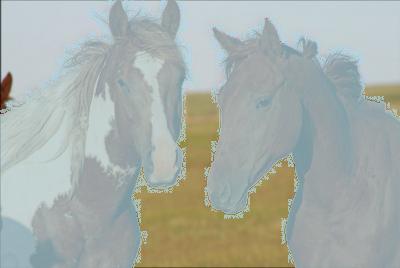} \\
		
		\includegraphics[width=0.16\textwidth, height=0.14\textwidth]{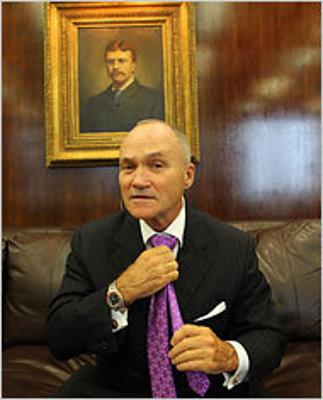} &
		\includegraphics[width=0.16\textwidth, height=0.14\textwidth]{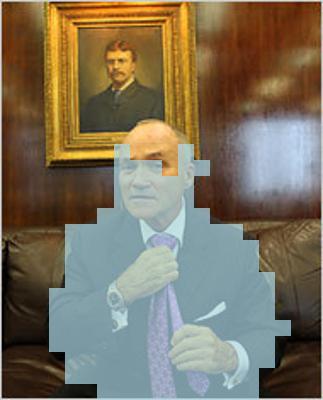} &
		\includegraphics[width=0.16\textwidth, height=0.14\textwidth]{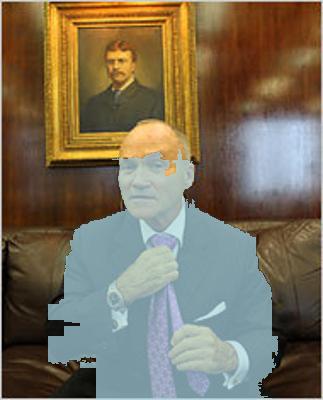} &
		\includegraphics[width=0.16\textwidth, height=0.14\textwidth]{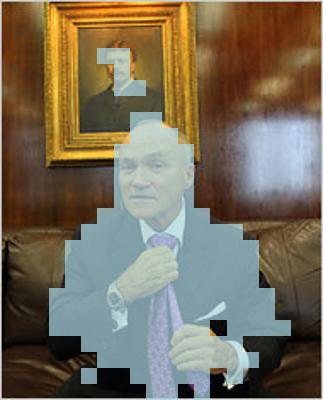} &
		\includegraphics[width=0.16\textwidth, height=0.14\textwidth]{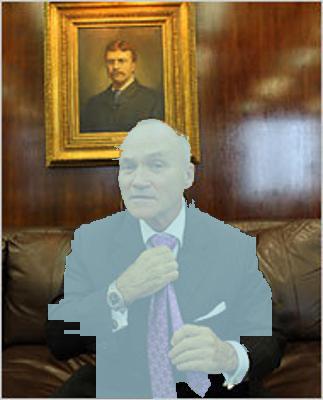} &
		\includegraphics[width=0.16\textwidth, height=0.14\textwidth]{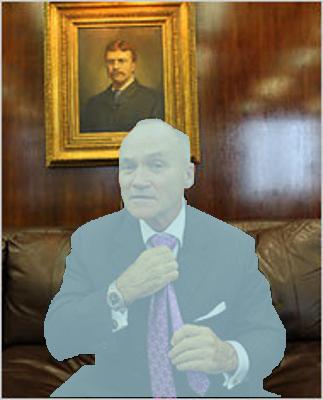} \\
		
		\makecell{(a) Input}  & \makecell{(b) LOST} & \makecell{(c) LOST + BS} & \makecell{(d) Ours} & \makecell{(e) Ours + BS} & \makecell{(f) GT} \\
\end{tabular}
\vspace{-0.2cm}
\caption{\textbf{Visual results of unsupervised segments on DUTS~\cite{wang2017learning}}}
\vspace{-0.2cm}
\label{fig: duts}

\end{figure*}

\begin{figure*}[!b]
\begin{tabular}{c@{\hskip 1.3pt}c@{\hskip 1.3pt}c@{\hskip 1.3pt}c@{\hskip 1.3pt}c@{\hskip 1.3pt}c}
		
		\includegraphics[width=0.16\textwidth, height=0.14\textwidth]{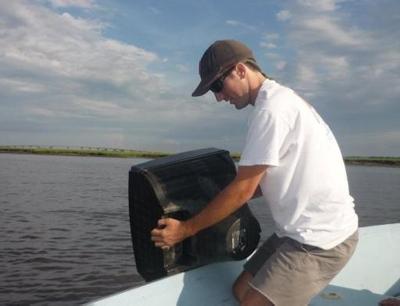} &
		\includegraphics[width=0.16\textwidth, height=0.14\textwidth]{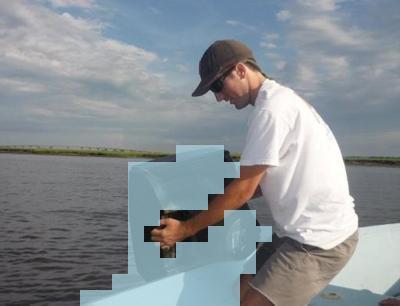} &
		\includegraphics[width=0.16\textwidth, height=0.14\textwidth]{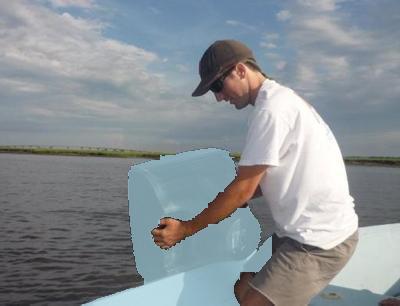} &
		\includegraphics[width=0.16\textwidth, height=0.14\textwidth]{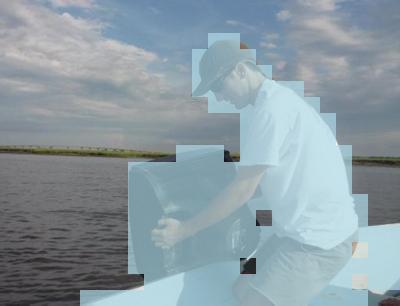} &
		\includegraphics[width=0.16\textwidth, height=0.14\textwidth]{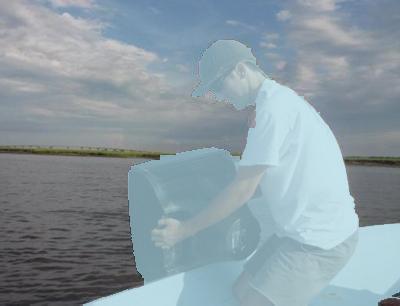} &
		\includegraphics[width=0.16\textwidth, height=0.14\textwidth]{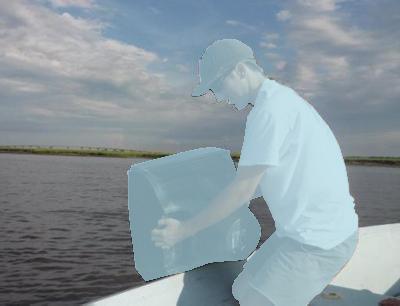} \\
		
		\includegraphics[width=0.16\textwidth, height=0.14\textwidth]{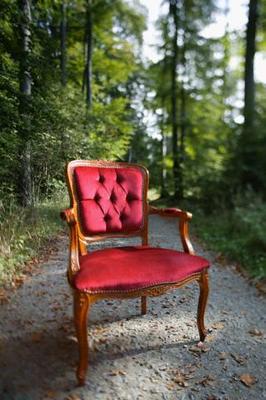} &
		\includegraphics[width=0.16\textwidth, height=0.14\textwidth]{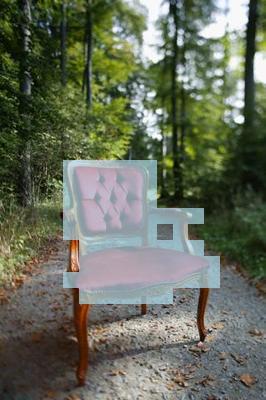} &
		\includegraphics[width=0.16\textwidth, height=0.14\textwidth]{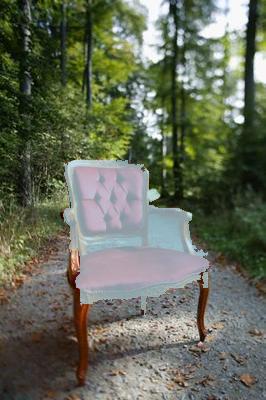} &
		\includegraphics[width=0.16\textwidth, height=0.14\textwidth]{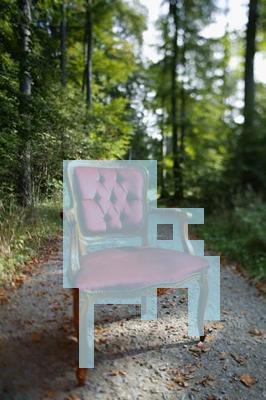} &
		\includegraphics[width=0.16\textwidth, height=0.14\textwidth]{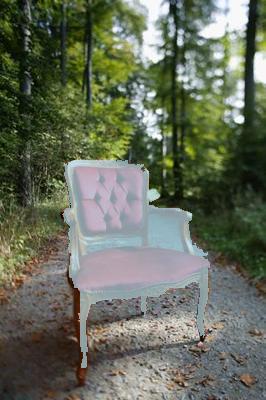} &
		\includegraphics[width=0.16\textwidth, height=0.14\textwidth]{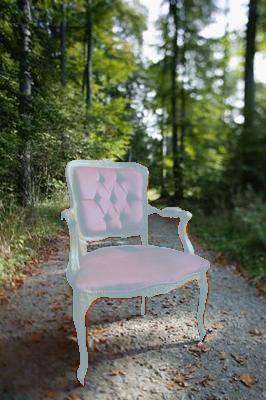} \\
		
		\includegraphics[width=0.16\textwidth, height=0.14\textwidth]{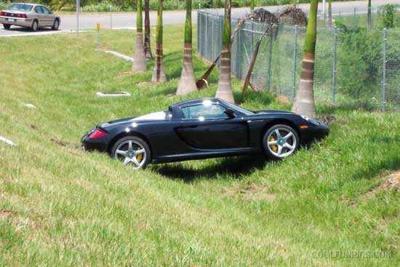} &
		\includegraphics[width=0.16\textwidth, height=0.14\textwidth]{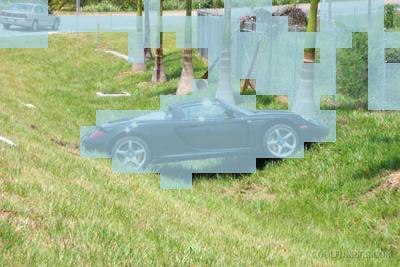} &
		\includegraphics[width=0.16\textwidth, height=0.14\textwidth]{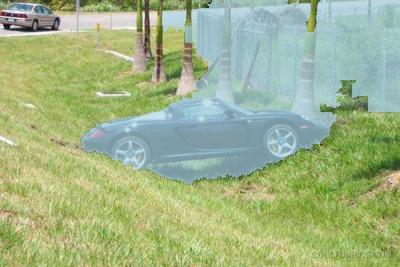} &
		\includegraphics[width=0.16\textwidth, height=0.14\textwidth]{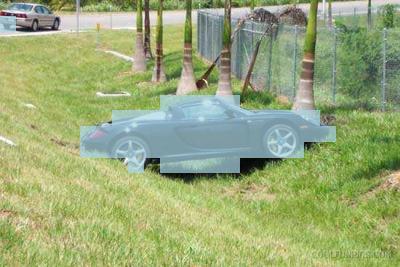} &
		\includegraphics[width=0.16\textwidth, height=0.14\textwidth]{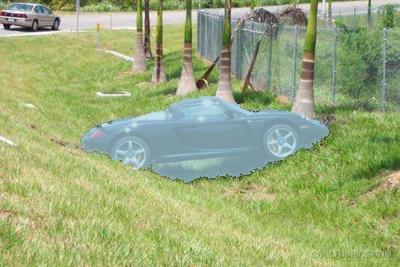} &
		\includegraphics[width=0.16\textwidth, height=0.14\textwidth]{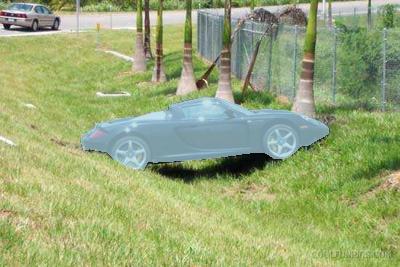} \\
		
		\includegraphics[width=0.16\textwidth, height=0.14\textwidth]{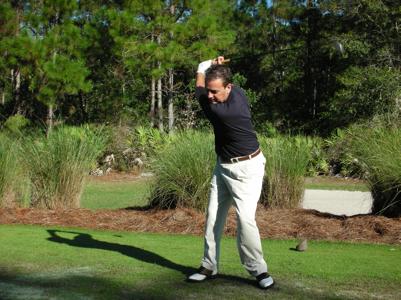} &
		\includegraphics[width=0.16\textwidth, height=0.14\textwidth]{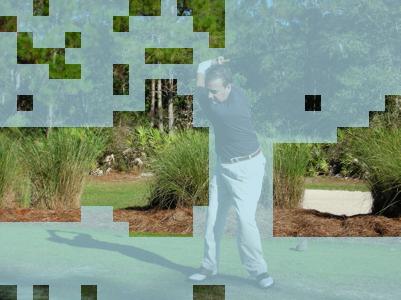} &
		\includegraphics[width=0.16\textwidth, height=0.14\textwidth]{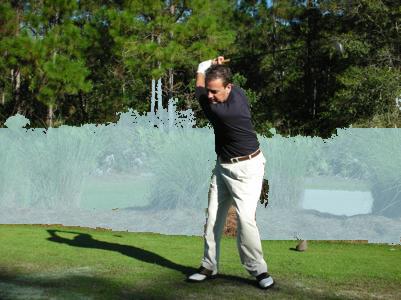} &
		\includegraphics[width=0.16\textwidth, height=0.14\textwidth]{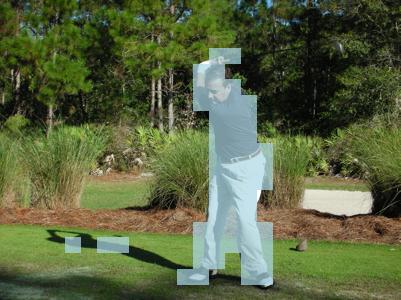} &
		\includegraphics[width=0.16\textwidth, height=0.14\textwidth]{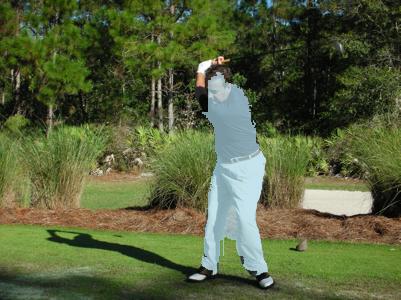} &
		\includegraphics[width=0.16\textwidth, height=0.14\textwidth]{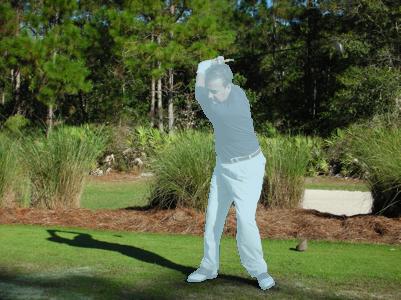} \\
		
		\makecell{(a) Input}  & \makecell{(b) LOST} & \makecell{(c) LOST + BS} & \makecell{(d) Ours} & \makecell{(e) Ours + BS} & \makecell{(f) GT} \\
\end{tabular}
\caption{\textbf{Visual results of unsupervised segments on DUT-OMRON~\cite{yang2013saliency}}}
\label{fig: dut}
\end{figure*}

We present visual results for unsupervised saliency detecion on ECSSD~\cite{shi2015hierarchical}, DUTS~\cite{wang2017learning} and DUT-OMRON~\cite{yang2013saliency} in Fig.~\ref{fig: ecssd},~\ref{fig: duts} and~\ref{fig: dut} respectively.

For each dataset, we compare LOST segmentation, LOST + Bilateral Solver and our approch. The TokenCut provides better segmentation on objects. The performance is further improved with Bilateral Solver.

\end{document}